%% file: CLBenchreal.tex
\documentclass{article}

\usepackage{iclr2026_conference,times}
\input{mypackage}

\title{CL-bench Life: Can Language Models Learn\\from Real-Life Context?}

\author{
\large
\textbf{Hunyuan Team, Tencent \quad Fudan University$^*$}
\\
\\
\href{https://clbench.com}{\texttt{https://clbench.com}}
}

\iclrfinalcopy

\begin{document}

\maketitle

\AddToShipoutPictureFG*{%
  \AtPageUpperLeft{%
    \raisebox{-1.2cm}{%
      \hspace{3.8cm}%
      \includegraphics[height=0.7cm]{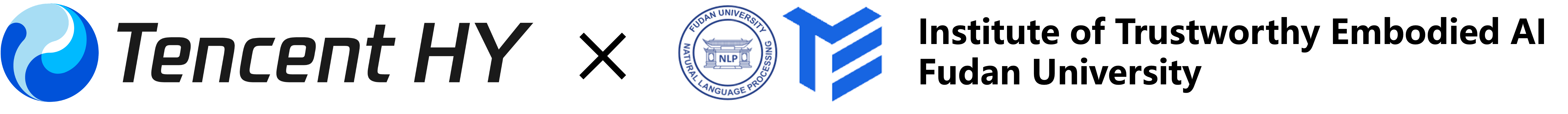}%
    }%
  }%
}

\begingroup
\renewcommand\thefootnote{}
\footnote{$^*$The full list of authors appears at the end of the paper. Correspondence can be sent to \texttt{\href{mailto:shihandou@foxmail.com}{shihandou@foxmail.com}}, \texttt{\href{mailto:tgui@fudan.edu.cn}{tgui@fudan.edu.cn}}, and \texttt{\href{mailto:plutozhou096@foxmail.com}{plutozhou096@foxmail.com}}.}
\addtocounter{footnote}{-1}
\endgroup

\vspace{-1em}
\begin{abstract}

Today’s AI assistants such as OpenClaw are designed to handle context effectively, making context learning an increasingly important capability for models.
As these systems move beyond professional settings into everyday life, the nature of the contexts they must handle also shifts.
Real-life contexts are often messy, fragmented, and deeply tied to personal and social experience, such as multi-party conversations, personal archives, and behavioral traces.
Yet it remains unclear whether current frontier language models can reliably learn from such contexts and solve tasks grounded in them.
To this end, we introduce \textbf{CL-bench Life}, a fully human-curated benchmark comprising 405 context-task pairs and 5,348 verification rubrics, covering common real-life scenarios.
Solving tasks in CL-bench Life requires models to reason over complex, messy real-life contexts, calling for strong \textbf{real-life context learning} abilities that go far beyond those evaluated in existing benchmarks.
We evaluate ten frontier LMs and find that real-life context learning remains highly challenging: even the best-performing model achieves only 19.3\% task solving rate, while the average performance across models is only 13.8\%.
Models still struggle to reason over contexts such as messy group chat histories and fragmented behavioral records from everyday life.
CL-bench Life provides a crucial testbed for advancing real-life context learning, and progress on it can enable more intelligent and reliable AI assistants in everyday life.

\end{abstract}

\section{Introduction}
\label{sec:introduction}

Real-world tasks often require models to search, organize, and reason over large amounts of context, rather than rely solely on pre-trained knowledge \cite{zhou2024webarena,lewis2020retrieval,xu2024searchinthechain,liu2024lost}.
This has driven the development of AI assistants such as OpenClaw \cite{openclaw2025} that excel at handling context.
For the models that power these systems, the focus of improvement is accordingly shifting from enhancing their ability to reason over pre-trained knowledge to improving their ability to reason over complex context \cite{anthropic2024claude,yang2025qwen3,qwen36plus,openai2025gpt5}. 
The latter is commonly referred to as \textbf{context learning}, a foundational capability for handling complex context effectively \cite{dou2026cl,mei2025survey,anthropic2025context}.

At the same time, as AI assistants become more widely used, users increasingly expect them to handle complex real-life context and support everyday tasks \cite{brynjolfsson2025generative,yao2024tau,wang2024survey}. 
By real-life context, we mean context grounded in everyday communication, scattered notes, and behavioral traces, rather than in professional or domain-specific sources.
Such context is often messier, more fragmented, more socially grounded, and temporally dispersed, with relevant information scattered across multi-party interactions and noisy records.
For example, users may ask models to reason over discussions in public threads or meetings, or to analyze personal digital traces and everyday behavioral records.

However, existing benchmarks primarily evaluate models’ context learning ability in more professional and domain-specific contexts \cite{dou2026cl,hu2026cl4sebenchmarkingcontextlearning,ding2026octobench}, such as finance, science, and code. 
They fail to capture whether current frontier language models (LMs) can reliably learn from such real-life context.
As a result, there is a pressing need for a benchmark that better reflects models’ \textbf{real-life context learning} ability and helps shape their future development.

To fill this gap, we introduce \textbf{CL-bench Life}, a benchmark for comprehensively evaluating context learning ability in real-life scenarios, comprising 405 expert-curated context-task pairs and 5,348 verification rubrics.
Each task requires models to learn from complex, messy real-life context and reason over new information, rather than merely relying on pre-trained knowledge.
To reflect the diversity of everyday context that models must reason over, CL-bench Life is organized into three categories: communication \& social interactions, fragmented information \& revisions, and behavioral records \& activity trails.
Specifically, \textbf{(1)} the first category covers socially grounded and multi-party contexts such as private chats, group discussions, meeting transcripts, and public community interactions.
\textbf{(2)} The second focuses on fragmented information and revision processes, including personal notes, public information streams, and creation or revision histories. 
\textbf{(3)} The third focuses on behavioral and activity-based records, such as game logs, digital footprints, browsing streams, and long-term records of daily activity.

These three categories capture three broad facets of everyday human life, corresponding to people's interpersonal interactions, the fragmented information they record or revise, and the behavioral traces they leave through daily activity.
Together, three categories cover common everyday scenarios, enabling a broad evaluation of models' real-life context learning ability.
Moreover, for each task, the corresponding context is self-contained, so solving the task does not require external retrieval, allowing the evaluation to more cleanly isolate the ability to learn from context.

\begin{figure}[t]
\centering
\includegraphics[width=1\textwidth]{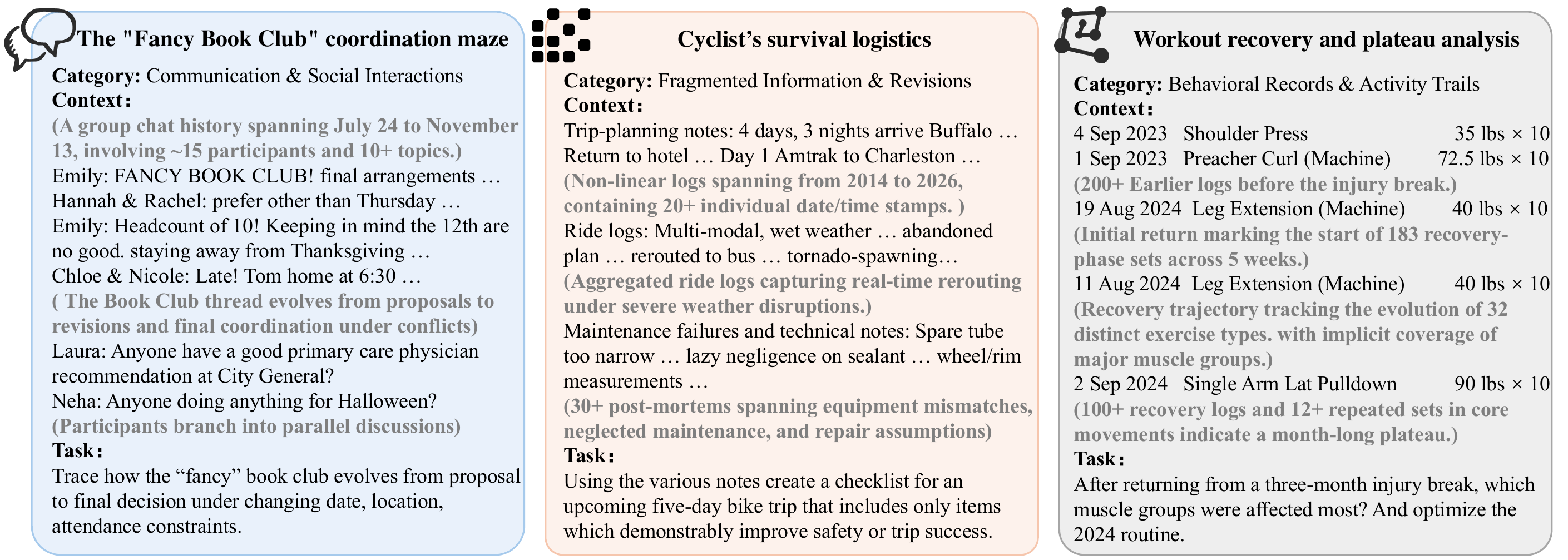}
\caption{
Three simplified examples of real-life contexts and tasks.
They illustrate how real-life contexts are often messier, more fragmented, and more socially grounded than professional-domain contexts.
Further explanation of these examples is provided in Figure~\ref{fig:case3}.
}
\label{fig:cl-bench}
\end{figure}

We evaluate ten state-of-the-art language models on CL-bench Life and find that real-life context learning remains highly challenging. 
On average, the models solve only 13.8\% of tasks, and even the best-performing model, GPT-5.4, solves only 19.3\%. 
Increasing reasoning effort generally improves context learning performance, but the extra token cost required to obtain similar gains varies substantially across models.
We also find that task solve rate is not strongly correlated with context length, suggesting that poor real-life context learning cannot be simply attributed to limited long-context ability.
Moreover, in group chat scenarios, models often exhibit confusion over user identities and references, leading them to misunderstand the context and fail to solve the task.

We present additional findings in Sections~\ref{sec:experiments} and~\ref{sec:analysis}. 
Overall, real-life context learning is a critical capability for models to solve everyday tasks effectively. 
Improving this ability can help enable more intelligent, practical, and capable AI assistants for real-life tasks.
CL-bench Life provides a rigorous benchmark for evaluating this capability and informing the future development of LMs in this direction.

\section{Related Work}
\label{sec:related-work}

Real-world tasks are rarely solved in a single step and typically involve multi-step processes in which AI systems must continually acquire, organize, and utilize large amounts of complex contextual information \cite{liu2024agentbench,zhou2024webarena,jimenez2024swebench,yang-etal-2018-hotpotqa}.
Accordingly, existing research can be broadly divided into two lines. 
One line focuses on the end-to-end capabilities of agents or AI assistants, such as how they leverage scaffolding to complete tasks \cite{schick2023toolformer,zhou2024language,mialon2023gaiabenchmarkgeneralai}. 
The other line focuses on the context-related abilities of the underlying models themselves, such as whether they can effectively understand, organize, and learn from context \cite{bai-etal-2024-longbench,bai-etal-2025-longbench,li-etal-2024-loogle,hsieh2024ruler,yen2025helmet,an-etal-2024-l}. 
CL-bench Life belongs more naturally to the latter line and further focuses on real-life context learning.

\textbf{Benchmarks for agents.}
Many benchmarks aim to evaluate the overall capabilities of agents \cite{ma2024agentboard,ToolEyes,xie2024travelplanner,jang2025videowebarena,deng2023mindweb,yao2022webshop,2019textworldlearningenvironmenttextbased}. 
These benchmarks formulate tasks as multi-step processes of reasoning and acting, where models must continuously acquire, organize, and use contextual information through interactions with users and diverse tools, and then decide on the next action \cite{qin2024toolllm,xie2024osworld,yao2024tau,schick2023toolformer}. 
In such settings, agent performance depends not only on task decomposition and tool use, but also on whether the underlying LM can effectively handle context and continue reasoning and acting based on that information \cite{sumers2024cognitive}. 
However, these end-to-end benchmarks usually couple together many different abilities. When a model fails, it is often unclear whether the failure comes from tool use, memory retrieval, or the more fundamental ability of context learning itself \cite{dou2026cl,cemri2025why}.

In contrast, CL-bench Life provides a more direct and controlled evaluation of context learning as a standalone ability. 
It can be viewed as simulating the final step of agent problem solving: all task-relevant context is already provided, and the model is asked to solve the task based on that context alone. 
This largely removes errors introduced by earlier stages such as search or context organization, allowing context learning to be evaluated more directly.

\textbf{Benchmarks for context-related abilities.}
Another line of work focuses more directly on the context-related abilities of LMs themselves \cite{garg2022what,mei2025survey,dong2024survey,zhang2024infty}. 
Some benchmarks evaluate long-context understanding, examining whether LMs can retrieve information and reason effectively under long-input settings \cite{kwan-etal-2024-m4le,an-etal-2024-l,bai2024longbench,dong2024bamboo,li2024loogle,yen2024helmet}. 
Unlike these benchmarks, CL-bench Life is more concerned with whether models can reason over messy and fragmented real-life context effectively. 
In fact, our results show that real-life context learning is not strongly correlated with long-context ability, and the contexts in CL-bench Life generally do not exceed the context windows of current frontier LMs.

\begin{figure}[t]
\centering
\includegraphics[width=0.92\textwidth]{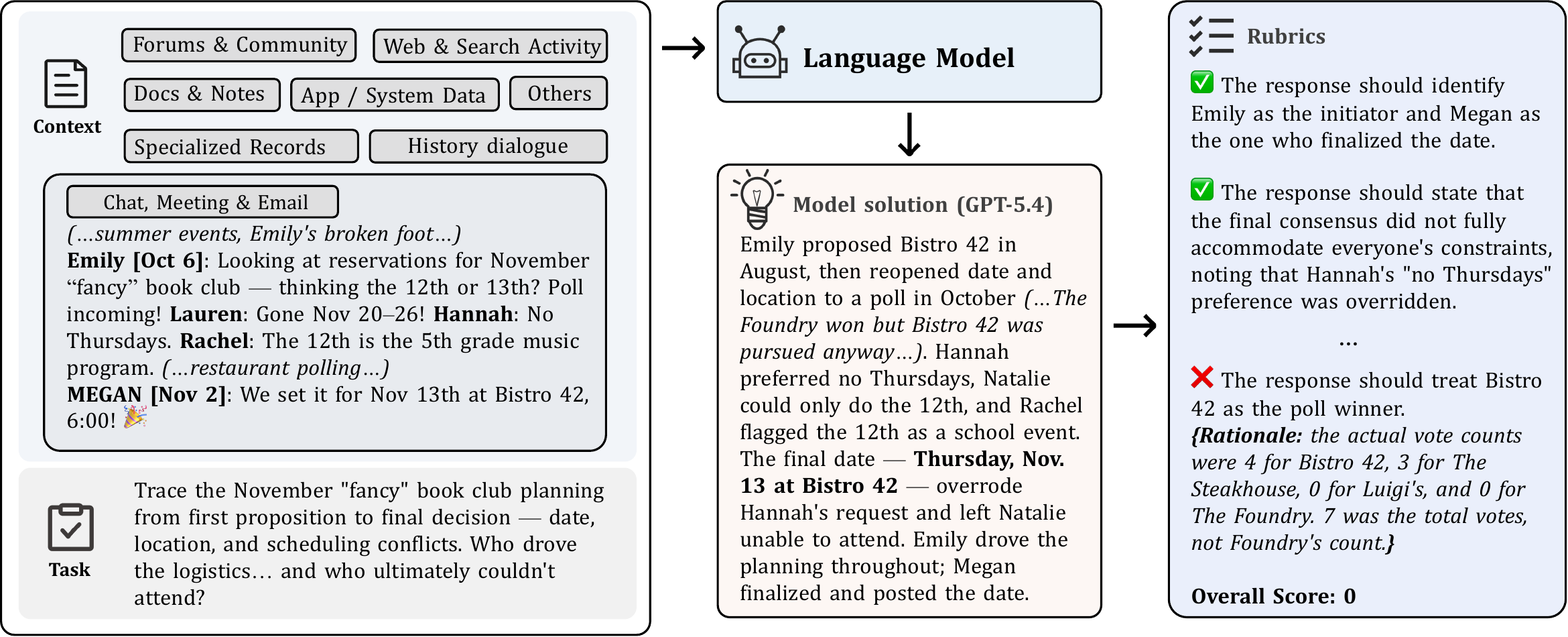}
\caption{
Each instance in CL-bench Life consists of a task, a context, and a set of verification rubrics. 
The language model must reason over the provided context to solve the task, without retrieving additional external information. 
This example task in this figure requires reasoning over a long, informal group chat history to reconstruct an evolving real-life coordination process. 
The model must trace the planning of the November ``fancy'' book club meeting over time, resolve conflicting constraints, determine the final date and location, and identify who led and finalized the logistics.
}
\label{fig:cl-bench}
\end{figure}

Some benchmarks evaluate context learning more directly in professional, scientific \cite{ying2026retrieval}, and coding domains \cite{ding2026octobench}, such as CL-bench \cite{dou2026cl} and CL4SE \cite{hu2026cl4sebenchmarkingcontextlearning}. 
CL-bench Life is closely related to this line of work, but targets a different context-learning setting. 
Prior benchmarks focus mainly on professional and domain-specific contexts, which are more focused and relatively post-organized, where models must master novel, complex knowledge such as rules or procedures. 
In contrast, CL-bench Life focuses on more fragmented, and less organized real-life contexts, where models must piece together scattered clues and reason robustly.

There are also benchmarks that focus on context management or memory, evaluating whether models can store and retrieve historical information across long-term interactions \cite{lee-etal-2025-ethic,wu2025longmemeval,kuratov2024babilong,song2025counting,zhang2025longcite}. 
CL-bench Life differs from these settings in an important way: its tasks are solved in a one-shot setting, and all task-relevant new information has already been organized and provided to the model. 
Models do not need to call additional tools or perform retrieval. 
As a result, CL-bench Life focuses less on how AI systems obtain context, and more on whether models can effectively learn from and use context once it is given.

\textbf{Benchmarks for real-life scenarios.}
Beyond benchmarks built around professional or domain-specific tasks, evaluating models in real-life scenarios is also highly important \cite{lin2025wildbench}.
Some benchmarks cover a wide range of settings, including long-term personal memory \cite{maharana-etal-2024-evaluating,MemoryBank,MEMORYLLM,backlund2025vendingbenchbenchmarklongtermcoherence,xu-etal-2022-beyond}, multi-agent or collaborative scenarios \cite{zhu-etal-2025-multiagentbench,Park2023GenerativeAgents,sun-etal-2025-collab,qian-etal-2024-chatdev}, multi-turn interactions with users and tools \cite{he2025vitabench,drouin2024workarena,qin2024toolllm,schick2023toolformer}, and everyday tasks such as web navigation or online shopping \cite{yao2022webshop,deng2023mindweb,zhou2024webarena,xie2024osworld}.

Most of these benchmarks still evaluate agents in an end-to-end manner and do not explicitly disentangle the underlying abilities of LMs \cite{yin-etal-2025-mmau,Evaluationagent}. 
In contrast, CL-bench Life does not directly evaluate the overall performance of an agent in a full real-world environment. 
Instead, it evaluates context learning ability and, through this, indirectly reflects whether an agent can handle complex context when solving real-world tasks. 
In this sense, CL-bench Life targets a more fundamental capability, enabling researchers to more clearly identify existing capability gaps and providing guidance for targeted model development.

\input{tables/statistics_table}

\section{CL-bench Life}

\label{sec:CL-bench Life}

In this section, we first provide an overview of CL-bench Life, then detail its context taxonomy and automatic evaluation method.

\subsection{Overview}
\label{sec:overview}

CL-bench Life aims to evaluate whether language models can learn from real-life context and apply what they learn to solve grounded tasks. 
The evaluation pipeline is illustrated in Figure~\ref{fig:cl-bench}. 
Given a context and a task, the model must generate a response based only on the provided information, and the solution is then automatically evaluated using task-level rubrics with a judge model.
The context is self-contained, and all new information needed to solve a task is already organized into the provided context, so LMs do not need to retrieve external information. 

Unlike existing context learning benchmarks that mainly focus on professional or domain-specific settings such as finance, science, and code \cite{dou2026cl,hu2026cl4sebenchmarkingcontextlearning}, CL-bench Life centers on common everyday-life contexts. 
These contexts are diverse, ranging from multi-party conversations, personal notes, and archives to community discussions, behavioral logs, game histories, app or system data, web and search activity, and other forms of digital footprints.
We present three cases with different types of context in Figure~\ref{fig:case3}.
Compared with professional-domain contexts, real-life contexts are often messier, more fragmented, and more personalized, which makes learning from it substantially more challenging.

The dataset statistics are reported in Table~\ref{tab:statistics}.
Each instance in CL-bench Life consists of a context, a task, and a set of task-level rubrics, all of which are manually curated. 
Moreover, about 59.8\% of instances in CL-bench Life are multi-turn. 
In these cases, the provided context also includes earlier rounds of interaction between the user and the assistant, while the final user turn specifies the target task to be solved. 
This increases task difficulty and more faithfully reflects how users interact with AI assistants in everyday scenarios.

\begin{figure}[htpb]
\centering
\includegraphics[width=1\textwidth]{figures/case.pdf}
\caption{
Three cases from CL-bench Life.
\textbf{Case 1:} A long, noisy group chat about planning a November “fancy” book club dinner. The task requires tracking changes in date and venue, reconciling conflicting member constraints, inferring final attendance from scattered updates, and judging whether the final plan satisfies the group’s stated preferences.
\textbf{Case 2:} Fragmented personal cycling archives, including trip plans, ride logs, weather disruptions, and maintenance notes. The task requires identifying trip-relevant evidence, inferring key failure modes, and producing a justified checklist for a future multi-day bike trip.
\textbf{Case 3:} Workout logs collected before and after an injury break. The task requires comparing performance over time, inferring the most affected muscle groups, distinguishing recovery from plateau, and suggesting training adjustments.
}
\vspace{-1em}
\label{fig:case3}
\end{figure}

\subsection{Context Taxonomy}
\label{sec:context-taxonomy}

\begin{wrapfigure}{r}{0.52\textwidth}
  \centering
  \vspace{-1.5em}
  \includegraphics[width=0.52\textwidth]{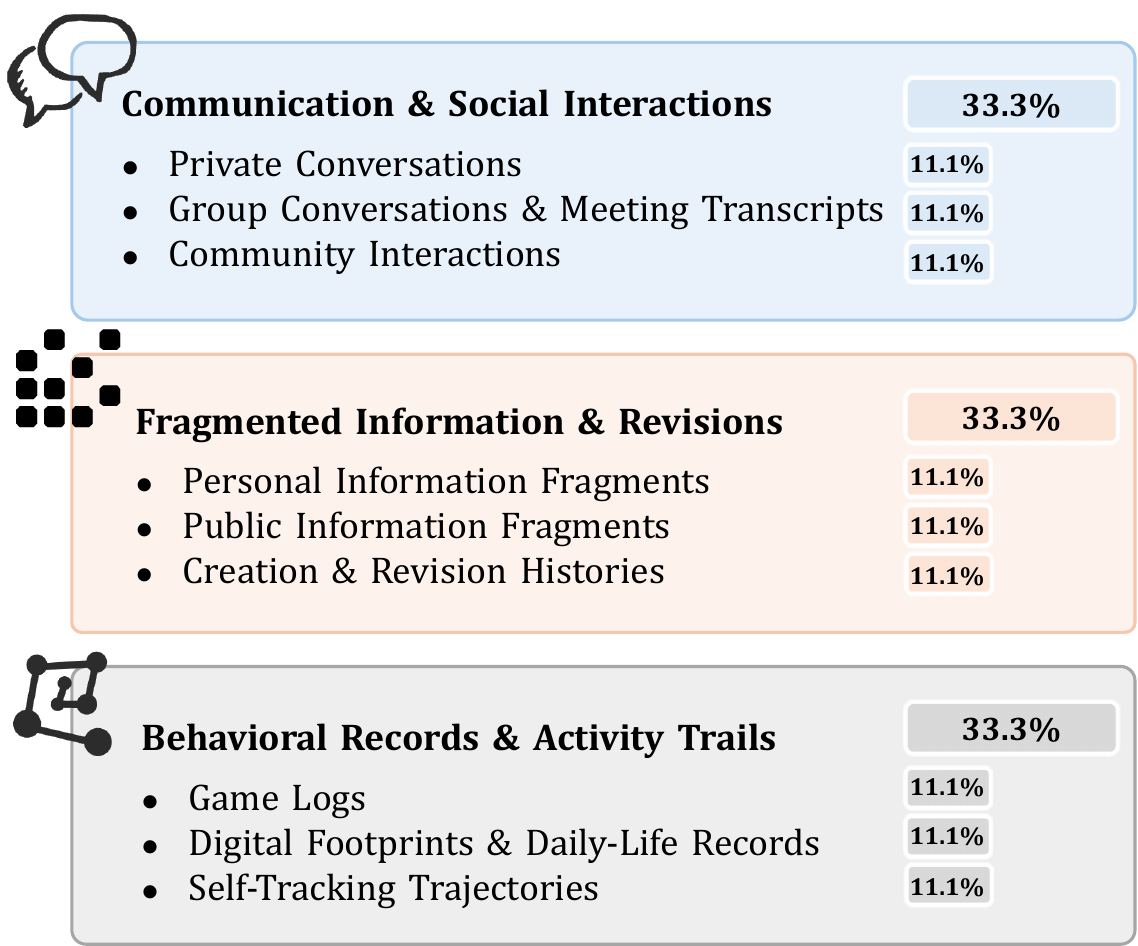} 
  \vspace{-1.5em}
  \caption{
Context taxonomy of CL-bench Life.
  }
\label{fig:taxonomy}
\end{wrapfigure}

We categorize the contexts in CL-bench Life into three broad types based on the kinds of context that people commonly encounter in everyday life, namely communication \& social interactions, fragmented information \& revisions, and behavioral records \& activity trails.
These three types correspond to interpersonal interactions, fragmented information that people record or revise, and behavioral traces generated through daily activity.
Each category is further divided into three subcategories, forming a taxonomy that captures the diversity of real-life contexts, as shown in Figure~\ref{fig:taxonomy}.
Each subcategory contains an equal number of contexts to avoid evaluation bias toward any single type.
To illustrate the distinctive characteristics of the three categories, we present one example case from each category in Figure~\ref{fig:case3}.

\textbf{Type 1: Communication \& social interactions.}
This type covers contexts arising from interpersonal communication and social interaction in everyday life. 
Such contexts come from one-on-one conversations, multi-party group chats, meeting transcripts, or public community discussions such as forum threads, blog comments, and social media interactions. 
Compared with clean and well-organized text, they are often more informal, socially grounded, and messy.
They often involve multiple participants, aliases or ambiguous references, topic shifts, interleaved threads, and nested replies.
Relevant information is often scattered across long interactions rather than presented in a single coherent passage, making the context fragmented and difficult to follow.

This type is divided into three subcategories:
\textbf{(1) private conversations}, which involve one-on-one interactions between individuals and often contain highly personal, informal, and topic-shifting exchanges;
\textbf{(2) group conversations \& meeting transcripts}, which contain multi-party discussions such as group chats, work meetings, or online voice-call transcripts;
and \textbf{(3) community interactions}, which cover public-facing discussions in forums, comment sections, blog discussions, or social media threads.

To solve tasks in this category, models must reason task-relevant information from messy conversational context, track topic changes and interleaved discussion threads, resolve aliases and references, and reason about the relationships, intentions, and viewpoints of different participants.

\textbf{Type 2: Fragmented information \& revisions.}
This type covers contexts characterized by scattered, incomplete, or evolving pieces of information. 
These contexts are often drawn from notes, snippets, lists, archives, reading histories, or revision records. 
Unlike structured documents with a clear narrative or logical sequence, they are often loosely organized and fragmented. 
Relevant information may be spread across multiple partial records, mixed with outdated or less relevant content, or embedded in iterative edits and revisions.

This type is divided into three subcategories:
\textbf{(1) personal information fragments}, which include personal notes, diaries, to-do lists, bookmarks, saved messages, and other private archives;
\textbf{(2) public information fragments}, which contain scattered public information such as RSS feeds, news snippets, blog reading histories, search results, or public records;
and \textbf{(3) creation \& revision histories}, which capture iterative content-production processes such as document edit histories, collaborative comments, or revision suggestions.

Tasks in this category require models to organize and reason over loosely structured information, integrate evidence from multiple fragments, and resolve inconsistencies or version changes across records. 
For revision-heavy contexts in particular, models must also trace how content evolves and understand the intent behind successive changes. 
Overall, this category evaluates whether models can learn effectively from fragmented, evolving, and weakly structured context.

\textbf{Type 3: Behavioral Records \& Activity Trails.}
This type covers contexts composed of passively accumulated behavioral and activity-based records that document actions, movements, and habits over time.
Rather than centering on dialogue or fragmented knowledge pieces, these contexts consist of raw activity traces, often recorded as logs, histories, timelines, or event sequences. 
They tend to be lengthy and highly structured, with task-relevant information often emerging from patterns across many records rather than from any single entry.

This type is divided into three subcategories:
\textbf{(1) game logs}, which contain behavioral records from games or other rule-based systems, such as MOBA match replays, poker hand histories, strategy execution logs, and event-aligned gameplay commentary or transcripts.
\textbf{(2) digital footprints \& daily-life records}, which include browsing histories, transaction records, location logs, and other digital traces of everyday activity;
and \textbf{(3) self-tracking trajectories}, which capture long-term self-monitoring data such as fitness logs, health records, learning progress, and other personal tracking histories.

Contexts in this category are also very common in real life. 
Tasks require models to reason over these real-life contexts, aggregate statistics, identify patterns, and provide useful insights. 
Depending on the subcategory, models may need to analyze strategy and game mechanics from logged events, infer routines and preferences from timestamped daily-life traces, or reason about trends, anomalies, and long-term changes in self-monitoring records.
Overall, this category evaluates whether models can learn effectively from lengthy, behavior-centered, and record-intensive context.

\subsection{Construction and automatic evaluation}
\label{sec:auto-verify}

\textbf{Simplified construction process.}
CL-bench Life is fully constructed by human annotators to ensure high data quality and to make the benchmark sufficiently challenging for frontier LMs. 
The simplified construction process for each instance involves the following steps\footnote{We provide a simplified description of the construction process due to policy restrictions.}. 
\textbf{(1)} We first defined and described common types of context that people encounter and deal with in everyday life through extensive discussion and investigation. 
\textbf{(2)} Given a specific context type, annotators then constructed a corresponding context by drawing from private sources, public sources, or newly created materials. 
As described in Section~\ref{sec:overview}, these contexts come from highly diverse sources.
Sensitive information is removed from all contexts to avoid infringing on the rights and interests of others.
\textbf{(3)} Annotators then designed sufficiently challenging tasks grounded in the context. 
These tasks are not intended to be simple needle-in-a-haystack retrieval problems or standard reading comprehension questions, nor can they be solved by relying only on pre-trained static knowledge. 
Instead, they evaluate models' real-life context learning ability by requiring them to reason over the provided context and use newly learned information.
\textbf{(4)} Finally, annotators write task-level rubrics for evaluation. 
These rubrics are required to be accurate, objective, and self-contained.
They avoid introducing requirements that are not grounded in the task itself and avoid relying on subjective judgment whenever possible.

Moreover, supervisors also conduct multiple rounds of sampled quality inspection and feedback to ensure annotation quality in the annotation process. 
On average, annotating each context and its corresponding tasks requires approximately 13 hours of expert effort.

\textbf{Automatic evaluation.}
Complex real-life tasks in CL-bench Life involve solutions that are difficult to check with predefined rules or may admit multiple valid solutions. 
To enable reliable automatic evaluation, we write task-level rubrics following prior work \cite{dou2026cl,dou2025evalearnquantifyinglearningcapability}. 
Specifically, each rubric is formulated as a binary question that only allows a ``yes'' or ``no'' answer, where ``yes'' indicates that the model's solution satisfies the rubric. 
On average, each task is associated with 13.2 rubrics.

\input{tables/main_table}

In all experiments, we use GPT-5.1 with high reasoning effort as the verifier to evaluate model-generated solutions based on the rubrics. 
We find that introducing the original context and task during judging tends to increase instruction-following failures of the verifier and may also introduce subjective bias into the assessment of model solutions. 
Therefore, the rubrics in CL-bench Life are decoupled from their corresponding contexts and tasks. 
For example, a rubric may take the form of ``The response should correctly list the total gold for all ten participants at the five-minute mark: Participant 1 (1,350), P2 (1,753), ...'' 
And during evaluation, we provide only the generated solution and the rubrics, without the original context or task. 
The system prompt used for the verifier follows \citet{dou2026cl}.

We adopt a strict evaluation criterion: a model is considered to have successfully solved a task only if its solution passes all associated rubrics.
We also analyze in Section~\ref{sec:eval-ana} whether model rankings change under judges with different levels of strictness.
To validate the reliability of this evaluation process, we randomly sample 100 model-generated solutions and their verification results and manually check them.
The results show that the evaluation accuracy exceeds 90\%. 
This observation is consistent with prior rubric-based evaluation work \cite{deshpande2025multichallenge,dou2025evalearnquantifyinglearningcapability}, suggesting that the framework is reliable and rigorous for our benchmark.
Moreover, in Table~\ref{tab:judge_consistency}, we further investigate agreement across different judge models.

\section{Benchmarking LMs with CL-bench Life}
\label{sec:experiments}

\textbf{Setup.}
We evaluate ten frontier language models on CL-bench Life through their official APIs: GPT-5.4 from OpenAI~\cite{openai2025gpt5}, Claude-Opus-4.6 from Anthropic~\cite{anthropic2026claudeopus46}, Gemini-3.1-Pro from Google~\cite{google2026gemini31pro}, Seed-2.0-Pro from ByteDance~\cite{bytedance2026seed20pro}, Kimi-K2.5-Thinking from Moonshot~\cite{moonshot2026k2_5}, Qwen-3.5-Plus from Alibaba~\cite{alibaba2026qwen35plus}, Grok-4.20 from xAI~\cite{xai2025grok4}, DeepSeek-V3.2-Thinking from DeepSeek~\cite{deepseek2025v32}, Hy3-preview from Tencent, and MiniMax-M2.5~\cite{minimax2026m25}. For models that support explicit reasoning, we evaluate them under high reasoning effort. For each model, we run three independent trials and report the mean $\pm$ standard deviation. A task is counted as correct only when the model satisfies every rubric item. We use GPT-5.1~\cite{openai2025gpt51} in the high reasoning effort as the default judge to ensure grading reliability.

\subsection{Main Results}

\textbf{Real-life context learning remains difficult for frontier LMs.} As shown in Table~\ref{tab:main_results}, even the best-performing model, GPT-5.4, achieves only a 19.3\% task solving rate. Most models cluster between 12\% and 17\%, with an average score of 13.8\%. Even when all information required to solve the task is already contained in the provided context, models still fail on the majority of tasks. 
These results indicate that learning from noisy, fragmented, and highly personal real-life context remains a major challenge for current AI systems.

\textbf{Different context categories pose different challenges for context understanding.} Averaged across the ten models, mean performance is 13.8\% overall, with 16.2\% on communication \& social interactions, 12.0\% on fragmented information \& revisions, and 13.1\% on behavioral records \& activity trails.
Communication \& social interactions is the easiest category on average, while fragmented information \& revisions is the most challenging, with the best model reaching only 16.5\%. 
Models also show category-specific strengths: GPT-5.4 performs best on communication \& social interactions and behavioral records \& activity trails, whereas Claude-Opus-4.6 leads on fragmented information \& revisions.

These differences likely reflect the distinct reasoning demands and data characteristics of each context type. 
Social interactions require tracking multi-party, interleaved discourse, such patterns are relatively common in web-scale conversational data. 
By contrast, fragmented notes and revisions require reconstructing scattered evidence across partially updated or overwritten records, a setting in which crucial information is often implicit and temporally distributed. 
Behavioral records further require more global reasoning over long event sequences and gradual changes, often without explicit discourse cues to highlight what matters.

\textbf{Frontier models exhibit distinct strengths across fine-grained context categories.} 
Table~\ref{tab:subcategory_breakdown} further exhibits model performance across all nine subcategories. 
Group conversations \& meeting transcripts and game logs are relatively tractable, with several top models reaching roughly 30\% or higher. 
By contrast, self-tracking trajectories is consistently the hardest subcategory: the best score is only 10.4\%, and most other models remain below 6\%. 
Figure~\ref{fig:case3} presents an example from this subcategory, where the model must reason over workout logs collected before and after an injury break to identify which muscle groups were affected and to distinguish genuine recovery from a prolonged plateau. 
This difficulty likely stems from the nature of self-tracking data, which is often raw, weakly structured, and only sparsely annotated compared with conversational or document-centric contexts. To solve such tasks, models must aggregate many small events over long time horizons and infer latent patterns that are rarely stated explicitly. 
These results suggest that different types of real-life context impose qualitatively different demands on context understanding and reasoning.

\input{tables/subcategory_table}

\begin{figure}[!b]
\centering
\includegraphics[width=1\textwidth]{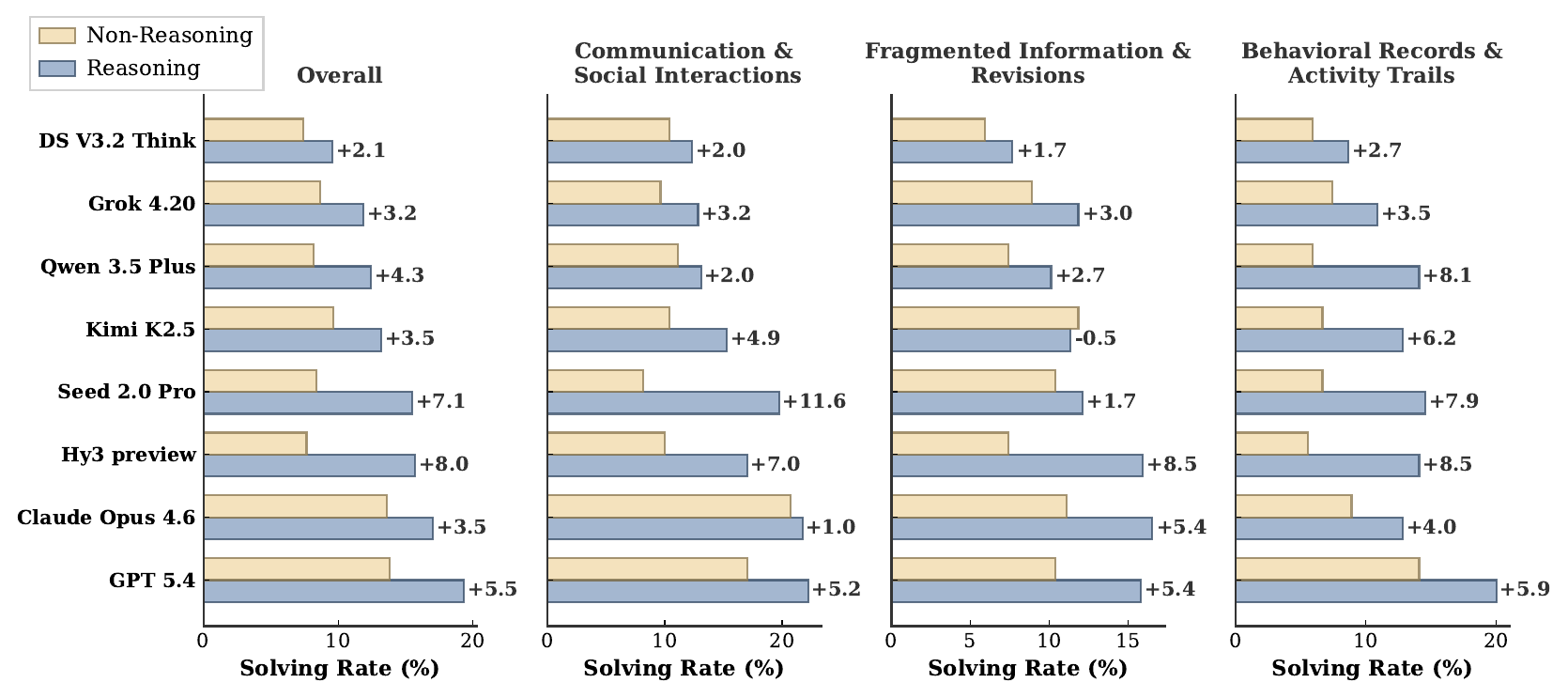}
\caption{
Performance comparison between reasoning and non-reasoning settings.
}
\label{fig:reasoning_comparison}
\end{figure}

\begin{figure}[t]
\centering
\includegraphics[width=0.9\textwidth]{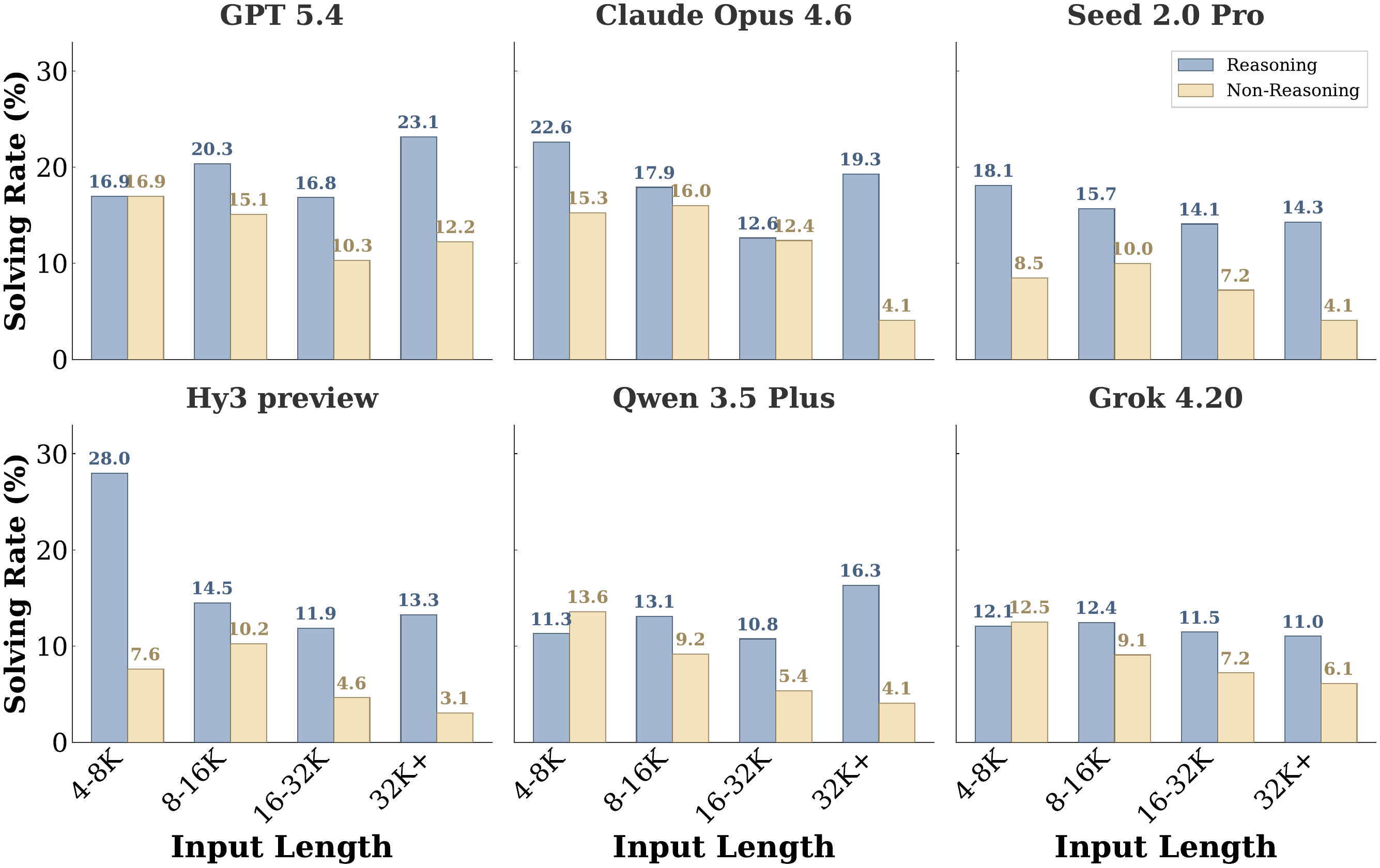}
\caption{
Task solving rate across context length bins under reasoning and non-reasoning settings.
}
\vspace{-1em}
\label{fig:token_length_bars}
\end{figure}

\begin{figure}[!b]
\centering
\begin{minipage}[t]{0.48\textwidth}
\centering
\includegraphics[width=\textwidth]{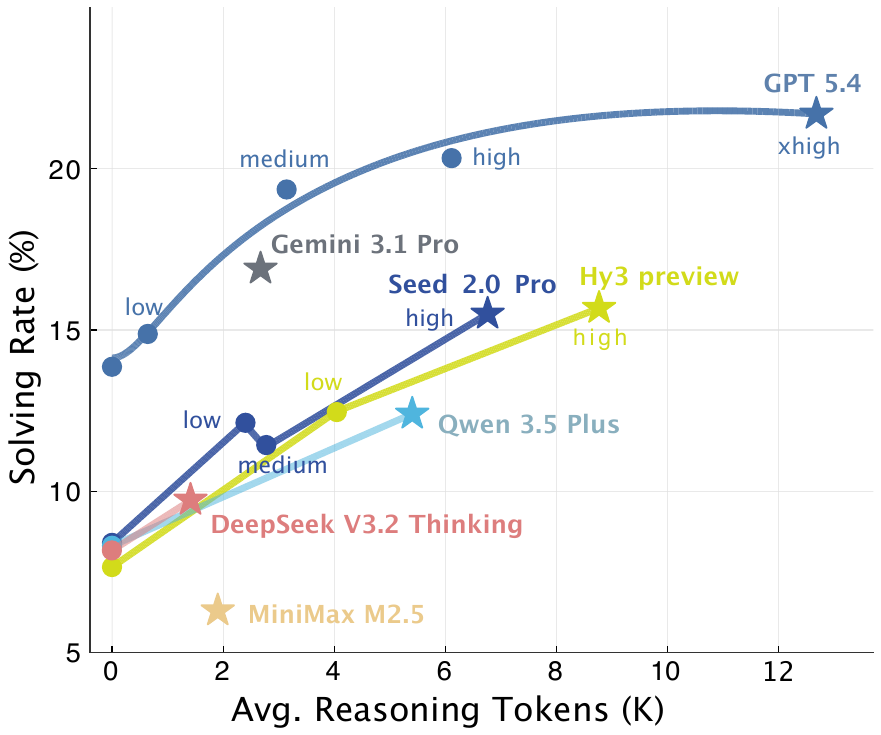}
\end{minipage}
\hfill
\begin{minipage}[t]{0.48\textwidth}
\centering
\includegraphics[width=\textwidth]{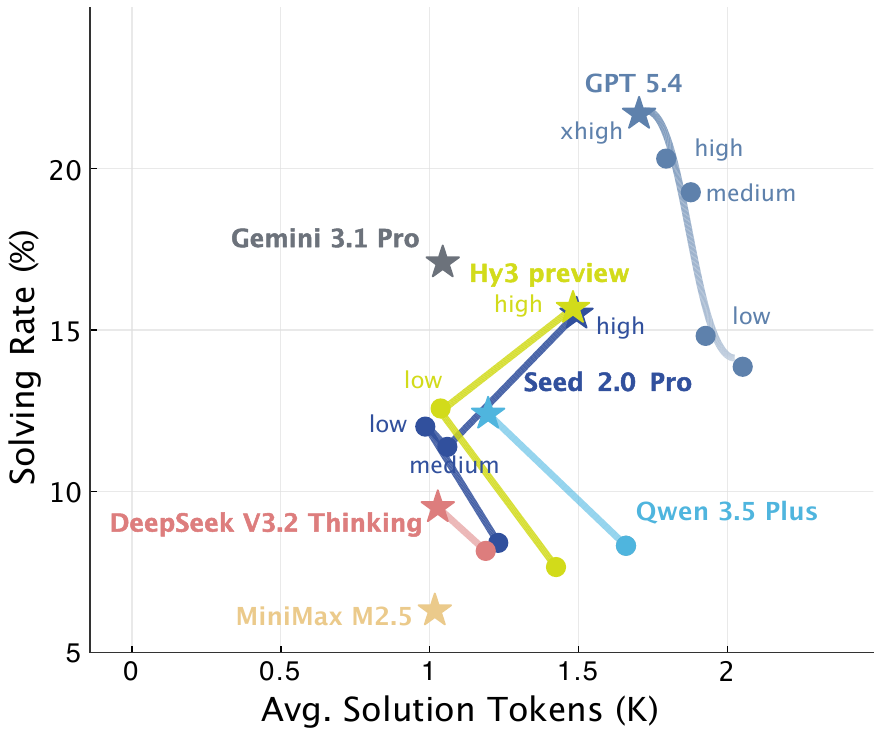}
\end{minipage}
\caption{
Average reasoning tokens \textbf{(left)} and solution tokens \textbf{(right)} versus task solving rate. 
Arrows connect reasoning and non-reasoning variants of the same model. 
More reasoning tokens generally correlate with a higher solving rate. 
Longer answers do not reliably predict a higher solving rate.
}
\label{fig:token_breakdown}
\end{figure}

\textbf{Real-life context learning benefits from explicit reasoning.} 
Figure~\ref{fig:reasoning_comparison} compares reasoning and non-reasoning settings for eight models that support both. 
Enabling reasoning improves performance for most models, suggesting that real-life context learning benefits from more deliberate, context-grounded inference.
The largest and most consistent gains appear on behavioral records \& activity trails, indicating that reasoning is especially helpful for reconstructing behavioral patterns from scattered evidence. 
Gains on fragmented information \& revisions and communication \& social interactions are more uneven. 
This suggests that reasoning helps models integrate dispersed evidence and track longer dependencies, but remains insufficient to address the broader challenge of real-life context learning.
Overall, the results are consistent with the design of CL-bench Life: these tasks require strong inference over realistic everyday context, and deeper reasoning improves performance, although the magnitude of the gain still depends on both the model and the context category.

\textbf{Poor performance on real-life context learning is not simply a long-context problem.} 
The input to a language model consists of the context and the task specification, with the context accounting for most of the total input length. 
Figure~\ref{fig:token_length_bars} shows task solving rate across context-length bins under reasoning and non-reasoning settings. 
Without reasoning, several models exhibit a decline in performance as context length increases, especially on the longest inputs. 
However, this pattern does not hold consistently when reasoning is enabled. 
GPT-5.4 achieves its highest task solving rate of 23.1\% on contexts above 32K, and Qwen-3.5-Plus peaks at 16.3\% in the longest bin. 
Claude-Opus-4.6 also recovers in the longest context bin after declining in the middle ranges. 
These results indicate that input length alone does not determine task difficulty. 
The key challenge lies in whether models can effectively organize and reason over noisy, fragmented, and weakly structured real-life context.

\input{tables/error_analysis}

\begin{wrapfigure}{r}{0.5\textwidth}
  \centering
  \vspace{-2em}
\includegraphics[width=0.5\columnwidth]{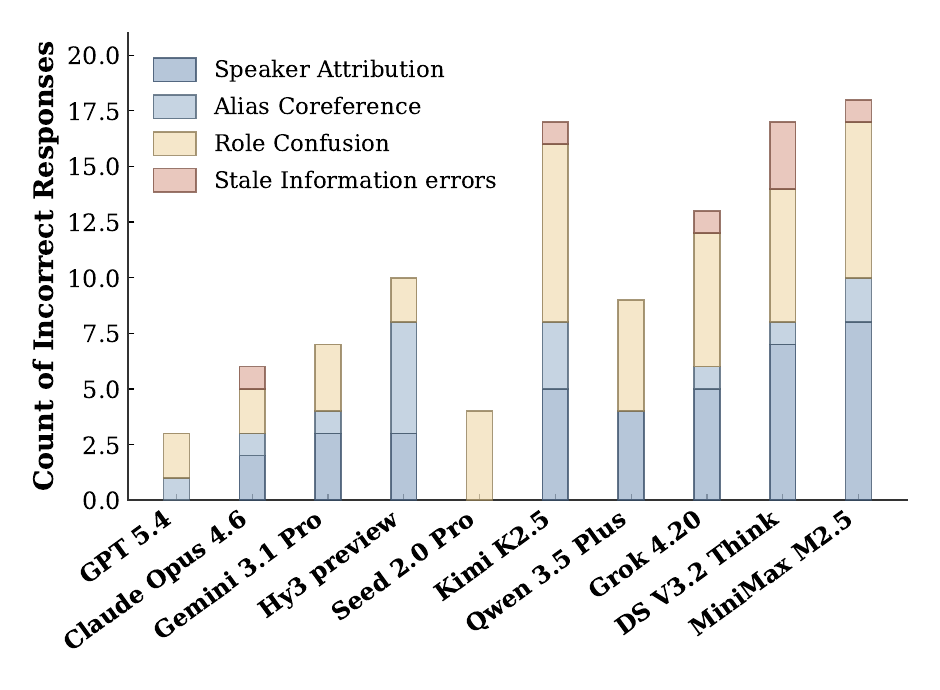}
 \vspace{-1.5em}
\caption{
Fine-grained breakdown of failures in group conversations \& meeting transcripts category.
}
\label{fig:interpersonal_errors}
 \vspace{-1em}
\end{wrapfigure}

\textbf{Reasoning quality matters more than reasoning volume.} Figure~\ref{fig:token_breakdown} relates task solving rate to output token usage, separating reasoning tokens \textbf{(left)} from answer tokens \textbf{(right)}. 
The arrows connecting reasoning and non-reasoning variants of the same model reveal substantial differences in \emph{token efficiency}: how much performance improves from additional reasoning tokens. 
For example, GPT-5.4 improves from roughly 15\% in the low setting to about 22\% in xhigh, but this gain comes with a large increase in reasoning tokens. 
Moreover, the marginal performance gain shrinks as reasoning effort increases, suggesting diminishing returns from simply allocating more reasoning tokens. 
Gemini-3.1-Pro reaches around 17\% with only about 2.7K reasoning tokens, and Seed-2.0-Pro reaches about 15\% with around 6.7K, showing that similar performance levels can arise from substantially different reasoning-token budgets.

At the same time, stronger reasoning does not necessarily produce longer final answers. 
For most model families, as reasoning increases, answers often become shorter while solving rate improves. 
Stronger reasoning may tend to make responses more selective, targeted, and accurate, rather than simply longer. 
This suggests that the value of reasoning lies not in generating more text, but in organizing context more effectively and converting it into more precise solutions. 
For practical AI assistants, the key takeaway is that better reasoning is not about saying more, but about understanding more and answering more directly.

\section{Further Analysis}
\label{sec:analysis}

\subsection{Error Analysis}

\textbf{Context misuse is the primary failure mode.}
Table~\ref{tab:error_types} presents the prevalence of different error patterns across models. 
We examine four non-mutually-exclusive failure dimensions. 
\textbf{Context-Ignored} refers to responses that leave relevant contextual information entirely unused, such that a necessary constraint or fact never appears in the answer. 
\textbf{Context-Misused} refers to cases where the model does engage with the context but fails to reason correctly over it, either by misinterpreting an individual piece of information or by failing to integrate multiple pieces into a coherent conclusion. 
\textbf{Format-Error} captures violations of output-format constraints regardless of whether the underlying content is correct, and \textbf{Refusal} denotes explicit task rejection or false claims of insufficient information.

Across models, failures are driven much more by misunderstanding or misusing contextual information than by simply ignoring it. 
Context-Misused is the most prevalent failure dimension, while Context-Ignored also remains substantial. 
A representative Context-Misused case appears in Table~\ref{table:category1_case1}, where the model reasons over the interaction history between the landlord and the tenant, but fails to properly weigh earlier and later evidence.
By comparison, Format-Error is much less common, and outright Refusal is rare. 
These results suggest that although stronger models may attend to context more reliably, the harder challenge remains correct contextual interpretation: models often read the context, yet still fail to use it in the right way.

\input{tables/no_context_ablation}

\textbf{Socially grounded group-chat settings expose a distinctive error pattern.} 
Group conversations and meeting transcripts provide a representative setting for studying social context understanding. Prior work has more often focused on long-context understanding or single-user assistant scenarios, while model behavior in messy multi-party social interactions remains relatively underexamined. In such group settings, success depends not only on recalling facts, but also on tracking participant identity, authority structure, responsibility, and information updates over time. We therefore conduct a focused error analysis on group conversations \& meeting transcripts tasks, where models must reason over multiple participants, their roles, and their evolving relationships throughout the interaction.

We identify four recurring error sources. 
\textbf{Speaker Attribution}: attributing statements to the wrong person, \textbf{Alias Coreference}: failing to recognize that different names refer to the same person, \textbf{Role Confusion}: misidentifying who holds authority or responsibility, and \textbf{Stale Information}: relying on outdated information that was later corrected.
As shown in Figure~\ref{fig:interpersonal_errors}, Role Confusion is the dominant error category.
We present a representative case of Role Confusion error in Table~\ref{table:category1_case2}, where Gemini-3.1-Pro fails: in a Slack channel where Alice, Brenda, and Clara collaboratively handle user questions about recipes and gardening, the model mistakes Alice, who created the channel and laid down the working rules, for the senior figure, and treats Clara, the one who actually makes the final calls, as her subordinate. With the interpersonal roles within the team thus inverted, the subsequent chain of reporting relationships is misread accordingly.
Speaker Attribution is also frequent, appearing in Hy3-preview and Gemini-3.1-Pro.
Alias Coreference and Stale Information errors are less common. 
Taken together, these patterns suggest that the core difficulty of social context understanding lies not just in tracking facts, but in maintaining role structure, speaker identity, and evolving participant-level distinctions throughout messy multi-party interactions.

\input{tables/judge_consistency}

\subsection{Evaluation Effectiveness Analysis}
\label{sec:eval-ana}

\begin{figure}[t]
\centering
\includegraphics[width=0.8\textwidth]{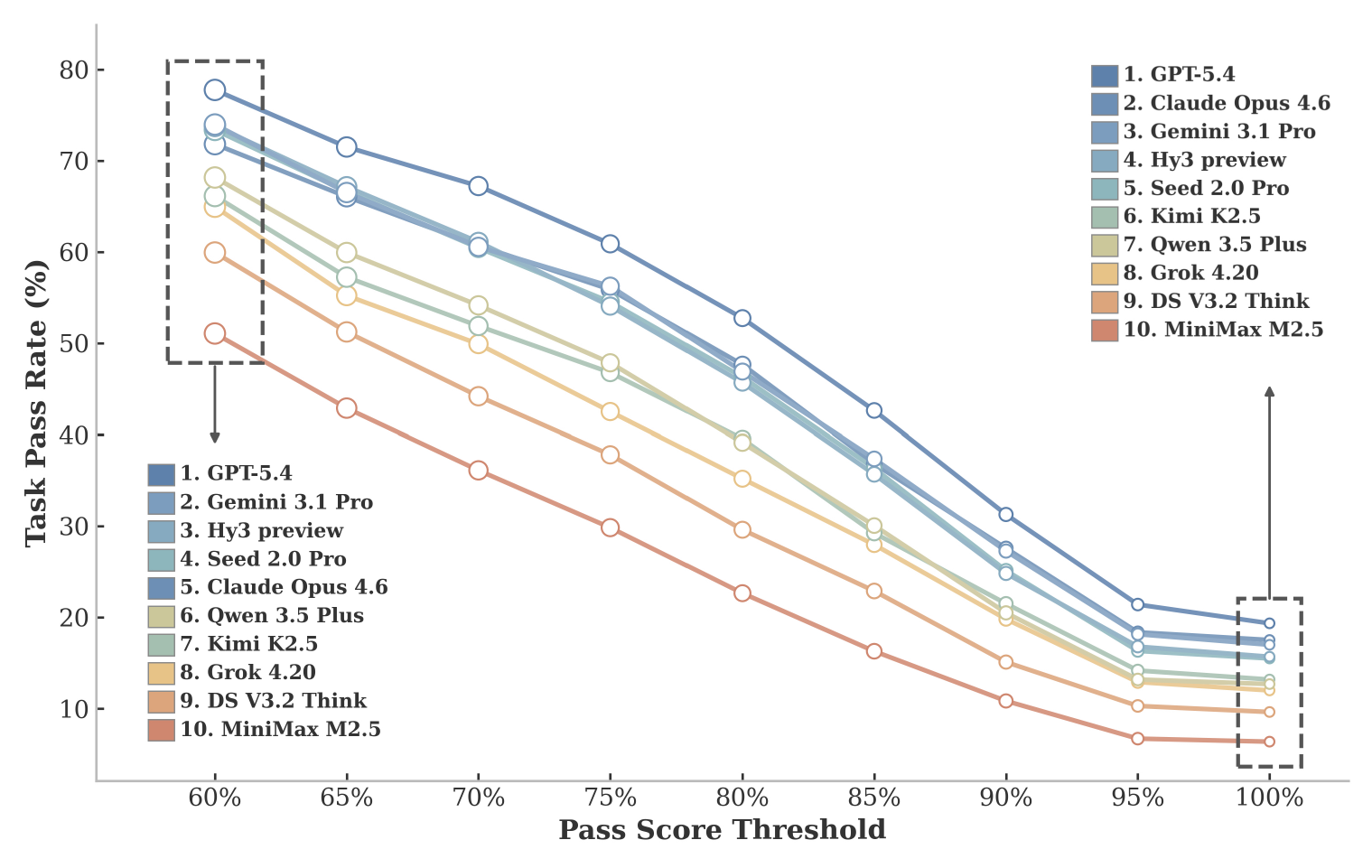}
\caption{
Task pass rate under different thresholds for all ten models. 
As the threshold increases, pass rates decrease monotonically for all models, showing that fully satisfying all rubrics is substantially harder than meeting only part of them. 
Stronger models remain consistently ahead across thresholds.
}
\label{fig:pass_score}
\end{figure}

\textbf{Solving CL-bench Life tasks requires models to learn from the provided context.} 
To verify that the benchmark cannot be solved from the final query alone, we conduct a no-context ablation in which all contexts are removed, and only the user’s final task is retained. As reported in Table~\ref{tab:no_context}, 
GPT-5.4 drops from 19.3\% to 1.7\% when context is stripped away. 
The degradation is consistent across all categories. These near-zero results show that CL-bench Life cannot be solved through parametric world knowledge alone, and instead requires models to acquire and use information from the provided real-life context.

\begin{figure}[!b]
\centering
\includegraphics[width=0.9\textwidth]{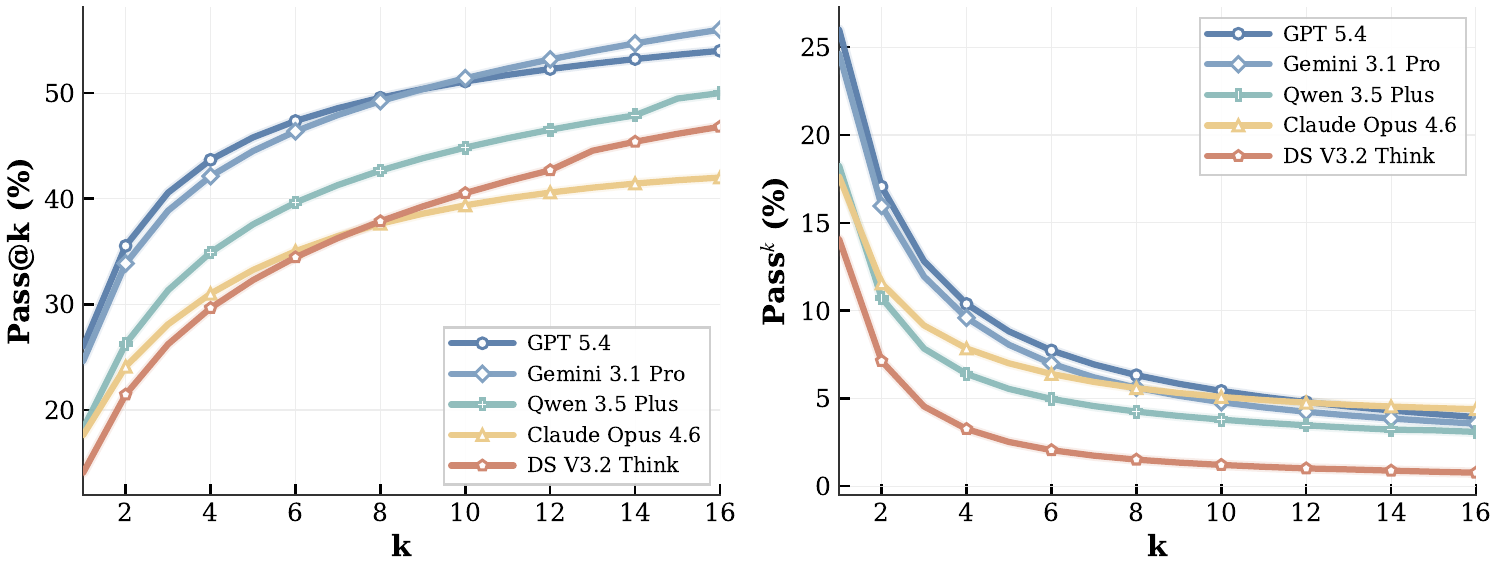}
\caption{
Model performance under Pass@k and Pass\textasciicircum k.
}
\label{fig:passk}
\end{figure}

\textbf{Fully correct solutions are relatively rare, but partially correct ones are much more common, while model rankings remain broadly stable across thresholds.}
We further analyze performance under varying rubric pass-rate thresholds rather than relying solely on strict binary pass/fail evaluation. 
In the default setting, a response is counted as correct only if it satisfies all rubrics for a task. 
Here, we relax that requirement by counting a task as correct whenever the proportion of satisfied rubrics exceeds a chosen threshold. 
Figure~\ref{fig:pass_score} shows that pass rates decrease for all models as the threshold increases from 60\% to 100\%, indicating that many responses satisfy part of the evaluation criteria under lenient settings, whereas fully satisfying all rubrics remains substantially more difficult. For example, GPT-5.4 drops from about 77\% at the 60\% threshold to 19\% at the 100\% threshold.
At the same time, model rankings remain broadly stable across thresholds, with stronger models maintaining an advantage under both lenient and strict criteria.
These results suggest that CL-bench Life can distinguish partially solved cases from fully correct ones while still providing stable and discriminative model comparisons.

\textbf{Rubric-based evaluation is consistent across different judge models.}
To test the stability of our rubric-based evaluation, we re-grade outputs from five evaluated models using two alternative judge models, Claude-Opus-4.6 and Gemini-3.1-Pro. Their ratings are then compared against those of the default judge, GPT-5.1, with high reasoning effort. We report average pairwise agreement rates and Cohen's $\kappa$ in Table~\ref{tab:judge_consistency}. 
Pairwise agreement exceeds 93\% for all judge pairs, and Cohen's $\kappa$ ranges from 0.710 to 0.773, indicating substantial agreement. These results suggest that our rubric-based evaluation remains stable across different judge models.

\textbf{Inference-time scaling helps only to a limited extent.}
To assess inference-time scaling, we evaluate five representative models using Pass@k and pass\textasciicircum k, as shown in Figure~\ref{fig:passk}. 
Following prior work \cite{yao2024tau, chen2021evaluating}, Pass@k measures whether at least one of $k$ independent runs succeeds, while Pass\textasciicircum k measures whether all $k$ runs succeed.
To reduce evaluation cost, we randomly sample a 50-example subset from CL-bench Life and report results on this subset.

Results show that Pass@k increases with $k$ for all models, but the gains slow substantially after $k=8$.
This suggests that repeated sampling helps only on a limited set of borderline cases, while most failures reflect tasks that are intrinsically difficult for current models.
Pass\textasciicircum k provides a complementary view of robustness.
It declines sharply as $k$ increases for every model, showing that correct answers are often not achieved consistently across runs.
Even GPT-5.4 exhibits a substantial drop by $k=4$, indicating that many tasks are solved only occasionally.
Taken together, the rapid saturation of Pass@k and the steep decay of Pass\textasciicircum k suggest that the main bottleneck is not insufficient inference-time exploration, but the fundamental difficulty of real-life context learning itself.

\begin{figure}[thpb]
\centering
\includegraphics[width=1\textwidth]{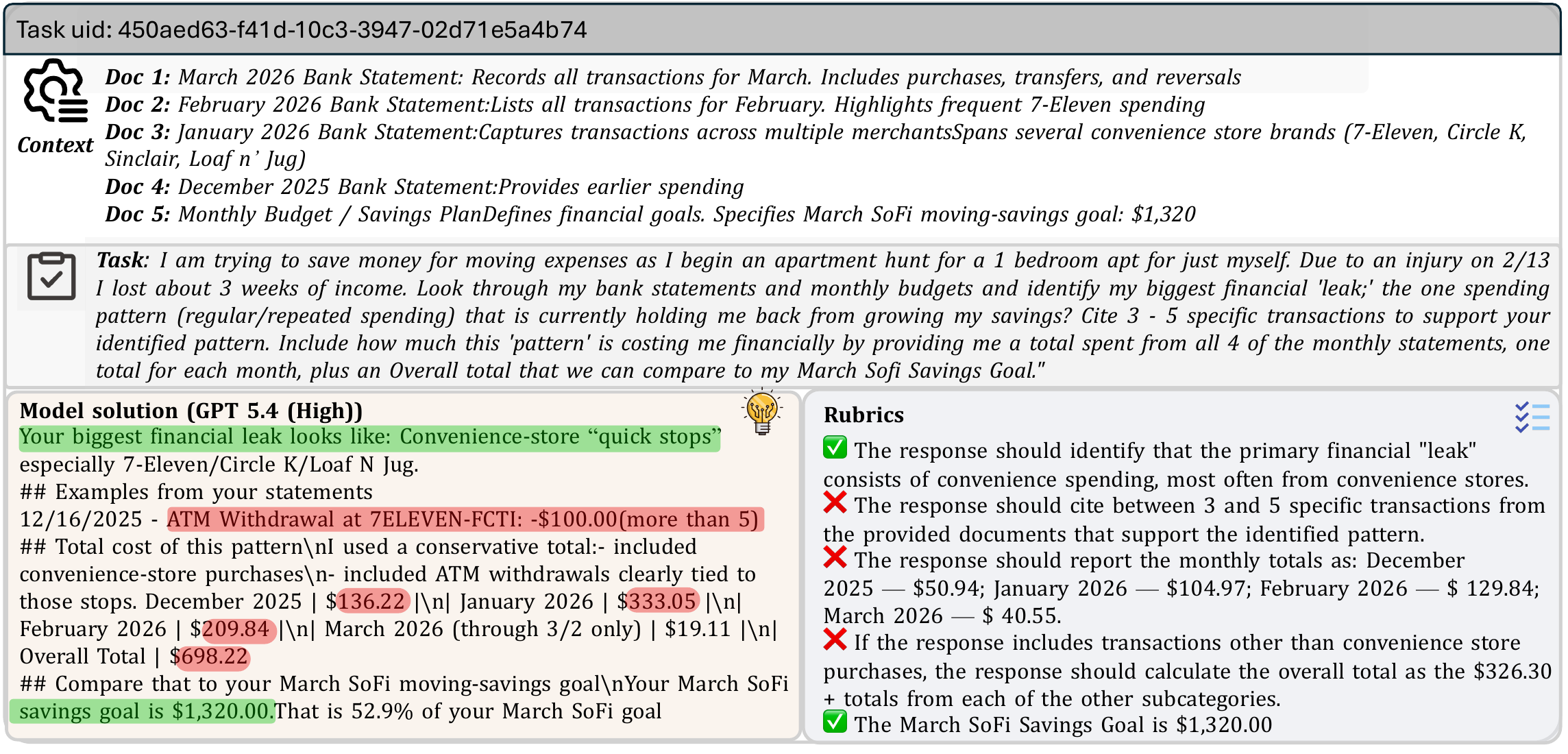}
\caption{
An example from CL-bench Life where the task requires identifying the largest recurring financial leak from four months of bank statements. 
While GPT-5.4 correctly identifies convenience store spending as the main leak, it fails to compute accurate monthly totals, misclassifies ATM withdrawals, and introduces unsupported speculative claims.
}
\label{fig:case}
\vspace{-1em}
\end{figure}

\subsection{Qualitative Analysis}
\label{sec:Qualitative-Analysis}

We select nine cases across multiple real-life context categories to gain deeper insights into model performance on context learning. 
These cases are drawn from GPT-5.4 (High), Claude-Opus-4.6 (High), Gemini-3.1-Pro (High) and Seed-2.0-Pro. 
We provide an in-depth analysis of remaining examples in Appendix~\ref{sec:in-depth analysis}.

In figure~\ref{fig:case}, we present an error case from behavioral records \&
activity trails category.
In this example, a user is planning to move, but has experienced income loss over the past three months due to an injury. She asks GPT-5.4 to identify the largest financial leak over the past four months, which means the recurring spending pattern that most significantly hinders her savings growth. She also provides more specific instructions: the model should (1) cite 3–5 concrete transactions as supporting evidence, (2) compute the monthly totals of this spending pattern across the four months as well as the overall total, and (3) compare this total against her March SoFi savings goal.

We then provide the user’s complete bank statements covering all income and expenses over the past four months. 
GPT-5.4 (High) successfully avoids being distracted by other spending categories (e.g., gambling-related transactions, subscription charges, etc.), and correctly identifies convenience store spending as the primary financial leak. It also demonstrates strong contextual understanding by locating the user’s March SoFi savings goal and incorporating it into a comparative analysis.

However, the model makes critical numerical and reasoning errors. Specifically, the monthly totals for convenience store spending are all incorrect. A key source of error is that the model incorrectly classifies ATM withdrawals at convenience stores as part of convenience store spending. The rubrics already consider this possible strategy and provide the correct results for them, yet the model still computes incorrect totals even under those same assumptions.
Furthermore, to mitigate hallucination, the user explicitly requires that all conclusions must be grounded in the provided documents, without speculative reasoning. Nevertheless, the model introduces unsupported claims such as “likely under-budgeted or invisible” and “a strong sign this habit is draining money,” which go beyond the evidence available in the bank statements.

Across all analyzed cases, we observe four consistent limitations of current frontier models. First, models struggle with evidence balancing and uncertainty calibration, often over-committing to a single narrative. Second, they fail at latent structure inference, relying on surface cues rather than recovering underlying relationships. Third, instruction following and constraint adherence remain unreliable, with frequent inclusion of irrelevant information or violation of task requirements. Finally, models exhibit weaknesses in fine-grained reasoning and grounding, particularly in aligning details and reconstructing implicit logic.
Overall, these results indicate that learning from noisy, fragmented, and dynamic real-life context remains a major challenge for current frontier models.

\section{Discussion}
\label{sec:discussion}

\textbf{Limitations and future work.}
CL-bench Life has several limitations that point to important directions for future work. 
First, group chat emerges as a particularly important yet challenging real-life setting. 
Models often struggle with phenomena such as alias and reference confusion in long multi-user interaction histories, suggesting the need for more targeted research on understanding and reasoning over group chat context.
Second, CL-bench Life currently focuses on text-only context. 
However, real-life context learning is inherently multimodal, and extending the benchmark to incorporate visual context is an important next step.
Thirdly, our rubric-based evaluation currently relies on GPT-5.1 as the judge model, which is effective but costly.
Meanwhile, we observe that smaller or open-source judge models exhibit noticeable bias. Designing low-cost and reliable rubric-based evaluation frameworks remains an open challenge.
Finally, we do not provide details of the construction process of CL-bench Life due to policy restrictions.

\textbf{Conclusion.}
We present CL-bench Life, the first benchmark for evaluating context learning in everyday real-life scenarios. 
CL-bench Life is fully human-annotated and covers common real-life settings. 
The contexts in it are sufficiently messy and complex to pose a meaningful challenge to current SOTA language models.
Our evaluations show that real-life context learning remains a major challenge for frontier LMs, with substantial room for improvement. 
In addition to overall performance gaps, we uncover several important findings about the relationships between performance and reasoning effort, solution length, context length, and long-context ability. 
We hope CL-bench Life will provide a strong foundation for future research on real-life context learning and help drive the development of more intelligent and practical AI assistants for everyday use.

\section*{Acknowledgments}

We greatly appreciate the substantial help from Speed Zhu, Clarence Ai, Deliang An, Ningxuan Wang, Xiaotong Yang, Yuhong Liu, and Lilin Wang (all at Tencent) on this project.

\clearpage

\section*{Full Author List}

Shihan Dou$^{*}$, Yujiong Shen$^{*}$, Chenhao Huang$^{*}$, Junjie Ye$^{*}$, Jiayi Chen, Junzhe Wang, Qianyu He, Shichun Liu, Changze Lv, Jiahang Lin, Jiazheng Zhang, Ming Zhang, Shaofan Liu, Tao Ji, Zhangyue Yin, Cheng Zhang, Huaibing Xie, Jianglu Hu, Jingcheng Deng, Lincheng Li, Minda Hu, Shaolei Wang, Syrus Zhao, Weichao Wang, Yan Lei, Yang Liu, Yanling Xiao, Yiting Liu, Zenan Xu, Zhen Guo, Ziliang Zhao

Pluto Zhou$^{\dagger}$, Tao Gui$^{\dagger}$, Qi Zhang, Xuanjing Huang, Yu-Gang Jiang, Di Wang, Shunyu Yao

\begingroup
\renewcommand\thefootnote{}
\footnote{$^*$Equal contribution. $^{\dagger}$Corresponding authors.}
\endgroup

\newpage
\bibliography{CLBenchreal}
\bibliographystyle{iclr2026_conference}

\clearpage

\appendix

\section{In-depth Analysis of Context Learning Successes and Failures}
\label{sec:in-depth analysis}
In this section, we present additional qualitative case studies following the style of Section~\ref{sec:Qualitative-Analysis}. Combined with the example discussed in the main paper, we present three cases for each of the following categories: Communication \& Social Interactions, Fragmented Information \& Revisions, and Behavioral Records \& Activity Trails. These cases further illustrate the bottlenecks encountered by frontier models in the real world. 

Tables~\ref{table:category1_case1} and \ref{table:category1_case8} illustrate failures in evidence balancing and uncertainty calibration. In Table~\ref{table:category1_case1}, Gemini-3.1-Pro (High) over-weights negative signals in a long-term landlord–tenant interaction and fails to incorporate subsequent positive evidence. Similarly, Table~\ref{table:category1_case8} shows that when presented with multiple plausible explanations for a user’s purchasing behavior, Seed 2.0 Pro (High) constructs a coherent hypothesis but over-commits to it, failing to preserve the ambiguity required by the evidence.

Tables~\ref{table:category1_case2} and \ref{table:category1_case3} highlight limitations in latent structure inference. In Table~\ref{table:category1_case2}, Gemini-3.1-Pro (High) misidentifies the underlying authority structure, conflating who initiates actions with who holds decision-making power. In Table~\ref{table:category1_case3},  GPT-5.4 (High) extracts surface-level disagreements from community discussions but fails to capture the deeper axes of contention and implicit norms governing the discourse.

Tables~\ref{table:category1_case4} and \ref{table:category1_case5} demonstrate weaknesses in instruction following and constraint adherence. In Table~\ref{table:category1_case4}, GPT-5.4 (High) includes non-actionable and irrelevant information when generating a planning checklist, indicating difficulty in distinguishing useful items from background reference data. In Table ~\ref{table:category1_case5}, despite producing a generally reasonable analysis, Claude-Opus-4.6 (High) violates strict citation constraints and misclassifies sources.

Finally, Tables~\ref{table:category1_case6} and \ref{table:category1_case7} reveal limitations in fine-grained reasoning and grounding. In Table~\ref{table:category1_case6}, Claude-Opus-4.6 (High) struggles to distinguish between true revisions and incremental elaborations, and fails to recover implicitly conveyed motivations. In Table~ \ref{table:category1_case7}, while Seed 2.0 Pro (High) identifies key events and performs necessary calculations, it does not precisely align natural language descriptions with formal event representations, nor does it fully establish the required causal chain.

These qualitative analyses suggest that the challenges faced by frontier models are not limited to traditionally difficult domains such as mathematical or formal reasoning. In contrast, real-world scenarios often pose even greater difficulty. These settings involve fragmented, noisy, and often implicitly expressed information, requiring models to infer latent relationships, hidden structures, and unstated assumptions from context. As a result, models frequently rely on surface-level cues or salient patterns, while failing to recover deeper, underlying signals necessary for correct reasoning.

\input{cases/case1}
\input{cases/case2}
\input{cases/case3}
\input{cases/case4}
\input{cases/case5}
\input{cases/case6}
\input{cases/case7}
\input{cases/case8}

\end{document}

%% file: mypackage.tex
\usepackage{natbib}
\usepackage{graphicx}
\usepackage{amsmath}
\usepackage{makecell}
\setcitestyle{numbers,square}
\usepackage[utf8]{inputenc} % allow utf-8 input
\usepackage[T1]{fontenc}    % use 8-bit T1 fonts
\usepackage{hyperref}       % hyperlinks
\usepackage{url}            % simple URL typesetting
\usepackage{booktabs}       % professional-quality tables
\usepackage{amsfonts}       % blackboard math symbols
\usepackage{nicefrac}       % compact symbols for 1/2, etc.

\usepackage{microtype}      % microtypography
\usepackage{xcolor}         % colors

\usepackage[most]{tcolorbox}
\usepackage{colortbl}
\usepackage{multirow}
\usepackage{graphicx}
\usepackage{lipsum}
\usepackage{tablefootnote}
\usepackage{tcolorbox}
\usepackage[tikz]{bclogo}
\usepackage{subfigure}
\usepackage{subcaption}
\usepackage{multirow}
\usepackage{wrapfig}
\usepackage{xcolor}
\usepackage{listings} % 代码风格展示
\usepackage{CJKutf8}

\usepackage{tabularx}
\usepackage{fancyvrb}
\usepackage{fvextra}

\usepackage{booktabs}
\usepackage{siunitx}

\newcommand{\Romannum}[1]{\texorpdfstring{\uppercase\expandafter{\romannumeral #1}}{#1}}

\newtcolorbox{titleEnv}{
colframe=black!80,
colback=gray!10,
fonttitle=\bfseries,
coltitle=black,
left=3pt,
right=3pt,
top=3pt,
bottom=3pt,
boxrule=0.4mm,
arc=3mm
}
\definecolor{mydeepblue}{RGB}{46, 90, 168}
\definecolor{myblue}{RGB}{166, 202, 236}
\definecolor{my_blue}{RGB}{0,120,255}
\definecolor{my_purple}{RGB}{161, 27, 155}
\definecolor{my_green}{RGB}{0, 176, 80}
\definecolor{msftBlue}{RGB}{0,164,239}
\definecolor{msftGreen}{RGB}{127,186,0}
\definecolor{msftYello}{RGB}{255,185,0}
\definecolor{msftBlack}{RGB}{0,0,0}

%% file: tables/statistics_table.tex
\begin{table}[thpb]
\caption{
Statistics of CL-bench Life, including counts of tasks and rubrics, average user turns, rubrics per task, and context length in tokens.
}
\centering
\resizebox{0.94\columnwidth}{!}{%
\setlength{\tabcolsep}{5pt}
    \begin{tabular}{lccccccccc}
    \toprule
    \multicolumn{1}{c}{\multirow{2}[4]{*}{\textbf{Context Category}}} & \multicolumn{1}{c}{\multirow{2}[4]{*}{\textbf{\# Contexts}}} & \multicolumn{1}{c}{\multirow{2}[4]{*}{\textbf{\# Rubrics}}} & \multicolumn{1}{c}{\multirow{2}[4]{*}{\makecell{\textbf{Avg.}\\\textbf{User Turns}}}} & \multicolumn{3}{c}{\textbf{\# Rubrics per Task}} & \multicolumn{3}{c}{\textbf{Context Length (tokens)}} \\
\cmidrule{5-10}          &       &       &       & \textbf{Min} & \textbf{Mean} & \textbf{Max} & \textbf{Min} & \textbf{Mean} & \textbf{Max} \\
    \midrule
    \makecell[l]{Communication \&\\Social Interactions} & 135   & 1,812 & 1.9   & 2     & 13.4  & 36    & 5.6K  & 12.9K & 34.9K \\
    \makecell[l]{Fragmented Information \&\\Revisions} & 135   & 1,877 & 2.0   & 3     & 13.9  & 38    & 5.4K  & 12.8K & 34.4K \\
    \makecell[l]{Behavioral Records \&\\Activity Trails} & 135   & 1,659 & 1.8   & 2     & 12.3  & 40    & 6.8K  & 32.5K & 170.8K \\
    \midrule
    Total & 405   & 5,348 & 1.9   & 2     & 13.2  & 40    & 5.4K  & 19.4K & 170.8K \\
    \bottomrule
    \end{tabular}%
    }
    \vspace{-1em}
\label{tab:statistics}
\end{table}

%% file: tables/main_table.tex
\newcommand{\gradientcellstd}[2]{%
    \begin{tikzpicture}[baseline]
        \fill[gray!25] (0,0) rectangle (0.9,0.3);
        \fill[gray!100] (0,0) rectangle ({0.9*#1/100},0.3);
        \node[anchor=west, font=\scriptsize] at (0.95,0.15) {
            \textbf{\makebox[1.2em][r]{#1}}\,{\tiny$\pm$}\,\makebox[0.8em][l]{#2}
        };
    \end{tikzpicture}%
}
\newcommand{\gradientcellsingle}[1]{%
    \begin{tikzpicture}[baseline]
        \fill[gray!25] (0,0) rectangle (0.9,0.3);
        \fill[gray!100] (0,0) rectangle ({0.9*#1/100},0.3);
        \node[anchor=west, font=\scriptsize] at (0.95,0.15) {
            \textbf{\makebox[1.2em][r]{#1}}
        };
    \end{tikzpicture}%
}

\begin{table*}[t]
    \caption{
    Task solving rate of ten frontier LMs on CL-bench Life. 
    All models are evaluated in reasoning mode, with results reported as mean $\pm$ std (\%) across three runs.
    }
    \centering
    \small
    \scalebox{0.84}{
    \begin{tabular}{@{}l|>{\centering\arraybackslash}p{2cm}|>{\centering\arraybackslash}p{2.4cm}|>{\centering\arraybackslash}p{2.8cm}|>{\centering\arraybackslash}p{2.4cm}@{}}
    \toprule
    \multicolumn{1}{l|}{\textbf{Model Names}} &
    \multicolumn{1}{c|}{\textbf{Overall}} &
    \multicolumn{1}{c|}{\makecell{\textbf{Communication \&} \\ \textbf{Social Interactions}}} &
    \multicolumn{1}{c|}{\makecell{\textbf{Fragmented} \\ \textbf{Information \& Revisions}}} &
    \multicolumn{1}{c}{\makecell{\textbf{Behavioral Records \&} \\ \textbf{Activity Trails}}} \\
    \midrule

    GPT 5.4 (High) & \gradientcellstd{19.3}{0.5} & \gradientcellstd{22.2}{0.7} & \gradientcellstd{15.8}{0.9} & \gradientcellstd{20.0}{0.7} \\

    Claude Opus 4.6 (High) & \gradientcellstd{17.0}{1.3} & \gradientcellstd{21.7}{3.3} & \gradientcellstd{16.5}{0.9} & \gradientcellstd{12.8}{1.5} \\

    Gemini 3.1 Pro (High) & \gradientcellstd{16.9}{1.2} & \gradientcellstd{18.8}{1.7} & \gradientcellstd{13.6}{1.9} & \gradientcellstd{18.3}{1.5} \\

     Hy3 preview (High) & \gradientcellstd{15.7}{0.9} & \gradientcellstd{17.0}{2.1} & \gradientcellstd{15.9}{1.6} & \gradientcellstd{14.1}{2.1} \\

    Seed 2.0 Pro (High) & \gradientcellstd{15.5}{1.7} & \gradientcellstd{19.8}{2.3} & \gradientcellstd{12.1}{2.4} & \gradientcellstd{14.6}{1.1} \\

    Kimi K2.5 (High) & \gradientcellstd{13.2}{0.8} & \gradientcellstd{15.3}{1.1} & \gradientcellstd{11.4}{1.5} & \gradientcellstd{12.8}{0.4} \\

    Qwen 3.5 Plus (High) & \gradientcellstd{12.4}{0.1} & \gradientcellstd{13.1}{0.9} & \gradientcellstd{10.1}{1.1} & \gradientcellstd{14.1}{0.7} \\

    Grok 4.20 & \gradientcellstd{11.9}{0.4} & \gradientcellstd{12.8}{0.9} & \gradientcellstd{11.9}{1.5} & \gradientcellstd{10.9}{2.6} \\

    DeepSeek V3.2 Thinking & \gradientcellstd{9.5}{0.4} & \gradientcellstd{12.3}{1.1} & \gradientcellstd{7.7}{1.1} & \gradientcellstd{8.6}{1.7} \\

    MiniMax M2.5 & \gradientcellstd{6.3}{1.0} & \gradientcellstd{8.6}{2.1} & \gradientcellstd{5.2}{1.3} & \gradientcellstd{5.2}{1.3} \\

    \bottomrule
    \end{tabular}
     }
     \vspace{-1em}
    \label{tab:main_results}
\end{table*}

%% file: tables/subcategory_table.tex
\definecolor{mygreen}{HTML}{DCE6F2}
\definecolor{myorange}{HTML}{FBE3C6}
\definecolor{mypink}{HTML}{F6D0D0}

\newcommand{\capval}[1]{\ifnum#1>100 100\else#1\fi}
\newcommand{\boost}[1]{\number\numexpr (#1*11)/10 \relax}

\newcommand{\orcell}[2]{\cellcolor{myorange!\capval{\boost{#1}}}#2}
\newcommand{\grcell}[2]{\cellcolor{mygreen!\capval{\boost{#1}}}#2}
\newcommand{\pkcell}[2]{\cellcolor{mypink!\capval{\boost{#1}}}#2}

\begin{table*}[t]
\caption{
Task solving rates across nine subcategories for all models. Columns are grouped by category. Cell shading indicates relative accuracy within each subcategory column, with darker cells indicating stronger performance.
}
\centering

\normalsize
\setlength{\tabcolsep}{5pt}   
\renewcommand{\arraystretch}{1.2}  

\resizebox{\textwidth}{!}{%
\begin{tabular}{@{}l*{9}{>{\centering\arraybackslash}p{2.0cm}}@{}}
\toprule
\multirow{2}{*}{\textbf{Model}} 
& \multicolumn{3}{c}{\textbf{\makecell{Communication \&\\Social Interactions}}} 
& \multicolumn{3}{c}{\textbf{\makecell{Fragmented Information\&\\Revisions}}} 
& \multicolumn{3}{c}{\textbf{\makecell{Behavioral Records\&\\Activity Trails}}} \\

\cmidrule(lr){2-4} \cmidrule(lr){5-7} \cmidrule(lr){8-10}

 & \textbf{\makecell{Comm.\\Inter.}} & \textbf{\makecell{Group\\Conv.}} & \textbf{\makecell{Private\\Conv.}} & \textbf{\makecell{Pers. Info.\\Frag.}} & \textbf{\makecell{Pub. Info.\\Frag.}} & \textbf{\makecell{Create \&\\ Rev.}} & \textbf{\makecell{Game\\Logs}} & \textbf{\makecell{Digital FP\\\& Daily}} & \textbf{\makecell{Self-Trk.\\Traj.}} \\
    
\midrule

GPT 5.4 (High)
& \grcell{60}{20.7} & \grcell{65}{30.4} & \grcell{40}{15.6} 
& \orcell{45}{14.8} & \orcell{65}{20.0} & \orcell{40}{12.6} 
& \pkcell{60}{30.4} & \pkcell{65}{19.3} & \pkcell{65}{10.4} \\

Claude Opus 4.6 (High)
& \grcell{45}{16.3} & \grcell{60}{28.1} & \grcell{65}{20.7} 
& \orcell{45}{14.8} & \orcell{50}{14.8} & \orcell{65}{20.0} 
& \pkcell{35}{17.8} & \pkcell{40}{13.3} & \pkcell{45}{7.4} \\

Gemini 3.1 Pro (High)
& \grcell{30}{10.4} & \grcell{60}{28.9} & \grcell{50}{17.0} 
& \orcell{30}{10.4} & \orcell{40}{11.9} & \orcell{60}{18.5} 
& \pkcell{65}{34.1} & \pkcell{50}{15.6} & \pkcell{30}{5.2} \\

Hy3 preview (High)
& \grcell{40}{14.4} & \grcell{45}{22.2} & \grcell{35}{14.4} 
& \orcell{65}{20.0} & \orcell{45}{13.3} & \orcell{45}{14.4} 
& \pkcell{45}{23.3} & \pkcell{35}{12.2} & \pkcell{40}{6.7} \\

Seed 2.0 Pro (High)
& \grcell{65}{23.0} & \grcell{55}{26.7} & \grcell{15}{9.6} 
& \orcell{30}{10.4} & \orcell{45}{14.1} & \orcell{40}{11.9} 
& \pkcell{50}{24.4} & \pkcell{30}{11.1} & \pkcell{50}{8.1} \\

Kimi K2.5 (High)
& \grcell{30}{9.6} & \grcell{50}{23.7} & \grcell{30}{12.6} 
& \orcell{30}{11.1} & \orcell{30}{9.6} & \orcell{45}{13.3} 
& \pkcell{35}{17.8} & \pkcell{35}{11.9} & \pkcell{55}{8.9} \\

Qwen 3.5 Plus (High)
& \grcell{40}{13.3} & \grcell{25}{14.8} & \grcell{20}{11.1} 
& \orcell{30}{11.1} & \orcell{25}{7.4} & \orcell{40}{11.9} 
& \pkcell{40}{20.7} & \pkcell{50}{15.6} & \pkcell{35}{5.9} \\

Grok 4.20 
& \grcell{30}{10.4} & \grcell{30}{16.3} & \grcell{25}{11.9} 
& \orcell{40}{13.3} & \orcell{40}{12.6} & \orcell{30}{9.6} 
& \pkcell{35}{15.6} & \pkcell{40}{13.3} & \pkcell{20}{3.7} \\

DeepSeek V3.2 Thinking 
& \grcell{25}{7.4} & \grcell{30}{17.0} & \grcell{30}{12.6} 
& \orcell{20}{8.1} & \orcell{25}{8.1} & \orcell{20}{6.7} 
& \pkcell{30}{12.6} & \pkcell{15}{7.4} & \pkcell{35}{5.9} \\

MiniMax M2.5
& \grcell{15}{4.4} & \grcell{15}{10.4} & \grcell{20}{11.1} 
& \orcell{15}{6.7} & \orcell{15}{4.4} & \orcell{15}{4.4} 
& \pkcell{15}{5.2} & \pkcell{15}{7.4} & \pkcell{15}{3.0} \\

\bottomrule
\end{tabular}%
}
\vspace{-1em}
\label{tab:subcategory_breakdown}
\end{table*}

%% file: tables/error_analysis.tex
\definecolor{deepred}{HTML}{C0392B}

\definecolor{mycellblue}{rgb}{0.71,0.773,0.945}

\newcommand{\gradientcellerror}[1]{%
  \makebox[\linewidth][c]{%
    \begin{tikzpicture}[baseline]
      \fill[gray!25] (0,0) rectangle (1.9,0.3);
      \fill[gray!100] (0,0) rectangle ({1.9*#1/100},0.3);
      \node[anchor=west, font=\scriptsize] at (2,0.15) {%
        \textbf{\makebox[1.2em][r]{#1}}%
      };
    \end{tikzpicture}%
  }%
}

\begin{table*}[t]
\caption{
Distribution of error types across models.
Error types are non-mutually exclusive, a single failed solution may exhibit multiple error types.
The majority of solving failures are attributed to misusing knowledge in the context.
}
\centering
\small
\scalebox{0.86}{
\begin{tabular}{@{}l|>{\centering\arraybackslash}p{2.3cm}|>{\centering\arraybackslash}p{2.3cm}|>{\centering\arraybackslash}p{2.3cm}|>{\centering\arraybackslash}p{2.3cm}@{}}
\toprule
\addlinespace[0.7em]
\multicolumn{1}{l|}{\textbf{Model Names}} &
\multicolumn{1}{c|}{\makecell{\textbf{Context} \\ \textbf{Ignored(\%)}}} &
\multicolumn{1}{c|}{\makecell{\textbf{Context} \\ \textbf{Misused (\%)}}} &
\multicolumn{1}{c|}{\makecell{\textbf{Format} \\ \textbf{Error (\%)}}} &
\multicolumn{1}{c}{\textbf{Refusal (\%)}} \\
\addlinespace[0.3em]
\midrule

GPT 5.4 (High) & \gradientcellerror{36.0} & \gradientcellerror{79.7} & \gradientcellerror{13.2} & \gradientcellerror{1.5} \\

Claude Opus 4.6 (High) & \gradientcellerror{38.9} & \gradientcellerror{76.0} & \gradientcellerror{16.2} & \gradientcellerror{0.9} \\

Gemini 3.1 Pro (High) & \gradientcellerror{44.4} & \gradientcellerror{77.8} & \gradientcellerror{14.3} & \gradientcellerror{2.3} \\

Hy3 preview (High) & \gradientcellerror{35.2} & \gradientcellerror{84.2} & \gradientcellerror{10.9} & \gradientcellerror{0.9} \\

Seed 2.0 Pro (High) & \gradientcellerror{41.4} & \gradientcellerror{80.2} & \gradientcellerror{13.8} & \gradientcellerror{1.4} \\

Kimi K2.5 (High) & \gradientcellerror{45.3} & \gradientcellerror{80.2} & \gradientcellerror{12.7} & \gradientcellerror{1.1} \\

Qwen 3.5 Plus (High) & \gradientcellerror{45.1} & \gradientcellerror{76.9} & \gradientcellerror{13.0} & \gradientcellerror{2.0} \\

Grok 4.20 & \gradientcellerror{40.0} & \gradientcellerror{83.7} & \gradientcellerror{12.4} & \gradientcellerror{2.5} \\

DeepSeek V3.2 Thinking  & \gradientcellerror{39.1} & \gradientcellerror{83.1} & \gradientcellerror{15.3} & \gradientcellerror{1.6} \\

MiniMax M2.5 & \gradientcellerror{43.5} & \gradientcellerror{84.4} & \gradientcellerror{13.4} & \gradientcellerror{0.8} \\

\bottomrule
\end{tabular}
}
\vspace{-1em}
\label{tab:error_types}
\end{table*}

%% file: tables/no_context_ablation.tex
\begin{table}[htpb]
\caption{
Context ablation study. We strip all contexts from each task, retaining only the final user question, and evaluate GPT-5.4 under the same conditions. Performance drops from 19.3\% to 1.7\%, confirming that CL-bench Life tasks cannot be solved without the provided context.
}
    \centering
    \small
    \resizebox{0.86\columnwidth}{!}{%
    \begin{tabular}{@{}l|c|c|c|c@{}}
    \toprule
    \textbf{Setting} &
    \textbf{Overall} &
    \makecell{\textbf{Communication \&} \\ \textbf{Social Interactions}} &
    \makecell{\textbf{Fragmented Information \&} \\ \textbf{ Revisions}} &
    \makecell{\textbf{Behavioral Records \&} \\ \textbf{Activity Trails}} \\
    \midrule
    With Context & 19.3\% & 22.2\% & 15.8\% & 20.0\% \\
    Without Context & 1.7\% & 2.2\% & 0.7\% & 2.2\% \\
    \midrule
    $\Delta$ & \textbf{--17.6} & --20.0 & --15.1 & --17.8 \\
    \bottomrule
    \end{tabular}
    }
    \label{tab:no_context}
\end{table}

%% file: tables/judge_consistency.tex
\begin{table}[htpb]
    \centering
    \small
    \caption{
Agreement across judge models. All judge model pairs show agreement rates above 93\%, with Cohen's $\kappa$ values ranging from 0.710 to 0.773, indicating substantial agreement beyond chance.
    }
    \begin{tabular}{@{}l|c|c@{}}
    \toprule
    \textbf{Judge model pair} &
    \textbf{Agreement (\%)} &
    \textbf{Cohen's $\kappa$} \\
    \midrule
    GPT-5.1 vs Claude-Opus-4.6 & 93.2 & 0.724 \\
    GPT-5.1 vs Gemini-3.1-Pro & 93.0 & 0.710 \\
    Claude-Opus-4.6 vs Gemini-3.1-Pro & 94.5 & 0.773 \\
    \bottomrule
    \end{tabular}
    % \vspace{-1em}
    \label{tab:judge_consistency}
\end{table}

%% file: cases/case1.tex
\begin{table}[ht]
\caption{
A discourse--action alignment task over a ${\sim}$3-year landlord--tenant chat history. Gemini 3.1 Pro detects individual maintenance contradictions but defaults to a monotonic trust-decay narrative, failing to weigh competing positive evidence.
}
\centering
\begin{tabular}{p{\linewidth}}
    \toprule
    \textbf{Task Information}
    \begin{itemize}
        \item uid: 852384e2-c518-5f03-168f-e96121e20080
        \item Model: Gemini 3.1 Pro (High)
        \item Context Category: Communication \& Social Interactions
        \item Context Sub-Category: Private Conversations
        \item Task: Identify specific instances where the landlord's polite reassurances directly contradicted his follow-through on maintenance issues. Based on these instances, how does the power dynamic and level of trust between the tenant and landlord evolve from the beginning of the lease to the end?
    \end{itemize} \\
    \midrule
    \textbf{Context} \\
    \input{cases/context/case1} \\
    \bottomrule
\end{tabular}
\label{table:category1_case1}
\end{table}

\begin{table}
\begin{tabular}{p{\linewidth}}
\toprule
\textbf{Rubrics} \\
\input{cases/rubric/case1} \\
\midrule
\textbf{Model Generated Solution} \\
\input{cases/response/case1} \\
\midrule
\textbf{Discussion.} This task demands discourse--action alignment over a ${\sim}$3-year, ${\sim}$500-message conversation---the model must contrast what the landlord \textit{said} against what \textit{actually happened} across messages separated by months, then synthesize a trust trajectory from competing evidence. The model detects salient contradictions (e.g., the water heater promise--failure gap), but exhibits two key reasoning failures: (i) it misses temporally proximate evidence---the washer declared fixed April 12, 2025 but leaking 8 days later---which is stronger than the distant episodes it selects, suggesting it retrieves by salience rather than evidential strength; (ii) it defaults to a monotonic trust-decay narrative (``Low \ldots\ they no longer trust his maintenance promises''), unable to weigh maintenance frustration against strong counter-signals (friend referrals, unsupervised apartment access, personal gifts through the final months). The core limitation is negativity bias in relational inference: once contradictions are identified, the model cannot integrate positive evidence to reach a non-binary trust assessment.
\\
\bottomrule
\end{tabular}
\end{table}
\newpage

%% file: cases/context/case1.tex
\textcolor{gray}{\textbf{(Part I)} [\textit{Tenancy setup and early maintenance}]}

[2/23/23, 6:18:21 PM] Sarah Miller: Hi Arben, this is Sarah \& David from the apartment! We just signed our lease today. we are hoping to move in on Saturday. Are you able to help us get the internet set up? We would like the Telecom company and the 100/50 mbps package for 1600 Lekë per month.

[3/7/23, 10:11:41 AM] Sarah Miller: Hi Arben, we have a few issues in the apartment. 1) The facet in the bathroom sink is unconnected from the counter top. It still works but doesn't stand up anymore. 2) there is some sort of leak under the washer that happens when it's running and not running. 3) the oven is tripping the power. Every time we turn it on, the whole apartment turns off.

. . . \textcolor{gray}{\textit{(Early months include internet installation, AC heating issues, repeated power outages requiring breaker resets, toilet replacement, and drain/plumbing problems---all resolved promptly by the property manager.)}}

\textcolor{gray}{\textbf{(Part II)} [\textit{Lease extension, visa assistance, and escalating infrastructure failures}]}

[9/11/23, 5:48:44 PM] Sarah Miller: Also, we are looking to extend our stay in Albania and would like to stay in the apartment another year.

[9/13/23, 11:01:41 AM] Sarah Miller: We would love to get an extended visa instead of having to leave for 3 months and then return for another year. However to get the visa we need some sort of work contract. We currently don't have any contracts since we freelance. Do you have any ideas on what we could do?

[7/17/24, 8:36:46 AM] Arben Rruga B: Gimi told me: Last night, some electric cables were burnt there. The fire brigade came and extinguished the fire. The black film is from the fire.

[7/17/24, 2:12:39 PM] Sarah Miller: We understand. It's been a bit hard with us trying to work from home though. We really love the place and you guys have been so wonderful to us but there have just been so many power issues.

. . . \textcolor{gray}{\textit{(This period includes two lease extensions, visa procedure referral, water heater repaired/replaced $\geq$4 times, an electrical fire requiring 4-night hotel relocation reimbursed at 300\texteuro{}, and pet dog negotiation.)}}

\textcolor{gray}{\textbf{(Part III)} [\textit{Ongoing issues, lifestyle requests, and deepening relationship}]}

[5/29/24, 8:00:26 PM] Arben Rruga B: Hi - while it is not our preference to have pets in the apartment, we want to accommodate your request. We are concerned about damage to the furniture, and in particular all the bedding (couch, bed, etc). We trust you will continue to take care of the apartment as you have so far - like it's your own. Good luck!

[4/8/25, 11:03:19 AM] Sarah Miller: Oh dear. The washer just started smoking when I opened it with a done load. Not sure if you can see it but it stinks.

. . . \textcolor{gray}{\textit{(Later period includes custom ottoman commissioning, olive oil sourcing, shipping logistics advice, recurring washer/internet failures, traffic ticket assistance, and continued pre-return cleanings---reflecting a relationship well beyond a typical landlord--tenant dynamic.)}}

%% file: cases/rubric/case1.tex
1. The response should not classify the toilet seat issue as an instance of failed follow-through.

2. The response should identify the contradiction regarding the water heater by noting that Arben promised it would be fixed or replaced in February and April 2024 (e.g., "Next week we will change it with a new one"), but the heater failed again by November 2024.

3. The response should identify the contradiction regarding the washing machine by noting that Arben stated the washer was fixed on April 12, 2025, but Sarah reported it leaking on April 20, 2025, and smoking on June 3, 2025.

4. The response should describe the evolution of the relationship as maintaining a high level of trust at the end.  For example, mentioning the tenants looking for recommendations for their friends.

5. The response should not claim that the relationship became "adversarial" or "strained" without acknowledging the counter-evidence. For example, the tenants trusting the landlord to enter their apartment while they are not there.

6. The response should explicitly connect the identified instances of contradiction to the overall evolution of trust and power dynamics, rather than addressing the two parts of the prompt separately. For example, it should address the contradictions and the evolution of trust at the same time.

7. The response should present events in the correct chronological order. For example, a message dated June 5 should be placed before a message dated June 6.

%% file: cases/response/case1.tex
. . . \textcolor{gray}{\textit{(Opening paragraph describes Arben's polite language as a ``buffer'' substituting for structural fixes, followed by a transition sentence.)}}

\#\#\# Instances of Contradicted Reassurances

**1. The ``New'' Hot Water Heater (April 2024 vs. November 2024)**

*   **The Reassurance:** On April 18, 2024, after the water heater breaks down again, Arben promises a permanent fix: *``Today they will come to fix it temporary. Next week we will change it with a new one.''*

*   **The Contradiction:** Seven months later, on November 4, 2024, the water heater breaks *again*. On November 5, Arben responds: *``Hi. Gimi will bring tomorrow a new one.''* This proves he never actually followed through on replacing the unit back in April, despite his explicit promise. He merely patched it and waited for it to fail again.

. . . \textcolor{gray}{\textit{(Omitted instances 2--4: ``Phantom Washer Inspections'' (Aug 2023 \& Dec 2024), outsourcing washer repair via DIY link (Mar 2025), and perpetual internet failures (2025--2026).)}}

\#\#\# The Evolution of the Power Dynamic and Trust

. . . \textcolor{gray}{\textit{(Omitted preamble sentence.)}}

**Phase 1: High Trust \& Cooperative Power (Early 2023)**

. . . \textcolor{gray}{\textit{(Omitted details: tenants deferential, Arben responds within 30 minutes, brings welcome wine.)}}

**Phase 2: Landlord Gatekeeping \& Eroding Trust (2024 -- Mid 2025)**

. . . \textcolor{gray}{\textit{(Omitted details: polite language as placation, tenants follow up and document issues.)}}

**Phase 3: Resignation, Deflection, and Burden-Shifting (Late 2025 -- 2026)**

. . . \textcolor{gray}{\textit{(Omitted Dynamic: Arben shifts maintenance labor onto tenants, asks them to diagnose error codes, blames them for faulty machinery; tenants' language becomes more direct and less apologetic.)}}

*   **Trust Level:** Low regarding competence and reliability. By the end of the lease, the tenants still like Arben personally (buying him wine, chocolate, and gifts), but they no longer trust his maintenance promises. They accept that living there means dealing with broken doors, leaking washers, and dropping internet, acknowledging that his polite ``I will check tomorrow'' is a euphemism for inaction.

%% file: cases/case2.tex
\begin{table}[ht]
\caption{
A role inference and workflow reconstruction task over Slack JSON logs. Gemini 3.1 Pro correctly reconstructs the query triage pattern and identifies bottlenecks, but mischaracterizes the organizational hierarchy by conflating workflow initiation with decision-making authority.
}
\centering
\begin{tabular}{p{\linewidth}}
    \toprule
    \textbf{Task Information}
    \begin{itemize}
        \item uid: 130ae25f-984a-efad-4741-2c80c2921368
        \item Model: Gemini 3.1 Pro (High)
        \item Context Category: Communication \& Social Interactions
        \item Context Sub-Category: Group Conversations \& Meeting Transcripts
        \item Task: Analyze the instances in this thread where the team initially lacks a clear answer for a user. Based on how these uncertainties are routed and resolved, deduce the specific domain expertise and hierarchical authority of the team. Provide the organizational hierarchy in a clear bulleted list with assumed job titles. Support your analysis by tracing the lifecycle of at least one food-related query and one gardening/DIY-related query from initial confusion to final resolution. Identify any instances where a query is left unresolved and explain why that happened based on the established team dynamics. Any log jams in the process should be identified with bold text and set apart from other text as to draw attention.
    \end{itemize} \\
    \midrule
    \textbf{Context} \\
    \input{cases/context/case2} \\
    \bottomrule
\end{tabular}
\label{table:category1_case2}
\end{table}

\begin{table}
\begin{tabular}{p{\linewidth}}
\toprule
\textbf{Rubrics}  \\
\input{cases/rubric/case2} \\
\bottomrule
\end{tabular}
\end{table}

\begin{table}
\begin{tabular}{p{\linewidth}}
\toprule
\textbf{Model Generated Solution (Continued)} \\
\input{cases/response/case2} \\
\midrule
\textbf{Discussion.} This task requires reverse-engineering organizational roles and workflow structure from raw Slack JSON---the model must infer who holds authority not from explicit titles but from how questions are intake, escalated, decided, and back-filled across threaded messages. This is harder than standard chat summarization, demanding role inference, workflow reconstruction, and bottleneck analysis simultaneously, though it stops short of the most challenging multi-turn identity-tracking tasks. Gemini 3.1 Pro reconstructs the workflow pattern well (Brenda routes $\rightarrow$ Alice defers $\rightarrow$ Clara resolves), correctly identifies both unresolved queries with causal explanations, and highlights the single-SME bottleneck. However, it mischaracterizes the hierarchy: it places Alice as ``Leadership'' and states Clara ``doesn't appear to manage the team administratively,'' over-indexing on workflow-initiating speech acts (Alice's ``We are going to try something new'') while under-weighting Clara's decision-making authority (e.g., her unilateral ``Delete it.''). It also anchors the log jam to Feb 5--8 rather than the required Feb 5--10, reflecting imprecise temporal boundary reasoning over threaded timestamps.
\\
\bottomrule
\end{tabular}
\end{table}
\newpage

%% file: cases/context/case2.tex
\textcolor{gray}{\textbf{(Part I)} [\textit{Workflow setup and comment triage pattern}]}

\begin{verbatim}
{"user": "U1AAAAAAA", "ts": "1611769201.001600",
 "text": "We are going to try something new with managing comments!
  Brenda, can you drop each comment that needs an answer here
  instead of in an email? I'll answer the ones I can, and tag in
  <@U3CCCCCCC> when I don't know the answer.",
 "reactions": [{"name":"+1::skin-tone-2",
  "users":["U3CCCCCCC","U2BBBBBBB"],"count":2}]}

{"user": "U2BBBBBBB", "ts": "1612372594.005000",
 "text": "Amy would like to know if she can sub cinnamon powder
  for cinnamon sticks in the elderberry syrup ..."}

{"user": "U1AAAAAAA", "ts": "1612372661.005400",
 "text": "I feel like it would mess with the texture of the syrup
  and make it kind of gritty? <@U3CCCCCCC>?"}

{"user": "U3CCCCCCC", "ts": "1612372760.006100",
 "text": "Yeah, ground cinnamon doesn't dissolve. It'll be fine,
  but they'll need to stir/shake vigorously ..."}
\end{verbatim}

. . . \textcolor{gray}{\textit{(This pattern repeats across ${\sim}$15 reader questions on recipes, DIY products, and gardening. Alice (U1AAAAAAA) consistently defers to Clara (U3CCCCCCC) as domain expert; Brenda (U2BBBBBBB) relays and posts replies.)}}

\textcolor{gray}{\textbf{(Part II)} [\textit{Content moderation and blog post corrections}]}

\begin{verbatim}
{"user": "U2BBBBBBB", "ts": "1613405863.003500",
 "text": "Is it okay to delete this comment from angryuser(at)
  yahoo(dot)com: \"...Reusable toilet paper? Not only is that
  disgusting but I feel bad for all the husbands and children
  ...\""}

{"user": "U3CCCCCCC", "ts": "1613486321.006900",
 "text": "Delete it."}
\end{verbatim}

. . . \textcolor{gray}{\textit{(The channel also handles blog corrections---e.g., Clara finds a screw size error and asks Brenda to edit the post---forming a lightweight editorial workflow alongside comment triage.)}}

%% file: cases/rubric/case2.tex
1. The response should present the organizational hierarchy in a clear bulleted list.

2. The response should include assumed job titles for Alice Smith, Brenda Jones, and Clara Williams within the hierarchy list. For example, it should list job titles as CEO, Manager, Editor, or Liaison.

3. The response should identify Clara Williams (U3CCCCCCC) as the primary authority or manager of the team. For example, it should cite her decision-making power in moderation during her Feb 16 instruction to "Delete it" regarding a troll comment.

4. The response should identify Alice Smith (U1AAAAAAA) as a person who manages the workflow of the team and answers general comments. For example, it should identify her as a content manager.

5. The response should identify Brenda Jones (U2BBBBBBB) as a support person responsible for collecting user feedback and tracking unanswered queries. For example, it should identify her as community liaison or manager.

6. The response should trace the lifecycle of a food-related query. For example, it should trace Amy’s elderberry syrup cinnamon substitution query from Brenda’s initial post asking if ground cinnamon is okay to use in place of stick cinnamon to Clara’s final resolution saying ground cinnamon will cause the final syrup to be grainy.

7. The response should trace the lifecycle of one gardening-related query. For example, it should trace Daniel’s raised bed screws query from Brenda asking what kind of screws were used to the final update where Clara provides the final resolution answer of 3 inch deck screws.

8. The response should identify Matthew’s query regarding organic certification for galvanized steel beds as an unresolved instance.

9. The response should identify the period between Feb 5 and Feb 10 as a log jam where multiple user comments (from Megan, Cora, Peter, Daniel, and Haley) remained unanswered.

10. The response should bold the specific text used to identify log jams in the process (e.g., **Log Jam** or **Process Stall**).

11. The response should set the log jam identifications apart from the main analysis text. For example, using a separate section, bold headers, or distinct formatting to draw attention.

12. The response should explain why Matthew's query went unresolved based on the team dynamics. For example, it should note that Clara, the sole domain expert, did not know the answer and there was no mechanism to escalate it further.

13. The response should explicitly state the specific domain expertise areas of the team members. For example, it should mention Clara's expertise as a "master gardener" and her knowledge of recipe chemistry/food science, Alice's general knowledge of cooking and gardening, and Brenda's general cooking knowledge.

14. The response should analyze how uncertainties are routed among the team. For example, it should note that Brenda collects the questions and routes them to Alice, who answers what she can before escalating to Clara.

15. The response should identify Jenna's query regarding blurry printed tags (which received zero replies) as an unresolved instance.

16. The response should explain why Jenna's query went unresolved. For example, it should explain it was completely missed by the team after Brenda failed to explicitly tag Alice or Clara to route it properly.\\
\midrule
\textbf{Model Generated Solution}

Based on the Slack logs provided, the team manages a wellness and gardening blog network (including "Wellnessblog" and "Gardenblog"). The team dynamics reveal a clear workflow where frontline inquiries are filtered, escalated to management, and ultimately resolved by a singular Subject Matter Expert (SME). 

Here is an analysis of the team’s domain expertise, authority, and workflows.

%% file: cases/response/case2.tex
\#\#\# Organizational Hierarchy and Assumed Job Titles

. . . \textcolor{gray}{\textit{(Introductory sentence on routing, delegation, and decision-making.)}}

*   **Alice Smith -- Managing Editor / Operations Manager (Leadership)**
    *   *Authority:* Sets team workflows (``We are going to try something new...''), has high-level knowledge of site migrations (``That post is about to move to Gardenblog''), and makes policy decisions (e.g., whether to delete abusive comments). She acts as a traffic director but openly lacks the specific domain expertise to answer technical user questions.
*   **Clara Williams -- Lead Content Creator / Subject Matter Expert (SME)**
    *   *Authority:* Holds the ultimate domain expertise. She is the author of the recipes, DIY tutorials, and cookbooks. While she doesn't appear to manage the team administratively, she holds creative authority, dictates recipe adjustments, and requests post edits.
*   **Brenda Jones -- Community Manager / Customer Support Specialist (Frontline)**
    *   *Authority:* Responsible for frontline moderation and customer engagement. She paraphrases or copies user comments, proposes potential responses for approval, and executes post updates once the SME provides the correct information.

. . . \textcolor{gray}{\textit{(Omitted two full query lifecycle traces: Instant Pot Coconut Yogurt (food-related) and Raised Bed Conduits (DIY/gardening-related). Both follow the same pattern: Brenda routes $\rightarrow$ Alice defers $\rightarrow$ Clara resolves.)}}

\#\#\# Unresolved Queries

. . . \textcolor{gray}{\textit{(Introductory sentence.)}}

1.  **The Organic Certification Question:** A user asks if using Galvanized Steel Beds will cause issues with getting an organic certification. . . . Clara explicitly replies, ``I'm not sure. I don't know anything about the organic gardening certification.'' *Why it happened:* This represents the absolute limit of Clara's domain expertise. The team relies entirely on her personal experience; if she hasn't experienced it, the team has no secondary expert to consult.
2.  **The Blurry Soup Tags Question:** Brenda posts that a user is having issues printing tags for soup jars. This message receives zero replies. *Why it happened:* Brenda posted the message but failed to explicitly tag Alice or Clara. . . . the query becomes invisible and dies in the feed.

\#\#\# Log Jams in the Process

**A significant bottleneck in this workflow is the team's total reliance on a single Subject Matter Expert (Clara) to resolve customer issues. Because Alice cannot answer technical queries, everything funnels to Clara. When Clara is unavailable---such as when she took a day off Slack to design ``Gardenblog''---the frontline workflow completely stalls.**

**This log jam is explicitly visible in the logs when queries pile up for nearly a week. Brenda posts five distinct user questions between February 5th and February 8th. Because Clara is absent or busy, Alice realizes the flaw in the system, noting, ``I think I need to tag Clara in on all of these.'' However, nothing moves until Brenda is forced to issue a bulk reminder days later: *``Just a reminder @Clara that there are some comments here in need of your expertise! Starting with Megan then Cora, Peter, Daniel, and Haley - all listed above.''* This single point of failure severely delays response times for the community.**

%% file: cases/case3.tex
\begin{table}[ht]
\caption{
A cross-thread opinion mining task over multiple outdoor forums. GPT-5.4 extracts salient disputes but substitutes surface-level disagreements for the required structural controversies and community ethics.
}
\centering
\begin{tabular}{p{\linewidth}}
    \toprule
    \textbf{Task Information}
    \begin{itemize}
        \item uid: 9f5f44db-d3cd-df93-6f43-8196dc6bab7f
        \item Model: GPT-5.4 (High)
        \item Context Category: Communication \& Social Interactions
        \item Context Sub-Category: Community Interactions
        \item Task: From these forums list 3 differing opinions and point of contention about shed hunting, provide 1 quote for each side of the disagreement. Also list 3 ethical decisions and any attempts to reconcile them (like scaring animals when they are vulnerable), support them with one quote each. If there is no attempt to reconcile the ethical issue within the threads provide "none stated". For all quotes provide username and thread name as citation.
    \end{itemize} \\
    \midrule
    \textbf{Context} \\
    \input{cases/context/case3} \\
    \bottomrule
\end{tabular}
\label{table:category1_case3}
\end{table}

\begin{table}
\begin{tabular}{p{\linewidth}}
\toprule
\input{cases/rubric/case3} \\
\midrule
\textbf{Model Generated Solution} \\
\input{cases/response/case3} \\
\midrule
\textbf{Discussion.} This task requires reconstructing debate axes and ethical boundaries from scattered posts across multiple community threads---the model must distinguish tactical disagreements from normative disputes, identify opposing sides with precise quotations, and assess whether the community attempts reconciliation. GPT-5.4 correctly identifies the media/commercialization contention and the wildlife disturbance ethic with an appropriate reconciliation quote, demonstrating competent cross-thread evidence retrieval. However, it substitutes its own plausible-but-incorrect controversy axes for the required ones: it provides ``miles vs.\ focused searching'' instead of ``methodical search vs.\ casual observation,'' and omits ``sharing information vs.\ competitive secrecy'' entirely---a pattern suggesting the model defaults to the most salient surface-level disagreements rather than inferring subtler structural tensions. The ethical issues show a similar failure: ``interfering with turkey hunters'' replaces the required ``competing with other hunters,'' and the trespassing issue lacks a reconciliation quote. These errors indicate that while the model can extract explicit disputes, it struggles to abstract latent community norms from indirect conversational cues.
\\
\bottomrule
\end{tabular}
\end{table}
\newpage

%% file: cases/context/case3.tex
\textcolor{gray}{\textbf{(Part I)} [\textit{Thread 1: ``Shed hunting'' --- outdoor forum, Jan--Feb 2019, bumped 8d ago}]}

hiker\_p \#1 · Jan 24, 2019: Want to go shed hunting for the first time this year and looking for some pointers of what to do/not to do and what to look for. Not looking for hotspots or coordinates to public storage sheds.

RamHunter \#2 · Jan 24, 2019: Good luck is about the only advice I can give you. I used to shed hunt hard in Utah way before Tines up and hushin ruined it for everyone who enjoyed it prior to their gay boy club. all the wannabe cool kids running around these days pushing all the 1 horned animals all over killed it for me.

FirstFelix \#4 · Jan 25, 2019: Tines Up started in 2005 when two of the five members were deployed to Ramadi, Iraq. . . . You want to edit your hateful homophobic rant or you gonna leave it up?

Chocolate \#7 · Jan 25, 2019: The funny thing is the Tines Up guys were probably stacking up sheds before he ever thought about it. . . . No reason to be so vitriolic all the time.

XYZ \#10 · Jan 26, 2019: find winter grounds look under head height trees. . . . if possible, try not to push the animals. it is an important time of the year for them to recover from winter.

commander77 \#13 · Jan 28, 2019: My favorite time of year to go is the first part of April. By then all the Moose and Deer have dropped, and elk are probably half way done.

JFK \#14 · Jan 28, 2019: The biggest advice I can give to those heading out to hunt sheds is: DON'T SCREW UP THE TURKEY HUNTERS!!

ForestWalker \#15 · 8d ago: Took the dog out this afternoon . . . stumbled on my first shed of the year. . . . Signs of spring and a friendly reminder to review the ethics course.

. . . \textcolor{gray}{\textit{(Omitted posts \#3, \#8, \#9, \#11 with regulation reminders, hipster jokes, and general search tips.)}}

\textcolor{gray}{\textbf{(Part II)} [\textit{Thread 2: ``Shed hunting tips and tricks'' --- Rokslide forum, Feb--Apr 2023}]}

OutdoorsUser01 \#1 · Feb 17, 2023: I'd love to hear some of your best tips and tricks on how to best find sheds in your area.

AB99 \#11 · Mar 7, 2023: focus 90\% of your time searching 10\% of the areas. . . . I think so many people think walking ten plus miles a day will yield results, when they'd be better off hitting the good spots several times over.

. . . \textcolor{gray}{\textit{(Omitted posts \#2--\#10, \#12--\#15 with tips on bedding areas, food sources, fence lines, south-facing slopes, and regional variation.)}}

\textcolor{gray}{\textbf{(Part III)} [\textit{Reddit post: ``Shed hunting advice (pic for attention)'' --- r/shedhunting, ${\sim}$2y ago}]}

m\_fisher (OP): I've been shed hunting for the last 4-5 years . . . I find bedding areas, food and water sources. I search in pine groves and on the south sides of hills and mountains. . . . and I have found exactly 0 antlers.

Distinct\_Hike5678: Big thing is don't go early. Might scare them all out before they drop

outlaw-9: I've been doing it since the 80's . . . We try to get out the first of Jan where we can follow trails etc , found antlers as early as Jan 1st

. . . \textcolor{gray}{\textit{(Omitted ${\sim}$20 additional comments with search technique advice, encouragement, and regional tips.)}}

\textcolor{gray}{\textbf{(Part IV)} [\textit{Thread 3: ``Looking for some Shed hunting tips'' --- archery forum, Feb 2019}]}

TenPointer \#12 · Feb 13, 2019: in many parts trespassing can be MUCH worse in shed/morel seasons than it is in hunting season. I ran into a guy last year whose response was, ``I'm not hunting''; kind of drives you crazy. 

%% file: cases/rubric/case3.tex
\textbf{Context (Continued)}

TenPointer \#17 · Feb 18, 2019: An overwhelming majority of the sheds that I found are found within 5-15 FEET of where I am standing. . . . Go slow and keep your eyes down.

ArrowSlinger \#25 · Feb 20, 2019: When i started shed hunting 15 years ago I wouldn't see anybody . . . Beginning about 7 years ago boot tracks everywhere and cameras missing.

. . . \textcolor{gray}{\textit{(Omitted ${\sim}$15 additional posts with general tips, gear advice, dog training, and property-specific anecdotes. The thread also includes ${\sim}$20 ``Related Threads'' links on topics ranging from hunting dogs to Colorado shed regulations.)}} \\
\midrule

\textbf{Rubrics}

1. The response should list 3 distinct ethical decisions and reconciliation attempts. For example, none of the ethical decisions should have major overlap between them, such as two referring to disturbing wildlife.

2. The response should state Minimizing Disturbance to Vulnerable Wildlife as one of the ethical decision and reconciliation attempts. Paraphrasing is acceptable.

3. The response should state Trespassing for Access as one of the ethical decision and reconciliation attempts. Paraphrasing is acceptable.

4. The response should state Competing with Other Hunters as one of the ethical decision and reconciliation attempts. Paraphrasing is acceptable.

5. The response should list 3 distinct points of contention about shed hunting. For example, there should not be major overlap between two or more of the points, such as two explicitly discussing competitive secrecy.

6. The response should state Impact of Media and Social Media on Shed Hunting Culture as one of the points of contention. Paraphrasing is acceptable.

7. The response should state Optimal Shed Hunting Strategy: Methodical Search vs. Casual Observation as one of the points of contention. Paraphrasing is acceptable.

8. The response should state Culture of Sharing Information vs. Competitive Secrecy as one of the points of contention. Paraphrasing is acceptable.

9. The response should provide a username citation for each quote. For example, for the quote "Good luck is about the only advice I can give you...", the response should cite RamHunter; for the quote "Tines Up started in 2005...", the response should cite FirstFelix; for the quote "I find most of mine on edges...", the response should cite antiqueantlers; for the quote "Put some miles on the hiking boots", the response should cite jsmith; for the quote "Walk, walk, and walk more...", the response should cite Alex; for the quote "Best advice I've heard..." the response should cite AB99; for the quote "My method is pretty simple...", the response should cite spanielguy; for the quote "You have to scout for shed season...", the response should cite 2happy4mylife; for the quote "search bedding and food and trails...", the response should cite HardWorker; for the quote "Fence lines, bedding areas..." the response should cite StrayArrow; for the quote "Drive around. Seriously...", the response should cite MJJones; for the quote "Honestly, I personally just walk...", the response should cite Nature\_walker; for the quote "Maybe this is bad or unhelpful advice...", the response should cite 9988abc; for the quote "...we have two rules in groups that I organize...", the response should cite TenPointer; for the quote "When i started shed hunting...", the response should cite ArrowSlinger; for the quote "if possible, try not to..." the response should cite XYZ; for the quote "The biggest advice I can give..." the response should cite JFK; for the quote "Do what they do around my farm..." the response should cite FreshLook; for the quote "As simple as it sounds..." the response should cite TenPointer; for the quote "As you are finding out...", the response should cite commander77; for the quote "Go slow and keep your eyes down..." the response should cite TenPointer; for the quote "It sounds to me like...", the response should cite jsmith404; for the quote "I typically find 40-50...", the response should cite ChickenOverTheSea.

10. The response should provide a thread name citation for each quote. For example, for the quotes "Good luck is about the only advice I can give you..." "Tines Up started in 2005...", "if possible, try not to...", "The biggest advice I can give...", and "As you are finding out...", the response should cite the thread Shed hunting.; for the quotes "I find most of mine on edges...", "Put some miles on the hiking boots...", "Walk, walk, and walk more...", "Best advice I've heard...", "My method is pretty simple...", "Fence lines, bedding areas...", and "Drive around. Seriously...", the response should cite Shed hunting tips and tricks; for the quotes "You have to scout for shed season...",
"Honestly, I personally just walk...", "Maybe this is bad or unhelpful advice...", "It sounds to me like...", and "I \\
\bottomrule
\end{tabular}
\end{table}

\begin{table}
\begin{tabular}{p{\linewidth}}
\toprule
\textbf{Rubrics (Continued)}

typically find 40-50...", the response should cite Shed hunting advice (pic for attention); for the quotes "search bedding and food and trails...", "...we have two rules in groups that I organize...", "When i started shed hunting...", "Do what they do around my farm...", "Go slow and keep your eyes down...", and "As simple as it sounds...", the response should cite Looking for some Shed hunting tips.

11. The response should provide 1 quote for each side in a point of contention.

12. The response should provide a quote criticizing the impact of media and social media on shed hunting culture. For example, 
"Good luck is about the only advice I can give you. I used to shed hunt hard in Utah way before Tines up and hushin ruined it for everyone who enjoyed it prior to their gay boy club. all the wannabe cool kids running around these days pushing all the 1 horned animals all over killed it for me. Expect company, and don't plan on the animals still being in the area you left them in the weekend before, cuz I promise someone else had gone through there during the week and run them out..." RamHunter, Shed hunting.

13. The response should provide a quote defending the impact of media and social media on shed hunting culture. For example, 
"Tines Up started in 2005 when two of the five members were deployed to Ramadi, Iraq. While we were in Iraq, we realized how much the passion for the outdoors meant for us... Profound when young men in a foreign hell hole during the height of the Iraq war find comfort and reflection in the outdoors they'd experienced..." FirstFelix, Shed hunting.

14. The response should provide a quote supporting methodical search as the optimal shed hunting strategy. For example, 
"I find most of mine on edges. I also have some success on well known travel routes and bedding as well. Cloudy/rainy days are better than sunny days. As you're walking be sure to turn around often and look back the way you came from and different directions. When you do find one, pause for a second and look around really good. Sometimes the match is laying in close proximity. Anytime I find a large antler, I devote about an hour looking for the match in the same area. Winter food sources are a great starting point and the bedding closest to it." - antiqueantlers, Shed hunting tips and tricks
"Put some miles on the hiking boots and enjoy the outdoors. Keep your eyes low if your main goal is sheds. I personally get distracted looking at new spots and trees to setup in for next season lol" - jsmith, Shed hunting tips and tricks
"Walk, walk and walk more. As noted find travel and bedding places...silly whitetails around here drop them next to the road and 3000 ft higher..good bedding timber and some small southern exposure offer good opportunity." - Alex, Shed hunting tips and tricks
"Best advice I've heard, and the best I can give, is focus 90\% of your time searching 10\% of the areas. If the spot has something going for it, grid pattern the entire spot..." - AB99, Shed hunting tips and tricks
"My method is pretty simple. I walk miles of deer trails with my "shed eyes" on. I picked up 27 on my March trip to my place in Oklahoma and three more a couple weeks ago." - spanielguy, Shed hunting tips and tricks
"You have to scout for shed season just like deer season. Drive around, find the areas the bucks are actually in because they group up. Start in those areas. I can't stress enough how good waterways in fields are for shed. First 20 yards of woods and first 20 yards of the fields. Anything that greens early draws them like a magnet, wheat honeysuckle etc. Stream edges are good as well" - 2happy4mylife, Shed hunting advice (pic for attention)
"search bedding and food and trails from bedding to food. any fence lines or other obstacles they have to jump over. The biggest thing I would say, and the reason I find more sheds than most guys in my area is just put the time in and put your miles of walking in. walk slowly and scan the ground, its not a race and if you walk too fast I promise you will walk past them. best of luck" - HardWorker, Looking for some Shed hunting tips

15. The response should provide a quote supporting casual observation as the optimal shed hunting strategy. For example, 
"Fence lines, bedding areas by cedar wind breaks but usually just pure luck while hiking around." - StrayArrow, Shed hunting tips and tricks
"Drive around. Seriously, I have found my best sheds from the road." - MJJones, Shed hunting tips and tricks
"Honestly, I personally just walk wherever I feel I should and I end up finding them, I'm not the master at it but I'm finding 1-5 every year for the past 4 years." - Nature\_walker, Shed hunting advice (pic for attention)
"Maybe this is bad or unhelpful advice, but I have luck and find it less frustrating when I go out seeing instead of looking. I'll see what I see, and maybe I'll see an antler, maybe I'll see some other cool stuff. It's hard do describe but I think stuff turns up when I'm not trying so hard, and I'm less disappointed if nothing does." - 9988abc, Shed hunting advice (pic for attention)

16. The response should provide a quote supporting the culture of sharing information in shed hunting. For example, 
"...We have two rules in groups that I organize, finders-keepers EXCEPT if a match set is found whomever found the first side gets the match from whomever found it. This has worked very well..." - TenPointer, Looking for some Shed hunting tips\\
\bottomrule
\end{tabular}
\end{table}

\begin{table}
\begin{tabular}{p{\linewidth}}
\toprule
\textbf{Rubrics (Continued)}

17. The response should provide a quote supporting competitive secrecy in shed hunting. For example, 
"When i started shed hunting 15 years ago I wouldn't see anybody or see any sign of another shed Hunter. Beginning about 7 years ago boot tracks everywhere and cameras missing. Shed hunting has really exploded in popularity. When I do encounter a shed hunter I have tried to have conversation with them and they're usually so frenzied up with finding an antler they just brush you off. Antler zombies..." - ArrowSlinger, Looking for some Shed hunting tips

18. The response should provide quotes that explicitly support their associated side of a point of contention or ethical issue. For example, the quote should not provide arguments in favor of the opposing side and should only show support for the assigned point of view.

19. The response should provide a quote illustrating the ethical dilemma of minimizing disturbance to wildlife. For example,
"...if possible, try not to push the animals. it is an important time of the year for them to recover from winter." - XYZ, Shed hunting.
"The biggest advice I can give to those heading out to hunt sheds is: DON'T SCREW UP THE TURKEY HUNTERS!! Remember: shed hunting isn't the only thing going on. There are other people out trying to do their thing too. Don't screw them up." JFK, Shed hunting.

20. The response should provide a quote illustrating the ethical dilemma of trespassing for access. For example,
"Do what they do around my farm. Get permission from the smallest landowner who could care less what happens on his land then walk every piece of property you can walk in a days time. Then go out the following day and walk the opposite direction as far as you can. And don't forget to steal those game cameras that might take your picture!" - FreshLook, Looking for some Shed hunting tips
"As simple as it sounds, you have to search where the deer winter and in many cases that may not be where you hunt or even have permission to be walking. Not sure what its like near Buffalo, but in many parts trespassing can be MUCH worse in shed/morel seasons than it is in hunting season..." - TenPointer, Looking for some Shed hunting tips

21. The response should provide a quote illustrating the ethical dilemma of competing with other hunters. For example,
"As you are finding out, there is alot of competition when it comes to shed hunting, and few that are willing to help out when it comes to finding them..." - commander77, Shed hunting.

22. The response should provide a quote attempting to reconcile the ethical dilemma of minimizing disturbance to wildlife. For example, 
"It sounds to me like your doing a lot right. But I will say this. In the winter bucks hate people and being disturbed, make sure your walking farther enough off trail to get to them..." -jsmith404, Shed hunting advice (pic for attention)

23. The response should provide a quote attempting to reconcile the ethical dilemma of trespassing for access. For example, 
"I typically find 40-50 per year on 75\% public land..." ChickenOverTheSea, Shed hunting advice (pic for attention)

24. The response should provide a quote attempting to reconcile the ethical dilemma of competing with other hunters. For example, 
“Maybe this is bad or unhelpful advice, but I have luck and find it less frustrating when I go out seeing instead of looking. I’ll see what I see, and maybe I’ll see an antler, maybe I’ll see some other cool stuff...” - 9988abc, Shed hunting advice (pic for attention)

%% file: cases/response/case3.tex
. . . \textcolor{gray}{\textit{(Opening line and structural overview omitted.)}}

\#\# 1) Differing opinions / points of contention

\#\#\# A. Did social media / branded shed content ``ruin'' shed hunting?

**Side 1: Yes --- media and image culture hurt the activity**

> ``I was bashing the guys who turned shed hunting into a social media dick measuring contest with each other and trying to convince everyone how cool they are.''
--- **RamHunter**, *``Shed hunting''*

**Side 2: No --- those guys were long-time participants, and the reaction is overblown**

> ``The funny thing is the Tines Up guys were probably stacking up sheds before he ever thought about it\ldots\ No reason to be so vitriolic all the time.''
--- **Chocolate**, *``Shed hunting''*

\#\#\# B. Should you start shed hunting very early, or wait to avoid pushing deer?\\
\bottomrule
\end{tabular}
\end{table}

\begin{table}
\begin{tabular}{p{\linewidth}}
\toprule
\textbf{Model Generated Solution (Continued)}

**Side 1: Start early**

> ``We try to get out the first of Jan where we can follow trails etc , found antlers as early as Jan 1st and saw deer the same year still packing in march.''
--- **outlaw-9**, *``Shed hunting advice (pic for attention)''*

**Side 2: Wait --- going early can push deer before they drop**

> ``Big thing is don't go early Might scare them all out before they drop''
--- **Distinct\_Hike5678**, *``Shed hunting advice (pic for attention)''*

\#\#\# C. Is success mostly about covering lots of miles, or searching small high-probability areas thoroughly?

**Side 1: Cover ground and put miles on**

> ``Find where they winter, put the miles on''
--- **TrackingPro**, *``Shed hunting tips and tricks''*

**Side 2: Focus intensely on the best spots instead of just piling up miles**

> ``focus 90\% of your time searching 10\% of the areas\ldots\ I think so many people think walking ten plus miles a day will yield results, when they'd be better off hitting the good spots several times over.''
--- **AB99**, *``Shed hunting tips and tricks''*

\#\# 2) Ethical issues / decisions and attempts to reconcile them

\#\#\# 1. Avoid stressing or pushing animals during a vulnerable winter period

> ``if possible, try not to push the animals. it is an important time of the year for them to recover from winter.''
--- **XYZ**, *``Shed hunting''*

**Attempt to reconcile:**

> ``My favorite time of year to go is the first part of April. By then all the Moose and Deer have dropped, and elk are probably half way done.''
--- **commander77**, *``Shed hunting''*

**Reconciliation summary:** wait until later in the season so animals are less likely to be disturbed before they've dropped.

\#\#\# 2. Don't interfere with other users of the outdoors, especially turkey hunters

> ``The biggest advice I can give to those heading out to hunt sheds is: DON'T SCREW UP THE TURKEY HUNTERS!!''
--- **JFK**, *``Shed hunting''*

**Attempt to reconcile:**

> ``Remember: shed hunting isn't the only thing going on. There are other people out trying to do their thing too. Don't screw them up.''
--- **JFK**, *``Shed hunting''*

\#\#\# 3. Respect property boundaries / don't use ``I'm not hunting'' as an excuse to trespass

> ``in many parts trespassing can be MUCH worse in shed/morel seasons than it is in hunting season. I ran into a guy last year whose response was, `I'm not hunting'; kind of drives you crazy.''
--- **TenPointer**, *``Looking for some Shed hunting tips''*

**Attempt to reconcile:** **none stated**

. . . \textcolor{gray}{\textit{(Closing line offers to produce a comparison chart or coded theme list.)}}

%% file: cases/case4.tex
\begin{table}[ht]
\caption{
An evidence-grounded planning task over a fragmented personal note archive. GPT-5.4 links past ride failures to checklist items effectively but over-includes archival reference data and misses specific safety-monitoring requirements.
}
\centering
\begin{tabular}{p{\linewidth}}
    \toprule
    \textbf{Task Information}
    \begin{itemize}
        \item uid: bca8921a-ba38-1b97-4970-52276370192e
        \item Model: GPT-5.4 (High)
        \item Context Category: Fragmented Information \& Revisions
        \item Context Sub-Category: Personal Information Fragments
        \item Task: So, using the various notes surrounding cycling and bike trips, maintenance, etc. can you create a checklist of things to do or pack for an upcoming, five day bike trip? Specifically I want a list geared towards ensuring a successful trip. Every item on the list should be something that will enhance my safety and success on this trip and be justified by something in my notes, and I'd like you to reference the note that led you to create the list item.
    \end{itemize} \\
    \midrule
    \textbf{Context} \\
    \input{cases/context/case4} \\
    \bottomrule
\end{tabular}
\label{table:category1_case4}
\end{table}

\begin{table}
\begin{tabular}{p{\linewidth}}
\toprule
\textbf{Rubrics} \\
\input{cases/rubric/case4} \\
\midrule
\textbf{Model Generated Solution} \\
\input{cases/response/case4} \\
\midrule
\textbf{Discussion.} This task requires evidence-grounded planning from a fragmented personal archive---the model must identify past failure points scattered across ride logs, maintenance notes, and trip itineraries, then synthesize them into a forward-looking actionable checklist where every item is justified by a specific note. The model demonstrates strong cross-note evidence linking, correctly deriving the spare-tube and sealant items from Day 25's failure narrative, and appropriately grounding rain gear in multiple wet-weather ride logs. However, it over-generalizes from the archive: it pulls in trailer hardware, spoke specs, and axle bolt notes as checklist items, despite these being permanent-installation references rather than trip-packing evidence. It also misses the rubric requirement for a weather-forecast-monitoring item tied to Day 19's tornado warning, instead offering generic ``weather margin'' advice. This pattern---strong at extracting explicit failure-to-lesson mappings but weak at distinguishing actionable trip items from archival reference data---suggests the model lacks a robust filter for \textit{relevance scoping} when the source material mixes heterogeneous note types.
\\
\bottomrule
\end{tabular}
\end{table}
\newpage

%% file: cases/context/case4.tex
\textcolor{gray}{\textbf{(Part I)} [\textit{Trip-planning notes with rough itineraries, transport links, and logistics}]}

Erie Canal Trail (Jan 10, 2019): 4 days, 3 nights, shuttles from and ends in Lockport . . . 1. arrive Buffalo. Hotel . . . 6. Noble day 4, Lockport. 17 miles plus Niagra falls (22 miles = 39) . . . Cheapest flights on Mondays?

Swamp fox trip (Apr 16, 2018): Day 1: Amtrak to Charleston, SC (wanted Kingstree, SC, but no baggage = no bikes) . . . Day 2: 30-35 miles to trailhead. 1st campsite is 9 miles in . . . Day 5: Back to train station

. . . \textcolor{gray}{\textit{(Also includes Seaboard Railway trip with day-by-day mileage/camping, San Diego to Phoenix route map, and New River Trail lodging links.)}}

\textcolor{gray}{\textbf{(Part II)} [\textit{Ride logs from \#30daysofbiking with weather and rerouting problems}]}

Day 12 (Apr 15, 2019): Multi-modal, wet weather riding. Loaded up to ride to the train station and abandoned that plan on the way down the driveway. Rerouted to the bus station because who wants to arrive at the train soaking wet? . . . a soggy train ride after all.

Day 19 (Apr 19, 2019): . . . a line of tornado-spawning thunderstorms. Well, I hunkered down at the bottom of a parking deck stairwell waiting for a tornado warning to pass . . .

Day 25 (Apr 26, 2019): . . . my as-of-yet-unidentified, deflating, front tire did not interfere with my appointment. The spare tube I carry is sized for my much narrower wheelset, and probably wouldn't have worked in my 2.8" tire even if the tube hadn't developed several holes while kicking around in my toolkit for three years or so . . . ``Lazy negligence sounds so much meaner than bad luck.''

Day 29 (Dec 20, 2025): . . . my bike took a turn on the way home, and I best not ride it again until it's dealt with. My little bike should be up to the challenge, but it needs some tweaks, so I've been ignoring it all month.

. . . \textcolor{gray}{\textit{(Also includes Days 13--14 (rain, terrain surprises), Day 21 (unintended night ride at 10pm; near-miss with distracted driver in a 2016 entry), Day 22 (wrong tire choice), Day 27 (tubeless re-setup with missing valve-stem nut).)}}

\textcolor{gray}{\textbf{(Part III)} [\textit{Maintenance notes and standalone technical/spec notes}]}

Brenda's tire (Oct 20, 2016): Rhino Lite 26 erd 541 (261[259) Weinman 544 (263/261)

Shopping list (May 24, 2019): Shimano anti-rotation/non-turn washers: 5L 5R 7L 7R . . . WTB Ranger 26 x 3" TCS . . . Compass Rat Trap Pass 26 x 2.3 . . .

. . . \textcolor{gray}{\textit{(Also includes Spokes with spoke-length calculations for multiple hub/rim combos, Troll Rims with rim specs and spoke counts, Troll trailer bolts with axle threading notes, Serial Numbers, and xmas with gear wishlist including RunOff bags and bike rack.)}}

. . . \textcolor{gray}{\textit{(The full archive also contains ${\sim}$25 non-cycling notes: harmonica tablature, holiday film reviews, elder care resources, retirement task lists, job search notes, recipes, and miscellaneous personal items.)}}

%% file: cases/rubric/case4.tex
1. The response should format the output using standard checklist conventions. For example, it should use bullet points, checkboxes, or numbered lists for each item.

2. The response should accompany every item in the checklist with a justification based on the user's provided notes. For example, if it suggest waterproof clothing or a bag the justification might reference the user's various mentions of being caught in the rain.

3. The response should include a specific reference to the note (e.g., "Day 25" or "Spokes note") that informed the recommendation for every item in the checklist.

4. The response should include an item regarding refreshing tire sealant, justified by the "Day 25" note where "lazy negligence" regarding sealant caused a flat.

5. The response should include an item for carrying a spare tube specifically compatible with the tires, based on the "Day 25" failure where the user's spare was the wrong size.

6. The response should include an item for waterproof gear (clothing or bags), referencing the frequent rain disruptions mentioned in "Day 12", "13", "14", or "26".

7. The response should include an item for monitoring weather forecasts for severe conditions, referencing the "Day 19" encounter with tornado-spawning thunderstorms.

8. The response should include an item for identifying alternative transportation (like buses or trains), justified by the "Day 12" or "Day 25" notes where the user successfully used or needed a bus to reroute.

9. The response should not recommend carrying full-sized workshop tools that are not practical for a multi-day cycling journey. For example, the response should not recommend a "floor pump."

10. The response should not include items that only appear in shopping lists or notes that are not related to a successful or unsuccessful trip. For example, the response should not include spoke wrenches, spare spokes, or trailer bolts.

11. The response should not include items that are typically permanent installation parts, and are not typically needed for roadside repairs. For example, the response should not include "anti-rotation washers" or "adaptor washers" as trip checklist items just because they are in the "Shopping list" note.

12. The response should not use specific mentions of items or tools to justify generalized recommendations of tangential items. For example, the response should not use the "Seaboard Railway trip" Google Maps link to justify a requirement for "offline maps and guidebooks" because no note mentions success or failure of a trip based on offline maps or guidebooks.

13. The response should not add items to the list with items that are mentioned solely in the notes as general bicycle components, but are not linked to safety or trip success in the context. For example, the response should not mention bicycle bolts, wrenches, washers, or other hub-specific hardware.

14. The response should not include items that are not explicitly mentioned as part of the upcoming trip or in the context document as being relevant in the previous trip. For example, the response should not mention "Troll trailer bolts" because there is no mention of taking a trailer on past trips or the future trip.

15. The response should use each justification to explain how the item "enhances safety and success." For example, if the model mentions waterproof clothing, it might mention that this will keep the user drier if caught in the rain.

%% file: cases/response/case4.tex
Absolutely. I built this to reflect the failure points and near-failures that actually show up in your notes, rather than a generic touring list.

\#\# 5-day bike trip success checklist

\#\# Before you leave

$\square$ **Fix any known bike ``kinks'' before departure**
- **Why:** You noted that a bike you'd been ignoring suddenly became the weak point when you were close to finishing a goal.
- **From:** **``Day 29''** (bike ``took a turn''; little bike needed tweaks)

$\square$ **Refresh tubeless sealant and check every tire for slow leaks**
- **Why:** One tire got skipped because it was ``holding air,'' and that became the problem tire.
- **From:** **``Day 25''** (``should probably leave well enough alone'' led to failure), **``Day 27''** (tubeless re-setup)

$\square$ **Choose tires for the actual route and load**
- **Why:** Your notes show tire choice affecting speed, comfort, and suitability for surface type.
- **From:** **``Day 21''** (fattened tires for Umstead paths), **``Day 22''** (bigger tires felt like a slog), **``Day 27''** (plus tires passed the test ride), **``Shopping list''** (different tire options)\\
\bottomrule
\end{tabular}
\end{table}

\begin{table}
\begin{tabular}{p{\linewidth}}
\toprule
\textbf{Model Generated Solution (Continued)}

$\square$ **Do a loaded shakedown ride with the exact gear you'll carry**
- **Why:** Your notes repeatedly show that loaded riding changes the experience, and you explicitly test-rode after repairs to avoid getting stranded.
- **From:** **``Day 12''** (loaded ride), **``Day 14''** (fully loaded bike), **``Day 27''** (``take a little test run''), **``Surly''** (bike chosen for carrying load)

$\square$ **Check all cargo/trailer/axle-specific hardware if using it**
- **Why:** Your setup includes some nonstandard hardware, which is exactly the sort of thing that can derail a trip if overlooked.
- **From:** **``Troll trailer bolts,'' ``Shopping list''** (anti-rotation washers, adapter washer, strap anchors), **``xmas''** (rack note)

$\square$ **Verify train/bus/flight/shuttle bike policies before the trip**
- **Why:** You've already run into transportation constraints that changed the whole plan.
- **From:** **``Swamp fox trip''** (``no baggage = no bikes''), **``Day 12,'' ``Day 14''** (train/bus integration), **``Erie Canal Trail''** (flights/shuttles)

. . . \textcolor{gray}{\textit{(Omitted 5 additional ``Before you leave'' items: tire inflation check, daily mileage planning, terrain verification, route maps/bailout points, and weather/daylight margin.)}}

\#\# Repair kit to pack

$\square$ **Correct spare tube(s) for the tires actually on the bike**
- **Why:** You had a spare, but it was for the wrong wheel/tire setup.
- **From:** **``Day 25''**

$\square$ **Fresh spare tube(s), not old ones that have lived in the toolkit for years**
- **Why:** Your backup failed because it had aged into uselessness.
- **From:** **``Day 25''**

$\square$ **If tubeless: sealant, valve tool, and a quick check that valve hardware is installed/tight**
- **Why:** Sealant mattered, and a missing valve-stem nut caused extra grief.
- **From:** **``Day 27''**

$\square$ **A quick-reference note with your oddball specs**
- **Why:** You've kept notes on tire/rim dimensions, spoke lengths, and trailer/axle hardware for a reason.
- **From:** **``Brenda's tire,'' ``Spokes,'' ``Troll Rims,'' ``Troll trailer bolts''**

. . . \textcolor{gray}{\textit{(Omitted 3 additional repair items: tire levers, pump, and second flat-repair fallback.)}}

\#\# Weather, clothing, and keeping gear dry

$\square$ **Rain gear**
- **Why:** Rain is not a rare edge case in your riding notes; it's a recurring trip-shaping factor.
- **From:** **``Day 12,'' ``Day 13,'' ``Day 19,'' ``Day 26''**

$\square$ **Waterproof storage for clothes and electronics**
- **Why:** You've had soggy rides, wet transit legs, and repeated storm exposure.
- **From:** **``Day 12,'' ``Day 19''**; supported by **``xmas''** (Niteize RunOff bags)

$\square$ **A storm-shelter plan for each day's route**
- **Why:** You've already had to hide from severe weather and wait out dangerous conditions.
- **From:** **``Day 19''** (parking deck stairwell, waiting for warning/rain to pass), **``Day 12''** (rerouting due to weather)

. . . \textcolor{gray}{\textit{(Omitted dry clothing set item.)}}

\#\# Safety and ride execution

$\square$ **Front and rear lights**
- **Why:** Your notes show that even when you intend to finish in daylight, delays happen.
- **From:** **``Day 20''** (late-night ride), **``Day 21''** (rode home at 10 pm)

$\square$ **Helmet**
- **Why:** You had a serious near-miss with a distracted driver.
- **From:** **``Day 21.''** (Apr 21, 2016; driver ran red light while looking at phone)

$\square$ **A fallback transportation plan and enough flexibility to use it**
- **Why:** Some of your most successful recoveries came from rerouting to bus/train instead of forcing the original plan.
- **From:** **``Day 12,'' ``Day 25,'' ``Day 26''**

. . . \textcolor{gray}{\textit{(Omitted closing ``shortest version'' summary listing 7 must-not-skip items and offer to reformat.)}}

%% file: cases/case5.tex
\begin{table}[ht]
\caption{
A cross-corpus thematic alignment task over reader and writer blog archives in the romantasy space. Claude Opus 4.6 produces insightful feedback-loop hypotheses but violates citation-count constraints, misclassifies a hybrid source, and over-stretches title-to-concept mappings.
}
\centering
\begin{tabular}{p{\linewidth}}
    \toprule
    \textbf{Task Information}
    \begin{itemize}
        \item uid: c98f8847-5a19-505e-7bc9-2bc4679ab009
        \item Model: Claude Opus 4.6 (High)
        \item Context Category: Fragmented Information \& Revisions
        \item Context Sub-Category: Public Information Fragments
        \item Task: I’ve been doing some research into the tensions between reader criticism and what readers want vs writing advice and writer dialogue in the romantasy space. In the attached context there are a bunch of blog headlines/titles from some popular book blogs and popular writing blogs. Can you scan through these and identify some themes that pop up both in the writing advice space and the reader/reviewer space to identify possible areas of both tension and synergy to explore? I am interested in the way that marketing is driving publishing and the feedback loops that might be going on between readers and authors. Pull out 2-3 key avenues of exploration. Identify one explicit area of tension and one explicit area of synergy for each and provide at least 2 and no more than 4 citations from the context from each side (writer/reader) to support your suggestions. Please label any inference of feedback loops as hypothesis vs directly supported by textual snippets. I’ve been doing some research into the tensions between reader criticism and what readers want vs writing advice and writer dialogue in the romantasy space. In the attached context there are a bunch of blog headlines/titles from some popular book blogs and popular writing blogs. Can you scan through these and identify some themes that pop up both in the writing advice space and the reader/reviewer space to identify possible areas of both tension and synergy to explore? I am interested in the way that marketing is driving publishing and the feedback loops that might be going on between readers and authors. Pull out 2-3 key avenues of exploration. Identify one explicit area of tension and one explicit area of synergy for each and provide at least 2 and no more than 4 citations from the context from each side (writer/reader) to support your suggestions. Please label any inference of feedback loops as hypothesis vs directly supported by textual snippets.
    \end{itemize} \\
    \midrule
    \textbf{Context} \\
    \input{cases/context/case5} \\
    \bottomrule
\end{tabular}
\label{table:category1_case5}
\end{table}

\begin{table}
\begin{tabular}{p{\linewidth}}
\toprule
% \textbf{Rubrics} \\
\input{cases/rubric/case5} \\
\bottomrule
\end{tabular}
\end{table}

\begin{table}
\begin{tabular}{p{\linewidth}}
\toprule
\textbf{Model Generated Solution} \\
\input{cases/response/case5} \\
\midrule
\textbf{Discussion.} This task requires cross-corpus thematic alignment across ${\sim}$5 distinct blog archives spanning reader/reviewer and writer/advice discourse---the model must identify shared concerns, distinguish tension from synergy within each theme, maintain strict citation counts (2--4 per side), correctly classify each source's discourse role, and explicitly label inferential claims as hypothesis. Claude Opus 4.6 produces a sophisticated three-avenue analysis with well-labeled hypothesis markers and genuinely insightful feedback-loop reasoning. However, it commits three types of errors. First, a \textit{citation-count violation}: Avenue 1's tension lists 6 distinct writer-side titles (bundled with slashes), exceeding the 4-citation maximum---suggesting the model treats slash-grouped titles as single citations when structurally they are not. Second, a \textit{source misclassification}: Sierra St John---a romance novelist who blogs critiques and craft reflections---is placed under writer-side citations in Avenue 3, despite the rubric requiring her to be classified as a reader/reviewer source. This likely reflects the model's difficulty distinguishing hybrid voices who both write and critique. Third, several citations are \textit{tenuously matched} to their claimed concepts (e.g., an anticipated-sequels list cited as evidence of ``trope-driven discovery''). These are precision failures in a task that rewards analytical depth but penalizes structural non-compliance.
\\
\bottomrule
\end{tabular}
\end{table}
\newpage

%% file: cases/context/case5.tex
\textcolor{gray}{\textbf{(Part I)} [\textit{Reader/reviewer blogs: LitLifeBlog, RIME (Reading Is My Escape)}]}

LitLifeBlog titles include: ``Books With Similar Themes to Wuthering Heights: Gothic Romance, Trauma, and Obsession'' (Feb 3, 2026); ``BookTok Drama Explained: The Jamie Controversy, MM Romance, and Why `I Couldn't See Myself' Is Harmful'' (Jan 30, 2026); ``Are These Once-Hyped Books Still Worth Reading Today?'' (Sep 25, 2025); ``What Is Alchemised by SenLinYu? A Dark Fantasy Reimagined'' (Sep 24, 2025); ``What's the Maria Winter TikTok Drama All About?'' (Sep 1, 2025)

RIME blog titles include: ``Don't Miss the Hype: 7 Fantasy and Romantasy Books to Read Before Their Anticipated Sequels Drop in 2026''; ``The Wolf King'' and ``The Night Prince'' reviews; ``26 Most Anticipated Fantasy + Romantasy Novels Releasing in 2026''; ``Spoiler Review: Is Shield of Sparrows, the Latest Romantasy Sensation Sweeping Booktok and Bookstagram, right for your TBR?''; ``11 Books + Fantasy Series with Dragons to Read if You're Suffering From an Onyx Storm Book Hangover''; ``Book Review: Silver Elite and Why You Might Hate This New Dystopian Romance Novel Taking Over Booktok and Bookstagram''; ``The Rose Bargain is a Fun YA Romantasy Romp: The Bachelor, but Make it Regency London + with Fae''

. . . \textcolor{gray}{\textit{(RIME also includes author interviews (``Storyteller Sessions''), seasonal reading challenges, bookish gift guides, desktop wallpapers, and Flodesk email marketing posts. Additional review titles cover ``Hekate,'' ``A Language of Dragons,'' ``The Raven Scholar,'' ``Blood Over Bright Haven,'' and others across fantasy, historical fiction, and romantasy.)}}

%% file: cases/rubric/case5.tex
\textbf{Context (Continued)}

\textcolor{gray}{\textbf{(Part II)} [\textit{Writing advice blogs: Cassandra Wills (Fiction Writing Made Easy), Tessa Malone}]}

Cassandra Wills blog categories: Mindset, Genre, Theme, Characters, Point of View, Structure, Setting, Editing, Success Stories, Publishing. Titles include: ``3 Signs Your Novel Doesn't Need a Prologue''; ``Scene Structure Made Easy: The 5 Essential Elements Every Scene Needs''; ``5 Tips For Crafting Morally Gray Characters Readers Love''; ``5 Secrets to Writing Dialogue That Sounds Natural''; ``How to Write Romantasy Power Couples Readers Obsess Over With Tessa Malone''; ``How to Build an Author Platform (Even if You're an Introvert) with Casey Lee''; ``What Agents \& Readers Want: How to Write a Story That Works''; ``How to Revise Your Novel Like a Pro''; multiple Student Spotlights

Tessa Malone titles include: ``10 Worst Tropes in Fiction''; ``The Worst Romance Tropes in Fiction''; ``10 Worst Tropes in Fantasy Genre Fiction''; ``TROPES VS CLICHES''; ``10 BEST ROMANCE TROPES IN FICTION''; ``10 Best Tropes in Fantasy Fiction''; ``How to Write Romantasy - What IS Romantasy?''; ``How to Write a Healthy Romance''; ``How to Write a Sex Scene''; ``Character Development: Morally Grey Characters''; ``How to Stay Motivated While Writing Your Book''; ``10 Biggest Mistakes Writers Make''; ``Get More Readers: How to Host Giveaways''; ``How to Make an Author Newsletter''; ``How to Write Toxic Relationships''

. . . \textcolor{gray}{\textit{(Tessa Malone's blog spans ${\sim}$80 posts covering trope rankings across all subgenres, craft advice on dialogue/pacing/structure, publishing Q\&As, and reading pet peeves. She is a bestselling author with a Penguin Random House contract.)}}

\textcolor{gray}{\textbf{(Part III)} [\textit{Romance critique blog: Sierra St John}]}

Sierra St John describes herself as a romance novelist in Denver. Blog posts critique specific romance films and series: ``Wish-fulfillment at its worst: A critique of The Idea of You''; ``A critique of The Commoner''; ``I watched all of Netflix's new 2024 Christmas romcoms (so you don't have to)''; ``Attention! This is how to achieve romantic tension''; ``Why I DIY self-publish + My cover art dilemma''; ``7 reasons why we love those terrible, cheesy, wonderful Christmas romance movies''

. . . \textcolor{gray}{\textit{(Sierra's blog also lists links to other romance writers and covers topics including self-publishing, romantic tension craft, and mother-daughter conflict in romance narratives.)}}\\
\midrule
\textbf{Rubrics}

1. The response should identify exactly 2 or 3 avenues of exploration concerning the relationship between reader criticism/wants and writing advice.

2. The response should include one specific "explicit area of tension" and one specific 'explicit area of synergy' for each avenue identified.

3. The response should provide at least 2 and no more than 4 citations from the writers (Tessa Malone, Cassandra Wills) and from the reviewers/readers (Sierra St. John, LitLifeBlog, RIME) for every area of tension and synergy discussed.

4. The response should label Sierra St John, LitLifeBlog, and RIME as reviewer/reader sources.

5. The response should label Tessa Malone and Cassandra Wills as writer/advice sources.

6. The response should label any inference regarding "feedback loops" that is not mentioned directly in textual snippets as "hypothesis."

7. The response should not hallucinate the internal content or specific arguments of the blog articles that are not explicitly stated in the provided titles or summaries (for example, assuming a specific reason for "The Jamie Controversy" beyond the provided headline).

8. The response should ensure that cited titles explicitly mention or paraphrase the concept being supported. For example, it should use a citation about "Anticipated Sequels" to support demand/hype.

9. The response should include one avenue that discusses marketing/publishing feedback loops. For example, it could discuss the synergy between an author's "TikTok Shop Gift Guide" and readers seeking "Hype" books.

10. The response should be structured with distinct headings for each avenue and their subsections for tension, synergy, citations, and feedback-loop hypothesis.

11. The response should label any inference regarding "feedback loops" that is directly mentioned in textual snippets as "directly supported by textual snippets."

%% file: cases/response/case5.tex
\# Tensions and Synergies Between Reader Criticism and Writing Advice in the Romantasy Space

\#\# Avenue 1: The Trope Economy --- Tropes as Marketing Language vs.\ Creative Prescription

**Area of Tension: Prescriptive trope hierarchies (writer side) vs.\ trope-driven discovery and consumption (reader side)**

*Writer-side citations:*
- Tessa Malone: ``10 Worst Tropes in Fiction'' / ``10 Worst Tropes in Fantasy Genre Fiction'' / ``The Worst Romance Tropes in Fiction''
- Tessa Malone: ``TROPES VS CLICHES''
- Tessa Malone: ``10 BEST ROMANCE TROPES IN FICTION'' / ``10 Best Tropes in Fantasy Fiction''

*Reader-side citations:*
- RIME: ``The Rose Bargain is a Fun YA Romantasy Romp: The Bachelor, but Make it Regency London + with Fae''
- LitLifeBlog: ``Books With Similar Themes to Wuthering Heights: Gothic Romance, Trauma, and Obsession''

**Area of Synergy: Tropes as shared genre vocabulary enabling both discoverability and craft**

*Writer-side:* Cassandra Wills: ``How to Write Romantasy Power Couples Readers Obsess Over With Tessa Malone''; Tessa Malone: ``How to Write Romantasy - What IS Romantasy?''

*Reader-side:* RIME: ``Spoiler Review: Is Shield of Sparrows \ldots\ right for your TBR?''; RIME: ``11 Books + Fantasy Series with Dragons to Read if You're Suffering From an Onyx Storm Book Hangover''

**Hypothesis --- Feedback Loop:** The sheer volume of trope-ranking content on the writer-advice side likely reflects *demand from writers who are themselves responding to trope-driven reader discovery on BookTok*. . . . *This is inferential --- no single source explicitly describes this cycle.*

\#\# Avenue 2: The Hype-Quality Gap --- BookTok Marketing Power vs.\ Craft and Longevity

**Area of Tension:** *Reader-side:* LitLifeBlog: ``Are These Once-Hyped Books Still Worth Reading Today?''; RIME: ``Book Review: Silver Elite and Why You Might Hate This New Dystopian Romance Novel Taking Over Booktok and Bookstagram''; LitLifeBlog: ``What's the Maria Winter TikTok Drama All About?''

*Writer-side:* Tessa Malone: ``Get More Readers: How to Host Giveaways''; ``How to Make an Author Newsletter''; Cassandra Wills: ``Flodesk's New Pricing Structure and Why It's Still the Best Email Marketing Platform for Authors''

**Area of Synergy: Community-driven discovery valued on both sides**

*Writer-side:* Cassandra Wills: ``How to Build an Author Platform (Even if You're an Introvert) with Casey Lee''; ``What Agents \& Readers Want: How to Write a Story That Works''

*Reader-side:* RIME: ``Storyteller Sessions: Meet Author Rachel Howzell Hall \ldots''; ``Storyteller Sessions: Meet Author Kady Ambrose \ldots''

**Hypothesis --- Feedback Loop:** Reader-side skepticism toward hype may itself be a market correction in response to writer-side marketing intensification. *This is entirely inferential.*

\#\# Avenue 3: The ``Healthy Romance'' Prescription vs.\ Reader Appetite for Complexity and Darkness

**Area of Tension:**

*Writer-side:* Tessa Malone: ``How to Write a Healthy Romance'' --- ``we're sick of douchebag love stories''; ``Rant Book Review: How NOT to Write a Second Chance Romance''; Sierra St John: ``Wish-fulfillment at its worst: A critique of The Idea of You''

. . . \textcolor{gray}{\textit{(Closing summary table maps all three avenues to their tension, synergy, and feedback loop status (all ``Hypothesized'').)}}

%% file: cases/case6.tex
\begin{table}[ht]
\caption{
A revision-aware canon reconstruction task over ${\sim}$2,700 lines of interleaved fantasy world-building notes and iterative chapter planning. Claude Opus 4.6 identifies most author revisions but misclassifies organic backstory elaborations as changes, omits three subtle setting/mechanic revisions embedded in incremental turns, and defaults to ``no reasoning given'' when user motivations are conveyed implicitly.
}
\centering
\begin{tabular}{p{\linewidth}}
    \toprule
    \textbf{Task Information}
    \begin{itemize}
        \item uid: f00fca7a-f206-fa4a-d548-4db4ef42a36e
        \item Model: Claude Opus 4.6 (High)
        \item Context Category: Fragmented Information \& Revisions
        \item Context Sub-Category: Creation \& Revision Histories
        \item Task: What were changes I made to Eastern characters, races, or locations? Include a brief description of why the changes were made (if available).
    \end{itemize} \\
    \midrule
    \textbf{Context} \\
    \input{cases/context/case6} \\
    \bottomrule
\end{tabular}
\label{table:category1_case6}
\end{table}

\begin{table}
\begin{tabular}{p{\linewidth}}
\toprule
\textbf{Rubrics} \\
\input{cases/rubric/case6} \\
\bottomrule
\end{tabular}
\end{table}

\begin{table}
\begin{tabular}{p{\linewidth}}
\toprule
\textbf{Model Generated Solution (Continued)} \\

\input{cases/response/case6} \\
\midrule
\textbf{Discussion.} This task demands revision-aware canon reconstruction across ${\sim}$2,700 lines of interleaved world-building notes, chapter skeletons, and iterative backstory updates---the model must distinguish genuine author revisions from organic elaborations that never contradicted prior canon, and recover stated reasoning or note its absence. Claude Opus 4.6 correctly identifies 7 changes but commits three categories of errors. First, a \textit{change-vs-addition misclassification}: Galidurn's progressively expanded backstory is listed as a ``change'' despite never overwriting prior canon. Second, \textit{omission of three genuine changes} embedded in incremental turns: the castle of Revalon refined to a lived-in setting, reincarnation expanded to all sentient races, and the Llatherosi clarified as the most openly humorous Eastern people. Third, \textit{reasoning attribution failures}: for the kidnapping relocation and the Curator's antagonist reveal, the user conveyed motivations implicitly through design logic, yet the model defaults to reporting ``no reasoning was given.''
\\
\bottomrule
\end{tabular}
\end{table}
\newpage

%% file: cases/context/case6.tex
\textcolor{gray}{\textbf{(Part I)} [\textit{Race codex and spiritual systems}]}

Humans -- Children of Illia

Origins \& Spiritual Nature

Humans were created by Illia, the Mother-Goddess, who became obsessively protective of them. They are the only race whose souls are bound to the world, capable of reincarnation and Incarnation---a divine gift that briefly restores memories of past lives.

\ldots

Vulkari: The Forgotten Kin

Core Identity

The Vulkari are a once-human race twisted by Vitrius or his mortal proxies to serve as hunters of dissenters and rebels. Wolf-like in form, primal in instinct, they were designed to obey---then abandoned. Now, they wander the world as memory-haunted beasts, half-feral and half-divine. They remain spiritually tethered to Illia's reincarnation cycle, suffering a tragic distortion of Incarnation.

. . . \textcolor{gray}{\textit{(Part~I also includes entries for Rathani, Valekin, Valekin Cousins, Cylvani, Nahara'jin, Serathi, Thral'kor, Undead, and Dragons; a Cycle of Souls doctrine; and a Revalon kingdom profile.)}}

\textcolor{gray}{\textbf{(Part II)} [\textit{World guide: cosmology, history, geography, factions, magic}]}

God Wars World Guide

1. Pantheon \& Cosmology

The world of the God Wars is shaped by three god-like beings:

Illia -- The Mother:
- Domain: Humanity, reincarnation, light, memory.
\ldots

Vitrius -- The God of Will / The Blood Father:
- Domain: Chaos, ambition, monsters, domination.
\ldots
- Current state: Imprisoned after the Third God War, but his influence leaks through Tears in Reality.

. . . \textcolor{gray}{\textit{(Sections 2--11 cover a 9-God-War timeline through era 1200\,9e; geography across two continents and a volcanic archipelago; factions and nations; expanded race profiles; religion/ritual; magic systems (Anima, Branded, forbidden Tear-magic) and technology; notable figures; prophecy of a Ninth God War; glossary; and a mini-storyline of the First Flameguard.)}}

\textcolor{gray}{\textbf{(Part III)} [\textit{Novel chapter planning and character development}]}

Ch 1. Addicus is called to the tavern to break a brawl, takes over Farthings post momentarily and witnesses Renning being overwhelmed with gratitude and astonishment having helped a trapped spirit move on \ldots Addicus' patrol takes him by the training grounds where he here's the voice of brandt, who is berating a young recruit, addicus intervenes, is humiliated by brandt in front of everyone, an older guard gives him a nod of respect

Ch 2- illia awakens from a nightmare/vision. Her father enters to try to coax her out of bed, where she has largely remained for several days. \ldots She hears, ``My princess!'', from below and sees her former bodyguard and friend, turned branded Galidurn who has returned from battling diabolics. She smiles, waves, and opens her door to make her way down.

Galidurn provided the fun uncle role to her father's loving but distant, still grieving, and immersed in ruling

Galidurn is also the only one who would speak about her mother

. . . \textcolor{gray}{\textit{(The conversation includes AI-structured Ch\,1--3 beat-by-beat skeletons with pacing feedback, and iterative character backstory development for Addicus, Illiadrel, Galidurn, Renning, Brandt, and Sir Anders.)}}

%% file: cases/rubric/case6.tex
1. The response should identify the specific name change of the character "Theric" to "Galidurn."

2. The response should state that the user's reason for the name change was a personal preference, according to the quote "I like Galidurn more than Theric."

3. The response should not attribute the reasoning that "Galidurn" has "more gravitas and mythic resonance" to the user, as this was a suggestion provided by the AI assistant, not a reason stated by the user.

4. The response should state that sarcasm is Illiadrel's personal trait, while fatalistic humor is the Llatherosi cultural trait.

5. The response should note that Princess Illiadrel’s kidnapping was moved from Book 2 to Book 1.

6. The response should state that the reason for moving the kidnapping was to adjust the pacing and structure of the trilogy.

7. The response should identify the role change for the Curator of the Illian Archives from a simple scholar/advisor to a major political antagonist.

8. The response should state that the reason for the Curator's role change was to add political complexity and a primary villain to the plot (as indicated by the user's description of him orchestrating the King's assassination).

9. The response should state that the previously unnamed King of Revalon was named "Bannon."

10. The response should state that no reason was provided by the user for the change of the king's name to Bannon.

11. The response should not categorize the addition of Galidurn's backstory as a change unless the context shows it replaced a different, conflicting backstory.

12. The response should identify that the user clarified the Llatherosi people to be the "more openly humorous people" in the East.

13. The response should not attribute the AI assistant's interpretations of character attributes to the user. For example, it should not state that the user referred to Bannon's name as "dignified."

14. The response should mention the refinement of the Curator's power, clarifying that he holds accumulated writings rather than being an oracle (as per the user's correction "No, they literally have writings... they are supposed to be a voice of our past").

15. The response should note that the changes to the Curator's character were made to define the limits of his power without making him omniscient, as per the quote "They only know what writings were preserved after godsfalls."

16. The response should note the refinement of the castle of Revalon from a static backdrop to brimming with courtiers, servants, guards, and staff, referred to by the user as "lively and lived in."

17. The response should note that no explicit reason was provided by the user for the changes made to the castle of Revalon.

18. The response should not state that Illiadrel being described as "smart, sarcastic, and funny" is a "change," since that was her original characterization.

19. The response should only list changes related to Eastern characters, races, or locations, such as Revalon, Princess Illiadrel, and Galidurn.

20. The response should note that humans being the only race whose souls are bound to the world via reincarnation was changed to all sentient races experiencing reincarnation.

21. The response should state that no explicit reason was given by the user for changing which races experience reincarnation.\\
\midrule
\textbf{Model Generated Solution}

Here are the changes made to Eastern characters, races, or locations throughout the conversation, along with reasoning where available.

---

\#\#\# Character Changes

**1. Theric $\to$ Galidurn (Name Change)**\\
The Branded knight who serves as Illiadrel's former bodyguard was originally named **Theric**. The user changed the name to **Galidurn** with the simple reasoning: *``I like Galidurn more than Theric.''*

%% file: cases/response/case6.tex
**2. Galidurn's Role (Progressively Expanded)**\\
Galidurn was initially described only as Illiadrel's former personal guard turned Branded. His role was expanded across several updates:\\
- Added as the **``fun uncle'' figure** to Illiadrel, filling the emotional gap left by her grieving, duty-consumed father Bannon.\\
- Added as **King Bannon's lifelong best friend** \ldots\\
- Made **the only person who would speak to Illiadrel about her mother** \ldots\\
No single reason was given; these additions built organically as the user developed the emotional core of the royal family.

**3. King of Revalon Named Bannon**\\
The king was unnamed throughout early development. After being offered suggestions, the user chose **Bannon** on their own without selecting from the provided list.

---

\#\#\# Cultural/Racial Changes

**4. Llatherosi Humor Trait Refined**\\
\ldots\\
- **Fatalistic humor** $\to$ remains a core Llatherosi cultural trait.\\
- **Sarcasm** $\to$ is Illiadrel's **personal** trait, not cultural.

The reasoning was precision: the user wanted Illiadrel's voice to be distinctly hers, not entirely inherited from her culture.

---

\#\#\# Plot/Structural Changes

**5. Illiadrel's Kidnapping Moved to Book One**\\
The AI initially assumed the Serathi kidnapping of Illiadrel occurred in **Book Two**. The user corrected this: *``Illiadrel is kidnapped in Book One.''* No additional reasoning was given beyond factual correction.

**6. The Curator's True Nature Revealed**\\
\ldots The user then revealed that the Curator:\\
- **Orchestrates King Bannon's assassination in Book Two.**\\
- **Believes Illiadrel and her mother are curses on the kingdom.**\\
\ldots\\
No explicit reasoning was given for the change; this appears to have been a planned reveal rather than a revision.

**7. Curator's Knowledge Base Clarified**\\
\ldots The final canon is that the Curator holds **physical, written records** accumulated across Ages, limited only by what was **preserved through Godsfalls**---not by any supernatural mechanism.

. . . \textcolor{gray}{\textit{(The response lists 7 changes total---3 character, 1 cultural/racial, and 3 plot/structural. Each entry includes a description of the change and, where available, the user's stated reasoning.)}}

%% file: cases/case7.tex
\begin{table}[ht]
\caption{
A multi-source temporal alignment task requiring causal chaining across a StarCraft~II replay log (${\sim}$1,550 events) and a caster transcript (${\sim}$180 lines). Seed 2.0 Pro correctly identifies the triggering upgrade and computes completion timing, but fails to name the formal \texttt{UnloadTargetMedivac} event and to explicitly link the caster's retraction quote to the preceding ``ling speed'' observation.
}
\centering
\begin{tabular}{p{\linewidth}}
    \toprule
    \textbf{Task Information}
    \begin{itemize}
        \item uid: 13bad95a-fc33-b01c-ae45-21af3841392e
        \item Model: Seed 2.0 Pro (High)
        \item Context Category: Behavioral Records \& Activity Trails
        \item Context Sub-Category: Game Logs
        \item Task: In addition to the timestamps in the Game Summary, here is some information linking certain parts of the Transcript to the game clock:

0:32 - "just just do a really normal build" [the game clock has already been counting for a few ticks as the transcript starts its first sentence]

2:09 - "quite doggedly good in his ZVP late game"

3:09 - "active with your metavac and keep diving"

4:09 - "with the queens there should be okay"

4:14 - "And he does go ling speed"

5:09 - "quick fourth hatchery and second gas, I should say"

5:31 - "dive, dive, dive"

6:09 - "getting so much value if he gets that fourth base"

7:09 - "Drilling claws is on the way"

8:09 - "No vehicle plating either on the armory"

9:09 - "The other ling bane counterattacks"

10:09 - "Got to watch out for these widow mines"

11:09 - "The bailings come in, but he picks up at the last second"

11:54 - "perhaps what I'm most impressed by" [end of match, although clock counts for a few ticks more]

When the videocaster says, "Everything I said in that analysis, you can throw it out, guys," which action (in the Game Summary) on the part of ZergCommander caused the videocaster to change his narrative? Had the player fully implemented this change (assume it takes 110 seconds from the time it starts) by the time the videocaster says, "dive, dive, dive" and the Terran player unloads his medivacs to attack the Zerg?
    \end{itemize} \\
    \midrule
    \textbf{Context} \\
    \input{cases/context/case7} \\
    \bottomrule
\end{tabular}
\label{table:category1_case7}
\end{table}

\begin{table}
\begin{tabular}{p{\linewidth}}
\toprule
% \textbf{Rubrics} \\
\input{cases/rubric/case7} \\
\bottomrule
\end{tabular}
\end{table}

\begin{table}
\begin{tabular}{p{\linewidth}}
\toprule
% \textbf{Model Generated Solution (Continued)} \\
\input{cases/response/case7} \\
\midrule
\textbf{Discussion.} This task requires multi-source temporal alignment across a ${\sim}$1,550-event game replay log and a ${\sim}$180-line caster transcript---the model must chain commentary quotes to in-game upgrade events, compute upgrade completion times, and determine whether a strategic change was ``live'' when a subsequent attack landed. Seed 2.0 Pro correctly identifies EvolveMetabolicBoost at 04:05 as the trigger, performs the 110-second calculation to 05:55, and concludes the upgrade was not complete by the dive at 05:31/05:34. However, it commits two types of errors: first, a \textit{missing event-name grounding}---it refers to ``Terran unloading medivacs to initiate the attack at 05:34'' without ever naming the specific \texttt{UnloadTargetMedivac} event from the Game Summary, failing to anchor the natural-language description to its formal log entry. Second, a \textit{causal-link omission}---it notes the caster says ``And he does go ling speed'' at 4:14 but never explicitly connects the subsequent ``Everything I said in that analysis, you can throw it out'' quote as a direct consequence of that observation, leaving the core causal chain between the two quotes unlinked.
\\
\bottomrule
\end{tabular}
\end{table}
\newpage

%% file: cases/context/case7.tex
\textcolor{gray}{\textbf{(Part I)} [\textit{Game Summary -- replay event log}]}

Summary of the replay:

Map: Ultralove

Duration: 11:59

Players:

ZergCommander as Zerg (Loss)

TerranOperator as Terran (Win)

Events:

At 00:00, TerranOperator used TrainSCV

At 00:00, ZergCommander used MorphDrone

At 00:02, ZergCommander used ScanMove

\ldots

At 00:35, ZergCommander used BuildHatchery

At 00:38, TerranOperator used BuildBarracks

\ldots

At 01:27, TerranOperator used TrainReaper

At 01:27, TerranOperator used UpgradeToOrbitalCommand

\ldots

%% file: cases/rubric/case7.tex
\textbf{Context (Continued)}

At 02:10, TerranOperator used BuildFactory

At 02:10, ZergCommander used BuildHatchery

\ldots

At 02:35, ZergCommander used MorphZergling

\ldots

At 03:03, TerranOperator used BuildHellion

\ldots

At 04:05, ZergCommander used EvolveMetabolicBoost

At 04:05, ZergCommander used SpawnLarva

\ldots

At 04:43, TerranOperator used TrainMedivac

\ldots

At 05:34, TerranOperator used UnloadTargetMedivac

At 05:34, TerranOperator used Attack

\ldots

At 11:57, TerranOperator used Attack

. . . \textcolor{gray}{\textit{(The Game Summary lists ${\sim}$1,550 timestamped events from 00:00 to 11:57, covering build orders, unit production, upgrades, combat actions, and economy management for both players.)}}

\textcolor{gray}{\textbf{(Part II)} [\textit{Caster transcript}]}

And it's like just just do a really normal build and and yet obviously he can't proxy hatch again \ldots So I think TerranOperator kind of says, ``Okay, let's calm things down, go for a one rax expand and just chill.''

. . . \textcolor{gray}{\textit{(Caster discusses TerranOperator's expected Hellion/Banshee style and advises standard defensive play.)}}

I think your goal as TerranOperator is, you know, with this sort of roach build, the Zerg plays low on queens because they're they're building roaches, which are very expensive. Whereas with a Zergling build, you're more mineral focused. You can afford more queens and you need them. But when you play roaches, you often stop at five. Very rarely do you play more than six queens. And that means if these marines can dart in, snipe a queen, pick up, boost somewhere else, and you just start multi-dropping, sometimes Zerg, if Zerg makes any mistakes, they can just get overrun.

\ldots Now, interestingly, we've got a lot of slow zerglings building. Eight slow zerglings. \ldots I I normally do not expect slow zurglings to come out, but they're not bad at kind of stopping units from running past the queens. On their own, they're useless, but with the queens there should be okay. It's actually six queens into seven queens. And he does go ling speed. Okay, mind games. Mind games. As long as ling speed is ready before the stim marines arrive, it's fine. So he is going to play a heavy queen style.

So everything I said that analysis, you can throw it out, guys. Meteavax there. Evo chamber comes up cuz lings are going to have a lot more mobility. The the multi drops aren't going to be as effective.

\ldots Oh, he's already hitting. He's already going to hit, guys. He's got four Heliens here. Looks like one's inside the medevac. He's going to go dive dive dive.

. . . \textcolor{gray}{\textit{(Transcript continues through hellbat dive, queen micro, widow mine trades, parasitic bombs, and TerranOperator's game-winning engagement. Total ${\sim}$180 lines.)}}\\
\midrule
\textbf{Rubrics}

1. The response should identify "EvolveMetabolicBoost" in the Game Summary as the action on the part of the ZergCommander that caused the videocaster to change their narrative.

2. The response should identify 04:05 as the time in the game summary where "EvolveMetabolicBoost" was activated by ZergCommander.

3. The response should identify 5:55 (or 05:55) as the time when the change in strategy started by ZergCommander with 'EvolveMetabolicBoost' was fully implemented.

4. The response should identify 05:55 as the specific time which is 110 seconds after "EvolveMetabolicBoost" was activated.

5. The response should identify that the videocaster says "dive, dive, dive" at 5:31 (or 05:31).

6. The response should identify that TerranOperator unloading their medivacs to attack the Zerg refers to an UnloadTargetMedivac event in the Game Summary.

%% file: cases/response/case7.tex
\textbf{Rubrics (Continued)}

7. The response should identify that the "UnloadTargetMedivac" event in which the Terran player unloads his medivacs to attack the Zerg character happened at 5:34 (or 05:34) in the game summary.

8. The response should give a definitive "no" to the question of whether the ZergCommander player had fully implemented the change in strategy initiated by 'EvolveMetabolicBoost' by the time the videocaster says "dive, dive, dive" and the TerranOperator player had unloaded their medivacs.

9. The response should indicate that the videocaster saying "Everything I said in that analysis, you can throw it out, guys," is a direct result of the observation they made where they said "And he does go ling speed".

10. The response should identify that the quote from the videocaster "And he does go ling speed" occurs at 4:14 (or 04:14)

11. The response should calculate the completion time for the upgrade initiated by ZergCommander using 'EvolveMetabolicBoost' by adding 110 seconds to its start time of 04:05.

12. The response should determine whether the upgrade initiated by ZergCommanders use of 'EvolveMetabolicBoost' at 04:05 had been fully implemented before the "dive, dive, dive" comment  made by the videocaster at 05:31 by comparing their timestamps.

13. The response should not identify "MorphZergling" at 02:03 as the cause of the videocaster saying "Everything I said in that analysis, you can throw it out, guys," as that comment instead specifically follows the videocasters observation of the speed upgrade research by the ZergCommander player

14. The response should determine whether the upgrade initiated by ZergCommanders use of 'EvolveMetabolicBoost'  had been fully implemented before the Terran player unloaded their medivacs to attack the Zerg by comparing their timestamps in the game summary.

15. The response should not assert that events occurred at particular times unless this is supported by the Game Summary or other documents provided in the conversation.\\
\midrule
\textbf{Model Generated Solution}

\#\#\# Trigger for the caster's narrative shift

The videocaster initially predicted ZergCommander would play a roach-focused build \ldots The action that invalidated this analysis, prompting him to discard his earlier take, was **ZergCommander starting the *EvolveMetabolicBoost* (Zergling movement speed upgrade) at 04:05 (wall-clock time)** per the Game Summary. \ldots The caster explicitly notes this shift when he says ``And he does go ling speed'' at the linked transcript timestamp 4:14, immediately before stating his earlier analysis.

---

\#\#\# Implementation status at the ``dive, dive, dive'' attack

\ldots

1.  The *EvolveMetabolicBoost* upgrade (the marker of Zerg's build shift, started at 04:05.

2.  110 seconds after this start time is **05:55.

3.  The caster says ``dive, dive, dive'' at the linked transcript timestamp 5:31, with Terran unloading medivacs to initiate the attack at 05:34 per the Game Summary.

This attack occurs **21-24 seconds before the 110-second implementation window ends**, so the build change (including the Zergling speed upgrade had not been fully completed when the Terran dive attack landed. \ldots

. . . \textcolor{gray}{\textit{(The response is structured in two sections: identification of EvolveMetabolicBoost at 04:05 as the trigger, and a timestamp calculation concluding the upgrade was not complete by the 05:31/05:34 attack.)}}

%% file: cases/case8.tex
\begin{table}[ht]
\caption{
An adversarial hypothesis evaluation task over ${\sim}$361 Amazon purchase records, requiring anomaly detection, consumption baseline construction, and epistemic restraint when competing explanations are both plausible. Seed 2.0 Pro builds a compelling guest-preparation narrative from circumstantial clues but over-commits to it, failing to acknowledge the ambiguity the rubric demands.
}
\centering
\begin{tabular}{p{\linewidth}}
    \toprule
    \textbf{Task Information}
    \begin{itemize}
        \item uid: 3cd2d0d6-42e4-1e09-8bbf-f0c32915a48f
        \item Model: Seed 2.0 Pro (High)
        \item Context Category: Behavioral Records \& Activity Trails
        \item Context Sub-Category: Digital Footprints \& Daily-Life Records
        \item Task: Examine the items bought in the month leading up to the January 13-17 home decor purchases. Could all these purchases suggest the user was redecorating a room or preparing for a guest? Based on these items bought, is the redecorating explanation or visitor better supported. Support your conclusion by identifying specific clues.
    \end{itemize} \\
    \midrule
    \textbf{Context} \\
    \input{cases/context/case8} \\
    \bottomrule
\end{tabular}
\label{table:category1_case8}
\end{table}

\begin{table}
\begin{tabular}{p{\linewidth}}
\toprule
% \textbf{Rubrics} \\
\input{cases/rubric/case8} \\
\bottomrule
\end{tabular}
\end{table}

\begin{table}
\begin{tabular}{p{\linewidth}}
\toprule
\textbf{Model Generated Solution (Continued)} \\
\input{cases/response/case8} \\
\midrule
\textbf{Discussion.} This task requires anomaly detection across ${\sim}$361 purchase records spanning two years, followed by adversarial hypothesis evaluation: the model must first establish a consumption baseline, identify the Jan 13--17 home decor burst as behavioral deviation, then evaluate whether pre-cluster purchases (voltage converter, walkie talkies, winter gear, tablecloth/placemats) better support redecorating or guest preparation---crucially, the rubric demands acknowledging that these clues are \textit{ambiguous} and do not definitively favor either explanation. Seed 2.0 Pro performs strong baseline separation and builds a detailed, well-structured argument linking the Japan-specific voltage converter and children's toys to an inferred visiting family. However, it commits a critical \textit{over-commitment error}: it declares the guest hypothesis ``strongly supported'' and the voltage converter ``the most definitive clue'' whose ``only logical use'' is for a Japanese visitor, when the rubric requires concluding that the evidence does \textit{not} clearly favor one hypothesis over the other. This reflects a broader tendency of the model to construct maximally coherent narratives from circumstantial evidence rather than maintaining appropriate epistemic uncertainty---precisely the failure mode the task is designed to probe.
\\
\bottomrule
\end{tabular}
\end{table}
\newpage

%% file: cases/context/case8.tex
\textcolor{gray}{\textbf{(Part I)} [\textit{Order history -- routine consumption baseline}]}

Order History,,

Order Date,Product Name,Unit Price

2024-01-05T15:28:43Z,"RAW SUGAR Sensitive Skin Simply Body Wash - Green Tea + Cucumber + Aloe Vera, Moisturizing \& Brightening Bath \& Shower Gel, Sulfate-Free, Paraben-Free \& Vegan (Pack of 3)",27.99

2024-01-08T10:49:10Z,"Burt's Bees Refreshing Facial Towelettes With Cucumber and Mint \ldots 30 ct. Package (3-Pack)",19.97

\ldots

2024-01-10T21:16:54Z,"Charmin Ultra Strong Clean Touch Toilet Paper, 24 Family Mega Rolls = 123 Regular Rolls",35.49

\ldots

2024-02-29T05:38:28Z,"MegaFood Women's 40+ One Daily Multivitamin for Women \ldots 60 Tabs",37.99

. . . \textcolor{gray}{\textit{(Regular restocking of personal care, supplements, household staples, and groceries recurs throughout the 2-year history---body wash ${\sim}$monthly, toilet paper ${\sim}$bimonthly, MegaFood supplements ${\sim}$quarterly, Mrs.\ Meyer's hand soap ${\sim}$bimonthly, etc.)}}

\textcolor{gray}{\textbf{(Part II)} [\textit{Holiday season and anomalous pre-cluster purchases}]}

2024-11-29T15:41:15Z,"Swarovski Annual Edition 2024 Snowflake Ornament \ldots",49.99

2024-12-11T21:01:24Z,Hasbro Gaming Simon Micro Series Game,9.84

2024-12-11T21:01:24Z,ROKR 3D Puzzles for Adults \ldots,32.99

2024-12-11T21:01:24Z,"The Japanese Grill: From Classic Yakitori to Steak, Seafood, and Vegetables [A Cookbook]",16.99

\ldots

2024-12-16T22:38:20Z,SaphiRose Hooded Rain Poncho Waterproof Raincoat Jacket for Men Women Adults (Grey),14.99

2024-12-16T22:38:20Z,"Karecel Rechargeable Hand Warmer, 10000mAh \ldots",9.99

2024-12-17T23:26:26.026Z,"Retevis RT628 Walkie Talkies for Kids,Toys Gifts for 3-14 Years Old Boys Girls \ldots",21.99

2024-12-21T00:06:29Z,LiteFuze 3000 Watt Japan Voltage Converter Transformer Step Up/Down - 100v to 120v / 120v to 100v Japanese Power Converter \ldots,109.99

\ldots

2025-01-03T22:33:29Z,"Deep Dream Tablecloths, 55 x 86 Inch \ldots (Rectangle/Oblong,6-8 Seats, Light Coffee)",26.99

2025-01-03T22:33:29Z,"SUEH DESIGN Faux Leather Placemats Set of 6 \ldots",18.86

2025-01-05T02:40:21Z,"Deep Dream Tablecloths, 55 x 86 Inch \ldots",0

. . . \textcolor{gray}{\textit{(The Jan 5 tablecloth at \$0 is a free replacement for the Jan 3 order. Routine restocking of body wash, iron supplements, and paper napkins also occurs in early January.)}}

%% file: cases/rubric/case8.tex
\textbf{Context (Continued)}

\textcolor{gray}{\textbf{(Part III)} [\textit{January 13--17 home decor cluster and immediate aftermath}]}

2025-01-13T22:35:24Z,MADIZZ Pack of 2 Boho Corduroy Striped Throw Pillow Covers 14x14 Inch Cream White \ldots,14.99

2025-01-13T22:35:24Z,"Utopia Bedding Throw Pillows Insert (Pack of 2, White) - 14 x 14 Inches \ldots",14.99

2025-01-13T22:35:24Z,"Artcest Set of 2 Soft Velvet Throw Pillow Covers \ldots 18""x18"" (Lilac)",9.99

2025-01-13T22:35:24Z,"Utopia Bedding Throw Pillows Insert (Pack of 2, White) - 18 x 18 Inches \ldots",16.99

2025-01-14T22:00:00Z,"Utopia Bedding Throw Pillows Insert (Pack of 2, White) - 14 x 14 Inches \ldots",14.99

2025-01-14T22:00:00Z,Phantoscope Pack of 2 Farmhouse Throw Pillow Covers Button Vintage Linen \ldots Purple 18 x 18 inches,15.99

\ldots

2025-01-17T18:18:00Z,"SAFFRON ROAD Butter Chicken with Basmati Rice, 10 OZ",6.19

2025-01-17T18:18:00Z,"LA FERMIERE Cafe Creme Dessert, 4.4 OZ",3.49

2025-01-17T18:27:29Z,"Sioloc Flower Shaped Throw Pillow \ldots 15.7"" White Flower Pillow \ldots",17.49

2025-01-17T18:27:29Z,"Snuggle Sac Purple Throw Blankets for Couch \ldots Heather Purple, 50x60 inches",39.99

2025-01-19T00:40:37Z,"puredown\textregistered{} 20x20 Pillow Inserts \ldots Set of 2 \ldots",25.99

2025-01-19T04:34:58Z,"NESCCI Birthday Wrapping Paper \ldots Colorful Birthday Balloon Pattern For Boys Girls \ldots",9.99

. . . \textcolor{gray}{\textit{(After the home decor cluster, purchasing returns to routine restocking. The full order history spans 2024-01-05 to 2026-03-05, totaling ${\sim}$361 records across personal care, groceries, household supplies, occasional clothing, gifts, and electronics.)}}\\
\midrule
\textbf{Rubrics}

1. The response should identify and reference specific purchases from the lead-up period when evaluating possible explanations. For example, it might mention items such as the voltage converter, groceries, tablecloths, placemats, or body wash when discussing whether the purchases suggest preparing for a guest or redecorating.

2. The response should compare the competing explanations (preparing for a guest vs redecorating) by evaluating how the listed purchases support or weaken each explanation. For example, the response might discuss whether items like groceries or table linens could indicate hosting a guest or whether decorative items might indicate redecorating.

3. The response should conclude that the purchase evidence does not clearly support the "preparing for a guest" explanation over redecorating.

4. The response should explain that purchases such as a voltage converter, groceries, and table linens are ambiguous and could have multiple explanations and therefore do not definitely indicate a visiting guest.

5. The response should explicitly state that the purchase history could possibly support both explanations of redecorating and preparing for a guest.\\
\midrule
\textbf{Model Generated Solution}

\#\#\# First: Full Context of Purchases in the Month Leading Up to the Jan 13--17 Home Decor Cluster

We first isolate all purchases from **December 13, 2024 to January 12, 2025**, separating routine purchases \ldots from non-routine, diagnostic purchases that shed light on the user's plans:

\#\#\#\# Routine, neutral purchases (no support for either hypothesis)

\ldots

- Personal care: Always maxi pads, RAW SUGAR body wash, MegaFood iron supplements, Eva NYC hair mask, Google Pixel 9 Pro XL case/screen protector

- Standard grocery runs (Dec 24 and Jan 10): Produce, dairy, pantry staples, seafood

\ldots

%% file: cases/response/case8.tex
\#\#\#\# Non-routine, diagnostic purchases (deviate from normal patterns)

These are the items that test our two hypotheses:

1. Dec 16: Waterproof hooded rain poncho, rechargeable hand warmer

2. Dec 17: Retevis RT628 walkie talkies, marketed as a toy for children ages 3--14

3. Dec 21: LiteFuze Japan-specific voltage converter (explicitly designed to convert between 100V Japanese power and 120V US power)

4. Jan 3: 55x86 dining tablecloth + set of 6 faux leather placemats

---

\#\#\# Analysis of Hypotheses

The pre-month purchases *could* align with either redecorating or guest preparation at first glance, but a close review **strongly supports that the user was preparing for an extended-stay visitor (a family with children traveling from Japan)**, not a general redecoration project.

\#\#\#\# Why the redecorating explanation is weakly supported

The only evidence for redecoration is the tablecloth and placemats \ldots However, this explanation has critical gaps:

- There are no other typical redecorating purchases in the pre-month (no paint, hardware, furniture, wall art, lighting, or renovation supplies). \ldots

- The voltage converter, kids' walkie talkies, and winter outdoor gear have no logical connection to redecorating \ldots

\#\#\#\# Why the guest preparation explanation is far better supported

Every non-routine pre-month purchase aligns perfectly with preparing to host an international family with children from Japan \ldots:

1. **Japan-specific voltage converter (Dec 21): This is the most definitive clue**

   The product is explicitly built for use between Japanese and US power grids. There is no evidence the user planned a trip to Japan in this timeframe \ldots The only logical use is for a visitor from Japan to power their personal electronics \ldots during their stay.

2. **Kids' walkie talkies (Dec 17)**

   The product is marketed as a toy for children 3--14, and the user has no history of purchasing children's items \ldots This is a low-cost entertainment item for visiting children.

3. **Winter outdoor gear (Dec 16)**

   The waterproof rain poncho and rechargeable hand warmer are practical for hosting guests in January \ldots who may not have packed appropriate outerwear for outdoor activities \ldots

4. **6 placemats + tablecloth (Jan 3)**

   A set of 6 placemats is intended for group dining, far more than a single person or couple would use for daily meals. The user rushed a free replacement for the damaged tablecloth on Jan 5, indicating they needed it by a specific upcoming date \ldots

\ldots

\#\#\# Additional Supporting Context

In the week immediately after the home decor cluster (Jan 17--24), the user purchased extra ready-to-eat meals (butter chicken rice, individual creme desserts) and wrapping paper, consistent with hosting out-of-town guests and exchanging belated holiday gifts with them. No further home decor purchases appear after this period, confirming the textile updates were a one-time, time-sensitive project tied to a guest arrival, not an ongoing redecoration.

%% file: CLBenchreal.bib
@misc{openclaw2025,
  author       = {Steinberger, Peter and {OpenClaw Contributors}},
  title        = {{OpenClaw}: The AI That Actually Does Things},
  year         = {2025},
  howpublished = {\url{https://github.com/openclaw/openclaw}},
  note         = {Open-source AI agent framework. MIT License. Accessed: 2025},
}

@misc{openai2025gpt51,
  title={GPT-5.1},
  author={OpenAI},
  year={2025},
  url={https://openai.com/zh-Hans-CN/index/gpt-5-1/}
}

@misc{anthropic2026claudeopus46,
  title={System Card: Claude Opus 4.6},
  author={Anthropic},
  year={2026},
  note={System card},
  url={https://www.anthropic.com/claude-opus-4-6-system-card}
}

@misc{google2026gemini31pro,
  title={Gemini 3.1 Pro Model Card},
  author={Google DeepMind},
  year={2026},
  note={Model card},
  url={https://deepmind.google/models/model-cards/gemini-3-1-pro/}
}

@misc{bytedance2026seed20pro,
  title={Seed 2.0 Official Launch},
  author={ByteDance Seed Team},
  year={2026},
  note={Official launch page},
  url={https://seed.bytedance.com/en/blog/seed-2-0-official-launch}
}

@article{moonshot2026k2_5,
  title={Kimi K2.5: Visual Agentic Intelligence},
  author={Kimi Team and Tongtong Bai and Yifan Bai and Yiping Bao and S. H. Cai and Yuan Cao and Y. Charles and H. S. Che and Cheng Chen and Guanduo Chen and Huarong Chen and Jia Chen and Jiahao Chen and Jianlong Chen and Jun Chen and Kefan Chen and Liang Chen and Ruijue Chen and Xinhao Chen and Yanru Chen and Yanxu Chen and Yicun Chen and Yimin Chen and Yingjiang Chen and Yuankun Chen and Yujie Chen and Yutian Chen and Zhirong Chen and Ziwei Chen and Dazhi Cheng and Minghan Chu and Jialei Cui and Jiaqi Deng and Muxi Diao and Hao Ding and Mengfan Dong and Mengnan Dong and Yuxin Dong and Yuhao Dong and Angang Du and Chenzhuang Du and Dikang Du and Lingxiao Du and Yulun Du and Yu Fan and Shengjun Fang and Qiulin Feng and Yichen Feng and Garimugai Fu and Kelin Fu and Hongcheng Gao and Tong Gao and Yuyao Ge and Shangyi Geng and Chengyang Gong and Xiaochen Gong and Zhuoma Gongque and Qizheng Gu and Xinran Gu and Yicheng Gu and Longyu Guan and Yuanying Guo and Xiaoru Hao and Weiran He and Wenyang He and Yunjia He and Chao Hong and Hao Hu and Jiaxi Hu and Yangyang Hu and Zhenxing Hu and Ke Huang and Ruiyuan Huang and Weixiao Huang and Zhiqi Huang and Tao Jiang and Zhejun Jiang and Xinyi Jin and Yu Jing and Guokun Lai and Aidi Li and C. Li and Cheng Li and Fang Li and Guanghe Li and Guanyu Li and Haitao Li and Haoyang Li and Jia Li and Jingwei Li and Junxiong Li and Lincan Li and Mo Li and Weihong Li and Wentao Li and Xinhang Li and Xinhao Li and Yang Li and Yanhao Li and Yiwei Li and Yuxiao Li and Zhaowei Li and Zheming Li and Weilong Liao and Jiawei Lin and Xiaohan Lin and Zhishan Lin and Zichao Lin and Cheng Liu and Chenyu Liu and Hongzhang Liu and Liang Liu and Shaowei Liu and Shudong Liu and Shuran Liu and Tianwei Liu and Tianyu Liu and Weizhou Liu and Xiangyan Liu and Yangyang Liu and Yanming Liu and Yibo Liu and Yuanxin Liu and Yue Liu and Zhengying Liu and Zhongnuo Liu and Enzhe Lu and Haoyu Lu and Zhiyuan Lu and Junyu Luo and Tongxu Luo and Yashuo Luo and Long Ma and Yingwei Ma and Shaoguang Mao and Yuan Mei and Xin Men and Fanqing Meng and Zhiyong Meng and Yibo Miao and Minqing Ni and Kun Ouyang and Siyuan Pan and Bo Pang and Yuchao Qian and Ruoyu Qin and Zeyu Qin and Jiezhong Qiu and Bowen Qu and Zeyu Shang and Youbo Shao and Tianxiao Shen and Zhennan Shen and Juanfeng Shi and Lidong Shi and Shengyuan Shi and Feifan Song and Pengwei Song and Tianhui Song and Xiaoxi Song and Hongjin Su and Jianlin Su and Zhaochen Su and Lin Sui and Jinsong Sun and Junyao Sun and Tongyu Sun and Flood Sung and Yunpeng Tai and Chuning Tang and Heyi Tang and Xiaojuan Tang and Zhengyang Tang and Jiawen Tao and Shiyuan Teng and Chaoran Tian and Pengfei Tian and Ao Wang and Bowen Wang and Chensi Wang and Chuang Wang and Congcong Wang and Dingkun Wang and Dinglu Wang and Dongliang Wang and Feng Wang and Hailong Wang and Haiming Wang and Hengzhi Wang and Huaqing Wang and Hui Wang and Jiahao Wang and Jinhong Wang and Jiuzheng Wang and Kaixin Wang and Linian Wang and Qibin Wang and Shengjie Wang and Shuyi Wang and Si Wang and Wei Wang and Xiaochen Wang and Xinyuan Wang and Yao Wang and Yejie Wang and Yipu Wang and Yiqin Wang and Yucheng Wang and Yuzhi Wang and Zhaoji Wang and Zhaowei Wang and Zhengtao Wang and Zhexu Wang and Zihan Wang and Zizhe Wang and Chu Wei and Ming Wei and Chuan Wen and Zichen Wen and Chengjie Wu and Haoning Wu and Junyan Wu and Rucong Wu and Wenhao Wu and Yuefeng Wu and Yuhao Wu and Yuxin Wu and Zijian Wu and Chenjun Xiao and Jin Xie and Xiaotong Xie and Yuchong Xie and Yifei Xin and Bowei Xing and Boyu Xu and Jianfan Xu and Jing Xu and Jinjing Xu and L. H. Xu and Lin Xu and Suting Xu and Weixin Xu and Xinbo Xu and Xinran Xu and Yangchuan Xu and Yichang Xu and Yuemeng Xu and Zelai Xu and Ziyao Xu and Junjie Yan and Yuzi Yan and Guangyao Yang and Hao Yang and Junwei Yang and Kai Yang and Ningyuan Yang and Ruihan Yang and Xiaofei Yang and Xinlong Yang and Ying Yang and Yi Yang and Yi Yang and Zhen Yang and Zhilin Yang and Zonghan Yang and Haotian Yao and Dan Ye and Wenjie Ye and Zhuorui Ye and Bohong Yin and Chengzhen Yu and Longhui Yu and Tao Yu and Tianxiang Yu and Enming Yuan and Mengjie Yuan and Xiaokun Yuan and Yang Yue and Weihao Zeng and Dunyuan Zha and Haobing Zhan and Dehao Zhang and Hao Zhang and Jin Zhang and Puqi Zhang and Qiao Zhang and Rui Zhang and Xiaobin Zhang and Y. Zhang and Yadong Zhang and Yangkun Zhang and Yichi Zhang and Yizhi Zhang and Yongting Zhang and Yu Zhang and Yushun Zhang and Yutao Zhang and Yutong Zhang and Zheng Zhang and Chenguang Zhao and Feifan Zhao and Jinxiang Zhao and Shuai Zhao and Xiangyu Zhao and Yikai Zhao and Zijia Zhao and Huabin Zheng and Ruihan Zheng and Shaojie Zheng and Tengyang Zheng and Junfeng Zhong and Longguang Zhong and Weiming Zhong and M. Zhou and Runjie Zhou and Xinyu Zhou and Zaida Zhou and Jinguo Zhu and Liya Zhu and Xinhao Zhu and Yuxuan Zhu and Zhen Zhu and Jingze Zhuang and Weiyu Zhuang and Ying Zou and Xinxing Zu},
  year={2026},
  eprint={2602.02276},
  archivePrefix={arXiv},
  primaryClass={cs.CL},
  url={https://arxiv.org/abs/2602.02276}
}

@misc{alibaba2026qwen35plus,
  title={{Qwen3.5}: Towards Native Multimodal Agents},
  author={{Qwen Team}},
  month={February},
  year={2026},
  url={https://qwen.ai/blog?id=qwen3.5}
}

@misc{xai2025grok4,
  title={Grok 4 Model Card},
  author={xAI},
  year={2025},
  note={Model card},
  url={https://data.x.ai/2025-08-20-grok-4-model-card.pdf}
}

@misc{deepseek2025v32,
  title={DeepSeek-V3.2: Pushing the Frontier of Open Large Language Models}, 
  author={DeepSeek-AI and Aixin Liu and Aoxue Mei and Bangcai Lin and Bing Xue and Bingxuan Wang and Bingzheng Xu and Bochao Wu and Bowei Zhang and Chaofan Lin and Chen Dong and Chengda Lu and Chenggang Zhao and Chengqi Deng and Chenhao Xu and Chong Ruan and Damai Dai and Daya Guo and Dejian Yang and Deli Chen and Erhang Li and Fangqi Zhou and Fangyun Lin and Fucong Dai and Guangbo Hao and Guanting Chen and Guowei Li and H. Zhang and Hanwei Xu and Hao Li and Haofen Liang and Haoran Wei and Haowei Zhang and Haowen Luo and Haozhe Ji and Honghui Ding and Hongxuan Tang and Huanqi Cao and Huazuo Gao and Hui Qu and Hui Zeng and Jialiang Huang and Jiashi Li and Jiaxin Xu and Jiewen Hu and Jingchang Chen and Jingting Xiang and Jingyang Yuan and Jingyuan Cheng and Jinhua Zhu and Jun Ran and Junguang Jiang and Junjie Qiu and Junlong Li and Junxiao Song and Kai Dong and Kaige Gao and Kang Guan and Kexin Huang and Kexing Zhou and Kezhao Huang and Kuai Yu and Lean Wang and Lecong Zhang and Lei Wang and Liang Zhao and Liangsheng Yin and Lihua Guo and Lingxiao Luo and Linwang Ma and Litong Wang and Liyue Zhang and M. S. Di and M. Y Xu and Mingchuan Zhang and Minghua Zhang and Minghui Tang and Mingxu Zhou and Panpan Huang and Peixin Cong and Peiyi Wang and Qiancheng Wang and Qihao Zhu and Qingyang Li and Qinyu Chen and Qiushi Du and Ruiling Xu and Ruiqi Ge and Ruisong Zhang and Ruizhe Pan and Runji Wang and Runqiu Yin and Runxin Xu and Ruomeng Shen and Ruoyu Zhang and S. H. Liu and Shanghao Lu and Shangyan Zhou and Shanhuang Chen and Shaofei Cai and Shaoyuan Chen and Shengding Hu and Shengyu Liu and Shiqiang Hu and Shirong Ma and Shiyu Wang and Shuiping Yu and Shunfeng Zhou and Shuting Pan and Songyang Zhou and Tao Ni and Tao Yun and Tian Pei and Tian Ye and Tianyuan Yue and Wangding Zeng and Wen Liu and Wenfeng Liang and Wenjie Pang and Wenjing Luo and Wenjun Gao and Wentao Zhang and Xi Gao and Xiangwen Wang and Xiao Bi and Xiaodong Liu and Xiaohan Wang and Xiaokang Chen and Xiaokang Zhang and Xiaotao Nie and Xin Cheng and Xin Liu and Xin Xie and Xingchao Liu and Xingkai Yu and Xingyou Li and Xinyu Yang and Xinyuan Li and Xu Chen and Xuecheng Su and Xuehai Pan and Xuheng Lin and Xuwei Fu and Y. Q. Wang and Yang Zhang and Yanhong Xu and Yanru Ma and Yao Li and Yao Li and Yao Zhao and Yaofeng Sun and Yaohui Wang and Yi Qian and Yi Yu and Yichao Zhang and Yifan Ding and Yifan Shi and Yiliang Xiong and Ying He and Ying Zhou and Yinmin Zhong and Yishi Piao and Yisong Wang and Yixiao Chen and Yixuan Tan and Yixuan Wei and Yiyang Ma and Yiyuan Liu and Yonglun Yang and Yongqiang Guo and Yongtong Wu and Yu Wu and Yuan Cheng and Yuan Ou and Yuanfan Xu and Yuduan Wang and Yue Gong and Yuhan Wu and Yuheng Zou and Yukun Li and Yunfan Xiong and Yuxiang Luo and Yuxiang You and Yuxuan Liu and Yuyang Zhou and Z. F. Wu and Z. Z. Ren and Zehua Zhao and Zehui Ren and Zhangli Sha and Zhe Fu and Zhean Xu and Zhenda Xie and Zhengyan Zhang and Zhewen Hao and Zhibin Gou and Zhicheng Ma and Zhigang Yan and Zhihong Shao and Zhixian Huang and Zhiyu Wu and Zhuoshu Li and Zhuping Zhang and Zian Xu and Zihao Wang and Zihui Gu and Zijia Zhu and Zilin Li and Zipeng Zhang and Ziwei Xie and Ziyi Gao and Zizheng Pan and Zongqing Yao and Bei Feng and Hui Li and J. L. Cai and Jiaqi Ni and Lei Xu and Meng Li and Ning Tian and R. J. Chen and R. L. Jin and S. S. Li and Shuang Zhou and Tianyu Sun and X. Q. Li and Xiangyue Jin and Xiaojin Shen and Xiaosha Chen and Xinnan Song and Xinyi Zhou and Y. X. Zhu and Yanping Huang and Yaohui Li and Yi Zheng and Yuchen Zhu and Yunxian Ma and Zhen Huang and Zhipeng Xu and Zhongyu Zhang and Dongjie Ji and Jian Liang and Jianzhong Guo and Jin Chen and Leyi Xia and Miaojun Wang and Mingming Li and Peng Zhang and Ruyi Chen and Shangmian Sun and Shaoqing Wu and Shengfeng Ye and T. Wang and W. L. Xiao and Wei An and Xianzu Wang and Xiaowen Sun and Xiaoxiang Wang and Ying Tang and Yukun Zha and Zekai Zhang and Zhe Ju and Zhen Zhang and Zihua Qu},
  year={2025},
  eprint={2512.02556},
  archivePrefix={arXiv},
  primaryClass={cs.CL},
  url={https://arxiv.org/abs/2512.02556}, 
}

@misc{minimax2026m25,
  title={MiniMax M2.5: Built for Real-World Productivity},
  author={MiniMax},
  year={2026},
  note={Official release page},
  url={https://minimax.io/news/minimax-m25}
}

@inproceedings{
zhou2024webarena,
title={WebArena: A Realistic Web Environment for Building Autonomous Agents},
author={Shuyan Zhou and Frank F. Xu and Hao Zhu and Xuhui Zhou and Robert Lo and Abishek Sridhar and Xianyi Cheng and Tianyue Ou and Yonatan Bisk and Daniel Fried and Uri Alon and Graham Neubig},
booktitle={The Twelfth International Conference on Learning Representations},
year={2024},
url={https://openreview.net/forum?id=oKn9c6ytLx}
}

@article{lewis2020retrieval,
  title={Retrieval-augmented generation for knowledge-intensive nlp tasks},
  author={Lewis, Patrick and Perez, Ethan and Piktus, Aleksandra and Petroni, Fabio and Karpukhin, Vladimir and Goyal, Naman and K{\"u}ttler, Heinrich and Lewis, Mike and Yih, Wen-tau and Rockt{\"a}schel, Tim and others},
  journal={Advances in neural information processing systems},
  volume={33},
  pages={9459--9474},
  year={2020}
}

@inproceedings{
xu2024searchinthechain,
title={Search-in-the-Chain: Interactively Enhancing Large Language Models with Search for Knowledge-intensive Tasks},
author={Shicheng Xu and Liang Pang and Huawei Shen and Xueqi Cheng and Tat-Seng Chua},
booktitle={The Web Conference 2024},
year={2024},
url={https://openreview.net/forum?id=tr0TcqitMH}
}

@article{liu2024lost,
  title={Lost in the middle: How language models use long contexts},
  author={Liu, Nelson F and Lin, Kevin and Hewitt, John and Paranjape, Ashwin and Bevilacqua, Michele and Petroni, Fabio and Liang, Percy},
  journal={Transactions of the association for computational linguistics},
  volume={12},
  pages={157--173},
  year={2024}
}

@article{dou2026cl,
  title={CL-bench: A Benchmark for Context Learning},
  author={Dou, Shihan and Zhang, Ming and Yin, Zhangyue and Huang, Chenhao and Shen, Yujiong and Wang, Junzhe and Chen, Jiayi and Ni, Yuchen and Ye, Junjie and Zhang, Cheng and others},
  journal={arXiv preprint arXiv:2602.03587},
  year={2026}
}

@article{mei2025survey,
  title={A survey of context engineering for large language models},
  author={Mei, Lingrui and Yao, Jiayu and Ge, Yuyao and Wang, Yiwei and Bi, Baolong and Cai, Yujun and Liu, Jiazhi and Li, Mingyu and Li, Zhong-Zhi and Zhang, Duzhen and others},
  journal={arXiv preprint arXiv:2507.13334},
  year={2025}
}

@online{anthropic2025context,
  author       = {Anthropic Applied AI Team},
  title        = {Effective context engineering for AI agents},
  year         = {2025},
  month        = {Sep},
  day          = {29},
  url          = {https://www.anthropic.com/engineering/effective-context-engineering-for-ai-agents},
}

@online{anthropic2024claude,
  title={Introducing the next generation of Claude},
  author={Anthropic},
  year={2024},
  url={https://www.anthropic.com/news/claude-3-family}
}

@article{yang2025qwen3,
  title={Qwen3 technical report},
  author={Yang, An and Li, Anfeng and Yang, Baosong and Zhang, Beichen and Hui, Binyuan and Zheng, Bo and Yu, Bowen and Gao, Chang and Huang, Chengen and Lv, Chenxu and others},
  journal={arXiv preprint arXiv:2505.09388},
  year={2025}
}

@misc{qwen36plus,
    title = {{Qwen3.6-Plus}: Towards Real World Agents},
    url = {https://qwen.ai/blog?id=qwen3.6},
    author = {{Qwen Team}},
    month = {April},
    year = {2026}
}

@misc{openai2025gpt5,
  author       = {{OpenAI}},
  title        = {Introducing {GPT-5}},
  year         = {2025},
  howpublished = {\url{https://openai.com/index/introducing-gpt-5-4/}},
  note         = {Accessed: 2025},
}

@misc{chen2021evaluating,
      title={Evaluating Large Language Models Trained on Code}, 
      author={Mark Chen and Jerry Tworek and Heewoo Jun and Qiming Yuan and Henrique Ponde de Oliveira Pinto and Jared Kaplan and Harri Edwards and Yuri Burda and Nicholas Joseph and Greg Brockman and Alex Ray and Raul Puri and Gretchen Krueger and Michael Petrov and Heidy Khlaaf and Girish Sastry and Pamela Mishkin and Brooke Chan and Scott Gray and Nick Ryder and Mikhail Pavlov and Alethea Power and Lukasz Kaiser and Mohammad Bavarian and Clemens Winter and Philippe Tillet and Felipe Petroski Such and Dave Cummings and Matthias Plappert and Fotios Chantzis and Elizabeth Barnes and Ariel Herbert-Voss and William Hebgen Guss and Alex Nichol and Alex Paino and Nikolas Tezak and Jie Tang and Igor Babuschkin and Suchir Balaji and Shantanu Jain and William Saunders and Christopher Hesse and Andrew N. Carr and Jan Leike and Josh Achiam and Vedant Misra and Evan Morikawa and Alec Radford and Matthew Knight and Miles Brundage and Mira Murati and Katie Mayer and Peter Welinder and Bob McGrew and Dario Amodei and Sam McCandlish and Ilya Sutskever and Wojciech Zaremba},
      year={2021},
      eprint={2107.03374},
      archivePrefix={arXiv},
      primaryClass={cs.LG},
      url={https://arxiv.org/abs/2107.03374}, 
}

@inproceedings{
jimenez2024swebench,
title={{SWE}-bench: Can Language Models Resolve Real-world Github Issues?},
author={Carlos E Jimenez and John Yang and Alexander Wettig and Shunyu Yao and Kexin Pei and Ofir Press and Karthik R Narasimhan},
booktitle={The Twelfth International Conference on Learning Representations},
year={2024},
url={https://openreview.net/forum?id=VTF8yNQM66}
}

@article{brynjolfsson2025generative,
  title={Generative AI at work},
  author={Brynjolfsson, Erik and Li, Danielle and Raymond, Lindsey},
  journal={The Quarterly Journal of Economics},
  volume={140},
  number={2},
  pages={889--942},
  year={2025},
  publisher={Oxford University Press}
}

@article{yao2024tau,
  title={$\tau$-bench: A benchmark for tool-agent-user interaction in real-world domains, 2024},
  author={Yao, Shunyu and Shinn, Noah and Razavi, Pedram and Narasimhan, Karthik},
  journal={URL https://arxiv. org/abs/2406.12045},
  year={2024}
}

@article{wang2024survey,
  title={A survey on large language model based autonomous agents},
  author={Wang, Lei and Ma, Chen and Feng, Xueyang and Zhang, Zeyu and Yang, Hao and Zhang, Jingsen and Chen, Zhiyuan and Tang, Jiakai and Chen, Xu and Lin, Yankai and others},
  journal={Frontiers of Computer Science},
  volume={18},
  number={6},
  pages={186345},
  year={2024},
  publisher={Springer}
}

@inproceedings{deshpande2025multichallenge,
  title={Multichallenge: A realistic multi-turn conversation evaluation benchmark challenging to frontier llms},
  author={Deshpande, Kaustubh and Sirdeshmukh, Ved and Mols, Johannes Baptist and Jin, Lifeng and Hernandez-Cardona, Ed-Yeremai and Lee, Dean and Kritz, Jeremy and Primack, Willow E and Yue, Summer and Xing, Chen},
  booktitle={Findings of the Association for Computational Linguistics: ACL 2025},
  pages={18632--18702},
  year={2025}
}

@article{he2025vitabench,
  title={Vitabench: Benchmarking llm agents with versatile interactive tasks in real-world applications},
  author={He, Wei and Sun, Yueqing and Hao, Hongyan and Hao, Xueyuan and Xia, Zhikang and Gu, Qi and Han, Chengcheng and Zhao, Dengchang and Su, Hui and Zhang, Kefeng and others},
  journal={arXiv preprint arXiv:2509.26490},
  year={2025}
}

@inproceedings{
liu2024agentbench,
title={AgentBench: Evaluating {LLM}s as Agents},
author={Xiao Liu and Hao Yu and Hanchen Zhang and Yifan Xu and Xuanyu Lei and Hanyu Lai and Yu Gu and Hangliang Ding and Kaiwen Men and Kejuan Yang and Shudan Zhang and Xiang Deng and Aohan Zeng and Zhengxiao Du and Chenhui Zhang and Sheng Shen and Tianjun Zhang and Yu Su and Huan Sun and Minlie Huang and Yuxiao Dong and Jie Tang},
booktitle={The Twelfth International Conference on Learning Representations},
year={2024},
url={https://openreview.net/forum?id=zAdUB0aCTQ}
}

@inproceedings{yang-etal-2018-hotpotqa,
    title = "{H}otpot{QA}: A Dataset for Diverse, Explainable Multi-hop Question Answering",
    author = "Yang, Zhilin  and
      Qi, Peng  and
      Zhang, Saizheng  and
      Bengio, Yoshua  and
      Cohen, William  and
      Salakhutdinov, Ruslan  and
      Manning, Christopher D.",
    booktitle = "Proceedings of the 2018 Conference on Empirical Methods in Natural Language Processing",
    month = oct # "-" # nov,
    year = "2018",
    address = "Brussels, Belgium",
    publisher = "Association for Computational Linguistics",
    url = "https://aclanthology.org/D18-1259/",
    doi = "10.18653/v1/D18-1259",
    pages = "2369--2380",
}

@inproceedings{
schick2023toolformer,
title={Toolformer: Language Models Can Teach Themselves to Use Tools},
author={Timo Schick and Jane Dwivedi-Yu and Roberto Dessi and Roberta Raileanu and Maria Lomeli and Eric Hambro and Luke Zettlemoyer and Nicola Cancedda and Thomas Scialom},
booktitle={Thirty-seventh Conference on Neural Information Processing Systems},
year={2023},
url={https://openreview.net/forum?id=Yacmpz84TH}
}

@misc{
zhou2024language,
title={Language Agent Tree Search Unifies Reasoning Acting and Planning in Language Models},
author={Andy Zhou and Kai Yan and Michal Shlapentokh-Rothman and Haohan Wang and Yu-Xiong Wang},
year={2024},
url={https://openreview.net/forum?id=6LNTSrJjBe}
}

@misc{mialon2023gaiabenchmarkgeneralai,
      title={GAIA: a benchmark for General AI Assistants}, 
      author={Grégoire Mialon and Clémentine Fourrier and Craig Swift and Thomas Wolf and Yann LeCun and Thomas Scialom},
      year={2023},
      eprint={2311.12983},
      archivePrefix={arXiv},
      primaryClass={cs.CL},
      url={https://arxiv.org/abs/2311.12983}, 
}

@inproceedings{
ma2024agentboard,
title={AgentBoard: An Analytical Evaluation Board of Multi-turn {LLM} Agents},
author={Chang Ma and Junlei Zhang and Zhihao Zhu and Cheng Yang and Yujiu Yang and Yaohui Jin and Zhenzhong Lan and Lingpeng Kong and Junxian He},
booktitle={The Thirty-eight Conference on Neural Information Processing Systems Datasets and Benchmarks Track},
year={2024},
url={https://openreview.net/forum?id=4S8agvKjle}
}

@misc{dou2025evalearnquantifyinglearningcapability,
      title={EvaLearn: Quantifying the Learning Capability and Efficiency of LLMs via Sequential Problem Solving}, 
      author={Shihan Dou and Ming Zhang and Chenhao Huang and Jiayi Chen and Feng Chen and Shichun Liu and Yan Liu and Chenxiao Liu and Cheng Zhong and Zongzhang Zhang and Tao Gui and Chao Xin and Chengzhi Wei and Lin Yan and Yonghui Wu and Qi Zhang and Xuanjing Huang},
      year={2025},
      eprint={2506.02672},
      archivePrefix={arXiv},
      primaryClass={cs.CL},
      url={https://arxiv.org/abs/2506.02672}, 
}

@inproceedings{bai-etal-2024-longbench,
    title = "{L}ong{B}ench: A Bilingual, Multitask Benchmark for Long Context Understanding",
    author = "Bai, Yushi  and
      Lv, Xin  and
      Zhang, Jiajie  and
      Lyu, Hongchang  and
      Tang, Jiankai  and
      Huang, Zhidian  and
      Du, Zhengxiao  and
      Liu, Xiao  and
      Zeng, Aohan  and
      Hou, Lei  and
      Dong, Yuxiao  and
      Tang, Jie  and
      Li, Juanzi",
    booktitle = "Proceedings of the 62nd Annual Meeting of the Association for Computational Linguistics (Volume 1: Long Papers)",
    month = aug,
    year = "2024",
    address = "Bangkok, Thailand",
    publisher = "Association for Computational Linguistics",
    url = "https://aclanthology.org/2024.acl-long.172/",
    doi = "10.18653/v1/2024.acl-long.172",
    pages = "3119--3137",
}

@inproceedings{bai-etal-2025-longbench,
    title = "{L}ong{B}ench v2: Towards Deeper Understanding and Reasoning on Realistic Long-context Multitasks",
    author = "Bai, Yushi  and
      Tu, Shangqing  and
      Zhang, Jiajie  and
      Peng, Hao  and
      Wang, Xiaozhi  and
      Lv, Xin  and
      Cao, Shulin  and
      Xu, Jiazheng  and
      Hou, Lei  and
      Dong, Yuxiao  and
      Tang, Jie  and
      Li, Juanzi",
    booktitle = "Proceedings of the 63rd Annual Meeting of the Association for Computational Linguistics (Volume 1: Long Papers)",
    month = jul,
    year = "2025",
    address = "Vienna, Austria",
    publisher = "Association for Computational Linguistics",
    url = "https://aclanthology.org/2025.acl-long.183/",
    doi = "10.18653/v1/2025.acl-long.183",
    pages = "3639--3664",
    ISBN = "979-8-89176-251-0",
}

@inproceedings{li-etal-2024-loogle,
    title = "{L}oo{GLE}: Can Long-Context Language Models Understand Long Contexts?",
    author = "Li, Jiaqi  and
      Wang, Mengmeng  and
      Zheng, Zilong  and
      Zhang, Muhan",
    booktitle = "Proceedings of the 62nd Annual Meeting of the Association for Computational Linguistics (Volume 1: Long Papers)",
    month = aug,
    year = "2024",
    address = "Bangkok, Thailand",
    publisher = "Association for Computational Linguistics",
    url = "https://aclanthology.org/2024.acl-long.859/",
    doi = "10.18653/v1/2024.acl-long.859",
    pages = "16304--16333",
}

@inproceedings{
hsieh2024ruler,
title={{RULER}: What{\textquoteright}s the Real Context Size of Your Long-Context Language Models?},
author={Cheng-Ping Hsieh and Simeng Sun and Samuel Kriman and Shantanu Acharya and Dima Rekesh and Fei Jia and Boris Ginsburg},
booktitle={First Conference on Language Modeling},
year={2024},
url={https://openreview.net/forum?id=kIoBbc76Sy}
}

@inproceedings{
yen2025helmet,
title={{HELMET}: How to Evaluate Long-context Models Effectively and Thoroughly},
author={Howard Yen and Tianyu Gao and Minmin Hou and Ke Ding and Daniel Fleischer and Peter Izsak and Moshe Wasserblat and Danqi Chen},
booktitle={The Thirteenth International Conference on Learning Representations},
year={2025},
url={https://openreview.net/forum?id=293V3bJbmE}
}

@inproceedings{an-etal-2024-l,
    title = "{L}-Eval: Instituting Standardized Evaluation for Long Context Language Models",
    author = "An, Chenxin  and
      Gong, Shansan  and
      Zhong, Ming  and
      Zhao, Xingjian  and
      Li, Mukai  and
      Zhang, Jun  and
      Kong, Lingpeng  and
      Qiu, Xipeng",
    booktitle = "Proceedings of the 62nd Annual Meeting of the Association for Computational Linguistics (Volume 1: Long Papers)",
    month = aug,
    year = "2024",
    address = "Bangkok, Thailand",
    publisher = "Association for Computational Linguistics",
    url = "https://aclanthology.org/2024.acl-long.776/",
    doi = "10.18653/v1/2024.acl-long.776",
    pages = "14388--14411",
}

@inproceedings{
qin2024toolllm,
title={Tool{LLM}: Facilitating Large Language Models to Master 16000+ Real-world {API}s},
author={Yujia Qin and Shihao Liang and Yining Ye and Kunlun Zhu and Lan Yan and Yaxi Lu and Yankai Lin and Xin Cong and Xiangru Tang and Bill Qian and Sihan Zhao and Lauren Hong and Runchu Tian and Ruobing Xie and Jie Zhou and Mark Gerstein and dahai li and Zhiyuan Liu and Maosong Sun},
booktitle={The Twelfth International Conference on Learning Representations},
year={2024},
url={https://openreview.net/forum?id=dHng2O0Jjr}
}

@inproceedings{
xie2024osworld,
title={{OSW}orld: Benchmarking Multimodal Agents for Open-Ended Tasks in Real Computer Environments},
author={Tianbao Xie and Danyang Zhang and Jixuan Chen and Xiaochuan Li and Siheng Zhao and Ruisheng Cao and Toh Jing Hua and Zhoujun Cheng and Dongchan Shin and Fangyu Lei and Yitao Liu and Yiheng Xu and Shuyan Zhou and Silvio Savarese and Caiming Xiong and Victor Zhong and Tao Yu},
booktitle={The Thirty-eight Conference on Neural Information Processing Systems Datasets and Benchmarks Track},
year={2024},
url={https://openreview.net/forum?id=tN61DTr4Ed}
}

@inproceedings{ToolEyes,
  author       = {Junjie Ye and
                  Guanyu Li and
                  Songyang Gao and
                  Caishuang Huang and
                  Yilong Wu and
                  Sixian Li and
                  Xiaoran Fan and
                  Shihan Dou and
                  Tao Ji and
                  Qi Zhang and
                  Tao Gui and
                  Xuanjing Huang},
  title        = {ToolEyes: Fine-Grained Evaluation for Tool Learning Capabilities of
                  Large Language Models in Real-world Scenarios},
  booktitle    = {Proceedings of the 31st International Conference on Computational
                  Linguistics, {COLING} 2025, Abu Dhabi, UAE, January 19-24, 2025},
  pages        = {156--187},
  publisher    = {Association for Computational Linguistics},
  year         = {2025},
  url          = {https://aclanthology.org/2025.coling-main.12/},
  bibsource    = {dblp computer science bibliography, https://dblp.org}
}

@inproceedings{xie2024travelplanner,
  title={TravelPlanner: A Benchmark for Real-World Planning with Language Agents},
  author={Xie, Jian and Zhang, Kai and Chen, Jiangjie and Zhu, Tinghui and Lou, Renze and Tian, Yuandong and Xiao, Yanghua and Su, Yu},
  booktitle={Forty-first International Conference on Machine Learning},
  year={2024}
}

@inproceedings{
jang2025videowebarena,
title={VideoWebArena:  Evaluating Long Context Multimodal Agents with Video Understanding Web Tasks},
author={Lawrence Keunho Jang and Yinheng Li and Dan Zhao and Charles Ding and Justin Lin and Paul Pu Liang and Rogerio Bonatti and Kazuhito Koishida},
booktitle={The Thirteenth International Conference on Learning Representations},
year={2025},
url={https://openreview.net/forum?id=unDQOUah0F}
}

@inproceedings{
deng2023mindweb,
title={Mind2Web: Towards a Generalist Agent for the Web},
author={Xiang Deng and Yu Gu and Boyuan Zheng and Shijie Chen and Samuel Stevens and Boshi Wang and Huan Sun and Yu Su},
booktitle={Thirty-seventh Conference on Neural Information Processing Systems Datasets and Benchmarks Track},
year={2023},
url={https://openreview.net/forum?id=kiYqbO3wqw}
}

@inproceedings{
yao2022webshop,
title={WebShop: Towards Scalable Real-World Web Interaction with Grounded Language Agents},
author={Shunyu Yao and Howard Chen and John Yang and Karthik R Narasimhan},
booktitle={Advances in Neural Information Processing Systems},
editor={Alice H. Oh and Alekh Agarwal and Danielle Belgrave and Kyunghyun Cho},
year={2022},
url={https://openreview.net/forum?id=R9KnuFlvnU}
}

@misc{2019textworldlearningenvironmenttextbased,
      title={TextWorld: A Learning Environment for Text-based Games}, 
      author={Marc-Alexandre Côté and Ákos Kádár and Xingdi Yuan and Ben Kybartas and Tavian Barnes and Emery Fine and James Moore and Ruo Yu Tao and Matthew Hausknecht and Layla El Asri and Mahmoud Adada and Wendy Tay and Adam Trischler},
      year={2019},
      eprint={1806.11532},
      archivePrefix={arXiv},
      primaryClass={cs.LG},
      url={https://arxiv.org/abs/1806.11532}, 
}

@article{ying2026retrieval,
  title={Retrieval-Infused Reasoning Sandbox: A Benchmark for Decoupling Retrieval and Reasoning Capabilities},
  author={Ying, Shuangshuang and Wang, Zheyu and Peng, Yunjian and Chen, Jin and Wu, Yuhao and Lin, Hongbin and He, Dingyu and Liu, Siyi and Yu, Gengchen and Piao, YinZhu and others},
  journal={arXiv preprint arXiv:2601.21937},
  year={2026}
}

@article{ding2026octobench,
  title={OctoBench: Benchmarking Scaffold-Aware Instruction Following in Repository-Grounded Agentic Coding},
  author={Ding, Deming and Liu, Shichun and Yang, Enhui and Lin, Jiahang and Chen, Ziying and Dou, Shihan and Guo, Honglin and Cheng, Weiyu and Zhao, Pengyu and Xiao, Chengjun and others},
  journal={arXiv preprint arXiv:2601.10343},
  year={2026}
}

@misc{hu2026cl4sebenchmarkingcontextlearning,
      title={CL4SE: Benchmarking Context Learning on Software Engineering}, 
      author={Haichuan Hu and Quanjun Zhang and Ye Shang and Guoqing Xie and Chunrong Fang and Zhenyu Chen and Liang Xiao},
      year={2026},
      eprint={2602.23047},
      archivePrefix={arXiv},
      primaryClass={cs.SE},
      url={https://arxiv.org/abs/2602.23047}, 
}

@inproceedings{
cemri2025why,
title={Why Do Multi-Agent {LLM} Systems Fail?},
author={Mert Cemri and Melissa Z Pan and Shuyi Yang and Lakshya A Agrawal and Bhavya Chopra and Rishabh Tiwari and Kurt Keutzer and Aditya Parameswaran and Dan Klein and Kannan Ramchandran and Matei Zaharia and Joseph E. Gonzalez and Ion Stoica},
booktitle={The Thirty-ninth Annual Conference on Neural Information Processing Systems Datasets and Benchmarks Track},
year={2025},
url={https://openreview.net/forum?id=fAjbYBmonr}
}

@article{
sumers2024cognitive,
title={Cognitive Architectures for Language Agents},
author={Theodore Sumers and Shunyu Yao and Karthik R Narasimhan and Thomas L. Griffiths},
journal={Transactions on Machine Learning Research},
issn={2835-8856},
year={2024},
url={https://openreview.net/forum?id=1i6ZCvflQJ},
note={Survey Certification, Featured Certification}
}

@inproceedings{kwan-etal-2024-m4le,
    title = "{M}4{LE}: A Multi-Ability Multi-Range Multi-Task Multi-Domain Long-Context Evaluation Benchmark for Large Language Models",
    author = "Kwan, Wai-Chung  and
      Zeng, Xingshan  and
      Wang, Yufei  and
      Sun, Yusen  and
      Li, Liangyou  and
      Jiang, Yuxin  and
      Shang, Lifeng  and
      Liu, Qun  and
      Wong, Kam-Fai",
    booktitle = "Proceedings of the 62nd Annual Meeting of the Association for Computational Linguistics (Volume 1: Long Papers)",
    month = aug,
    year = "2024",
    address = "Bangkok, Thailand",
    publisher = "Association for Computational Linguistics",
    url = "https://aclanthology.org/2024.acl-long.832/",
    doi = "10.18653/v1/2024.acl-long.832",
    pages = "15568--15592",
}

@inproceedings{
garg2022what,
title={What Can Transformers Learn In-Context? A Case Study of Simple Function Classes},
author={Shivam Garg and Dimitris Tsipras and Percy Liang and Gregory Valiant},
booktitle={Advances in Neural Information Processing Systems},
editor={Alice H. Oh and Alekh Agarwal and Danielle Belgrave and Kyunghyun Cho},
year={2022},
url={https://openreview.net/forum?id=flNZJ2eOet}
}

@inproceedings{dong2024survey,
  title={A survey on in-context learning},
  author={Dong, Qingxiu and Li, Lei and Dai, Damai and Zheng, Ce and Ma, Jingyuan and Li, Rui and Xia, Heming and Xu, Jingjing and Wu, Zhiyong and Chang, Baobao and others},
  booktitle={Proceedings of the 2024 conference on empirical methods in natural language processing},
  pages={1107--1128},
  year={2024}
}

@inproceedings{bai2024longbench,
  title={Longbench: A bilingual, multitask benchmark for long context understanding},
  author={Bai, Yushi and Lv, Xin and Zhang, Jiajie and Lyu, Hongchang and Tang, Jiankai and Huang, Zhidian and Du, Zhengxiao and Liu, Xiao and Zeng, Aohan and Hou, Lei and others},
  booktitle={Proceedings of the 62nd annual meeting of the association for computational linguistics (volume 1: Long papers)},
  pages={3119--3137},
  year={2024}
}

@inproceedings{dong2024bamboo,
  title={Bamboo: A comprehensive benchmark for evaluating long text modeling capacities of large language models},
  author={Dong, Zican and Tang, Tianyi and Li, Junyi and Zhao, Wayne Xin and Wen, Ji-Rong},
  booktitle={Proceedings of the 2024 Joint International Conference on Computational Linguistics, Language Resources and Evaluation (LREC-COLING 2024)},
  pages={2086--2099},
  year={2024}
}

@inproceedings{li2024loogle,
  title={Loogle: Can long-context language models understand long contexts?},
  author={Li, Jiaqi and Wang, Mengmeng and Zheng, Zilong and Zhang, Muhan},
  booktitle={Proceedings of the 62nd Annual Meeting of the Association for Computational Linguistics (Volume 1: Long Papers)},
  pages={16304--16333},
  year={2024}
}

@article{zhang2024infty,
  title={$\infty$-Bench: Extending Long Context Evaluation Beyond 100K Tokens},
  author={Zhang, Xinrong and Chen, Yingfa and Hu, Shengding and Xu, Zihang and Chen, Junhao and Hao, Moo and Han, Xu and Thai, Zhen and Wang, Shuo and Liu, Zhiyuan and others},
  journal={arXiv preprint arXiv:2402.13718},
  year={2024}
}

@article{yen2024helmet,
  title={Helmet: How to evaluate long-context language models effectively and thoroughly},
  author={Yen, Howard and Gao, Tianyu and Hou, Minmin and Ding, Ke and Fleischer, Daniel and Izsak, Peter and Wasserblat, Moshe and Chen, Danqi},
  journal={arXiv preprint arXiv:2410.02694},
  year={2024}
}

@article{kuratov2024babilong,
  title={Babilong: Testing the limits of llms with long context reasoning-in-a-haystack},
  author={Kuratov, Yury and Bulatov, Aydar and Anokhin, Petr and Rodkin, Ivan and Sorokin, Dmitry and Sorokin, Artyom and Burtsev, Mikhail},
  journal={Advances in Neural Information Processing Systems},
  volume={37},
  pages={106519--106554},
  year={2024}
}

@inproceedings{song2025counting,
  title={Counting-stars: A multi-evidence, position-aware, and scalable benchmark for evaluating long-context large language models},
  author={Song, Mingyang and Zheng, Mao and Luo, Xuan},
  booktitle={Proceedings of the 31st International Conference on Computational Linguistics},
  pages={3753--3763},
  year={2025}
}

@inproceedings{zhang2025longcite,
  title={Longcite: Enabling llms to generate fine-grained citations in long-context qa},
  author={Zhang, Jiajie and Bai, Yushi and Lv, Xin and Gu, Wanjun and Liu, Danqing and Zou, Minhao and Cao, Shulin and Hou, Lei and Dong, Yuxiao and Feng, Ling and others},
  booktitle={Findings of the Association for Computational Linguistics: ACL 2025},
  pages={5098--5122},
  year={2025}
}

@inproceedings{lee-etal-2025-ethic,
    title = "{ETHIC}: Evaluating Large Language Models on Long-Context Tasks with High Information Coverage",
    author = "Lee, Taewhoo  and
      Yoon, Chanwoong  and
      Jang, Kyochul  and
      Lee, Donghyeon  and
      Song, Minju  and
      Kim, Hyunjae  and
      Kang, Jaewoo",
    booktitle = "Proceedings of the 2025 Conference of the Nations of the Americas Chapter of the Association for Computational Linguistics: Human Language Technologies (Volume 1: Long Papers)",
    month = apr,
    year = "2025",
    address = "Albuquerque, New Mexico",
    publisher = "Association for Computational Linguistics",
    url = "https://aclanthology.org/2025.naacl-long.283/",
    doi = "10.18653/v1/2025.naacl-long.283",
    pages = "5497--5512",
    ISBN = "979-8-89176-189-6",
}

@inproceedings{
wu2025longmemeval,
title={LongMemEval: Benchmarking Chat Assistants on Long-Term Interactive Memory},
author={Di Wu and Hongwei Wang and Wenhao Yu and Yuwei Zhang and Kai-Wei Chang and Dong Yu},
booktitle={The Thirteenth International Conference on Learning Representations},
year={2025},
url={https://openreview.net/forum?id=pZiyCaVuti}
}

@inproceedings{maharana-etal-2024-evaluating,
    title = "Evaluating Very Long-Term Conversational Memory of {LLM} Agents",
    author = "Maharana, Adyasha  and
      Lee, Dong-Ho  and
      Tulyakov, Sergey  and
      Bansal, Mohit  and
      Barbieri, Francesco  and
      Fang, Yuwei",
    booktitle = "Proceedings of the 62nd Annual Meeting of the Association for Computational Linguistics (Volume 1: Long Papers)",
    month = aug,
    year = "2024",
    address = "Bangkok, Thailand",
    publisher = "Association for Computational Linguistics",
    url = "https://aclanthology.org/2024.acl-long.747/",
    doi = "10.18653/v1/2024.acl-long.747",
    pages = "13851--13870",
}

@inproceedings{MemoryBank,
author = {Zhong, Wanjun and Guo, Lianghong and Gao, Qiqi and Ye, He and Wang, Yanlin},
title = {MemoryBank: enhancing large language models with long-term memory},
year = {2024},
isbn = {978-1-57735-887-9},
publisher = {AAAI Press},
url = {https://doi.org/10.1609/aaai.v38i17.29946},
doi = {10.1609/aaai.v38i17.29946},
booktitle = {Proceedings of the Thirty-Eighth AAAI Conference on Artificial Intelligence and Thirty-Sixth Conference on Innovative Applications of Artificial Intelligence and Fourteenth Symposium on Educational Advances in Artificial Intelligence},
articleno = {2198},
numpages = {8},
series = {AAAI'24/IAAI'24/EAAI'24}
}

@inproceedings{MEMORYLLM,
author = {Wang, Yu and Gao, Yifan and Chen, Xiusi and Jiang, Haoming and Li, Shiyang and Yang, Jingfeng and Yin, Qingyu and Li, Zheng and Li, Xian and Yin, Bing and Shang, Jingbo and McAuley, Julian},
title = {MEMORYLLM: towards self-updatable large language models},
year = {2024},
publisher = {JMLR.org},
booktitle = {Proceedings of the 41st International Conference on Machine Learning},
articleno = {2065},
numpages = {14},
location = {Vienna, Austria},
series = {ICML'24}
}

@misc{backlund2025vendingbenchbenchmarklongtermcoherence,
      title={Vending-Bench: A Benchmark for Long-Term Coherence of Autonomous Agents}, 
      author={Axel Backlund and Lukas Petersson},
      year={2025},
      eprint={2502.15840},
      archivePrefix={arXiv},
      primaryClass={cs.AI},
      url={https://arxiv.org/abs/2502.15840}, 
}

@inproceedings{xu-etal-2022-beyond,
    title = "Beyond Goldfish Memory: Long-Term Open-Domain Conversation",
    author = "Xu, Jing  and
      Szlam, Arthur  and
      Weston, Jason",
    booktitle = "Proceedings of the 60th Annual Meeting of the Association for Computational Linguistics (Volume 1: Long Papers)",
    month = may,
    year = "2022",
    address = "Dublin, Ireland",
    publisher = "Association for Computational Linguistics",
    url = "https://aclanthology.org/2022.acl-long.356/",
    doi = "10.18653/v1/2022.acl-long.356",
    pages = "5180--5197",
}

@inproceedings{zhu-etal-2025-multiagentbench,
    title = "{M}ulti{A}gent{B}ench : Evaluating the Collaboration and Competition of {LLM} agents",
    author = "Zhu, Kunlun  and
      Du, Hongyi  and
      Hong, Zhaochen  and
      Yang, Xiaocheng  and
      Guo, Shuyi  and
      Wang, Zhe  and
      Wang, Zhenhailong  and
      Qian, Cheng  and
      Tang, Xiangru  and
      Ji, Heng  and
      You, Jiaxuan",
    booktitle = "Proceedings of the 63rd Annual Meeting of the Association for Computational Linguistics (Volume 1: Long Papers)",
    month = jul,
    year = "2025",
    address = "Vienna, Austria",
    publisher = "Association for Computational Linguistics",
    url = "https://aclanthology.org/2025.acl-long.421/",
    doi = "10.18653/v1/2025.acl-long.421",
    pages = "8580--8622",
    ISBN = "979-8-89176-251-0",
}

@inproceedings{
lin2025wildbench,
title={WildBench: Benchmarking {LLM}s with Challenging Tasks from Real Users in the Wild},
author={Bill Yuchen Lin and Yuntian Deng and Khyathi Chandu and Abhilasha Ravichander and Valentina Pyatkin and Nouha Dziri and Ronan Le Bras and Yejin Choi},
booktitle={The Thirteenth International Conference on Learning Representations},
year={2025},
url={https://openreview.net/forum?id=MKEHCx25xp}
}

@inproceedings{
drouin2024workarena,
title={WorkArena: How Capable are Web Agents at Solving Common Knowledge Work Tasks?},
author={Alexandre Drouin and Maxime Gasse and Massimo Caccia and Issam H. Laradji and Manuel Del Verme and Tom Marty and David Vazquez and Nicolas Chapados and Alexandre Lacoste},
booktitle={Forty-first International Conference on Machine Learning},
year={2024},
url={https://openreview.net/forum?id=BRfqYrikdo}
}

@inproceedings{Park2023GenerativeAgents,  
author = {Park, Joon Sung and O'Brien, Joseph C. and Cai, Carrie J. and Morris, Meredith Ringel and Liang, Percy and Bernstein, Michael S.},  
title = {Generative Agents: Interactive Simulacra of Human Behavior},  
year = {2023},  
publisher = {Association for Computing Machinery},  
address = {New York, NY, USA},  
booktitle = {In the 36th Annual ACM Symposium on User Interface Software and Technology (UIST '23)},  
location = {San Francisco, CA, USA},  
series = {UIST '23}
}

@inproceedings{yin-etal-2025-mmau,
    title = "{MMAU}: A Holistic Benchmark of Agent Capabilities Across Diverse Domains",
    author = "Yin, Guoli  and
      Bai, Haoping  and
      Ma, Shuang  and
      Nan, Feng  and
      Sun, Yanchao  and
      Xu, Zhaoyang  and
      Ma, Shen  and
      Lu, Jiarui  and
      Kong, Xiang  and
      Zhang, Aonan  and
      Yap, Dian Ang  and
      Zhang, Yizhe  and
      Ahnert, Karsten  and
      Kamath, Vik  and
      Berglund, Mathias  and
      Walsh, Dominic  and
      Gindele, Tobias  and
      Wiest, Juergen  and
      Lai, Zhengfeng  and
      Wang, Xiaoming Simon  and
      Shan, Jiulong  and
      Cao, Meng  and
      Pang, Ruoming  and
      Wang, Zirui",
    booktitle = "Findings of the Association for Computational Linguistics: NAACL 2025",
    month = apr,
    year = "2025",
    address = "Albuquerque, New Mexico",
    publisher = "Association for Computational Linguistics",
    url = "https://aclanthology.org/2025.findings-naacl.267/",
    doi = "10.18653/v1/2025.findings-naacl.267",
    pages = "4752--4780",
    ISBN = "979-8-89176-195-7",
}

@inproceedings{Evaluationagent,
author = {Mohammadi, Mahmoud and Li, Yipeng and Lo, Jane and Yip, Wendy},
title = {Evaluation and Benchmarking of LLM Agents: A Survey},
year = {2025},
isbn = {9798400714542},
publisher = {Association for Computing Machinery},
url = {https://doi.org/10.1145/3711896.3736570},
doi = {10.1145/3711896.3736570},
booktitle = {Proceedings of the 31st ACM SIGKDD Conference on Knowledge Discovery and Data Mining V.2},
pages = {6129–6139},
numpages = {11},
series = {KDD '25}
}

@inproceedings{sun-etal-2025-collab,
    title = "Collab-Overcooked: Benchmarking and Evaluating Large Language Models as Collaborative Agents",
    author = "Sun, Haochen  and
      Zhang, Shuwen  and
      Niu, Lujie  and
      Ren, Lei  and
      Xu, Hao  and
      Fu, Hao  and
      Zhao, Fangkun  and
      Yuan, Caixia  and
      Wang, Xiaojie",
    booktitle = "Proceedings of the 2025 Conference on Empirical Methods in Natural Language Processing",
    month = nov,
    year = "2025",
    address = "Suzhou, China",
    publisher = "Association for Computational Linguistics",
    url = "https://aclanthology.org/2025.emnlp-main.249/",
    doi = "10.18653/v1/2025.emnlp-main.249",
    pages = "4922--4951",
    ISBN = "979-8-89176-332-6",
}

@inproceedings{qian-etal-2024-chatdev,
    title = "{C}hat{D}ev: Communicative Agents for Software Development",
    author = "Qian, Chen  and
      Liu, Wei  and
      Liu, Hongzhang  and
      Chen, Nuo  and
      Dang, Yufan  and
      Li, Jiahao  and
      Yang, Cheng  and
      Chen, Weize  and
      Su, Yusheng  and
      Cong, Xin  and
      Xu, Juyuan  and
      Li, Dahai  and
      Liu, Zhiyuan  and
      Sun, Maosong",
    booktitle = "Proceedings of the 62nd Annual Meeting of the Association for Computational Linguistics (Volume 1: Long Papers)",
    month = aug,
    year = "2024",
    address = "Bangkok, Thailand",
    publisher = "Association for Computational Linguistics",
    url = "https://aclanthology.org/2024.acl-long.810/",
    doi = "10.18653/v1/2024.acl-long.810",
    pages = "15174--15186",
}
